\documentclass[lettersize,journal]{IEEEtran}
\usepackage{amsmath,amsfonts}
\usepackage{amssymb}
\usepackage{amsthm}
\usepackage{algorithm}
\usepackage{array}
\usepackage{algpseudocode}
\usepackage[caption=false,font=normalsize,labelfont=sf,textfont=sf]{subfig}
\usepackage{textcomp}
\usepackage{stfloats}
\usepackage{bbding}
\usepackage{url}
\usepackage{pifont}
\usepackage{verbatim}
\usepackage{graphicx}
\usepackage{cite}
\usepackage{multirow}
\usepackage{booktabs}
\usepackage{colortbl}
\usepackage{xcolor}
\usepackage{hyperref}
\hypersetup{hypertex=true,
            colorlinks=true,
            linkcolor=blue,
            anchorcolor=blue,
            citecolor=blue}
\begin{document}

\title{M\textsuperscript{3}Depth: Wavelet-Enhanced Depth Estimation on Mars via Mutual Boosting of Dual-Modal Data}

\author{Junjie Li,~\IEEEmembership{Graduate Student Member,~IEEE,} Jiawei Wang,~\IEEEmembership{Graduate Student Member,~IEEE,} Miyu Li, \\
Yu Liu,~\IEEEmembership{Member,~IEEE,} Yumei Wang,~\IEEEmembership{Member,~IEEE,} and Haitao Xu
\thanks{This work was supported in part by the National Key Research
and Development Program of China under Grant 2022YFB2902705, in part
by Beijing University of Posts and Telecommunications (BUPT) Excellent
Ph.D. Students Foundation under Grant CX20241090, and in part by BUPT
Innovation and Entrepreneurship Support Program under Grant 2025-YC-
T025. \textit{Corresponding author: Yu Liu }(e-mail:liuy@bupt.edu.cn).}
\thanks{J. Li, M. Li, Y. Liu, and Y. Wang are with the School of Artificial Intelligence, Beijing University of Posts and Telecommunications, Beijing 100876, China (e-mail: junjie@bupt.edu.cn; miyuli@bupt.edu.cn; liuy@bupt.edu.cn; ymwang@bupt.edu.cn).}

\thanks{J. Wang is with State Key Laboratory of Networking and Switching Technology, Beijing University of Posts and Telecommunications, Beijing 100876, China (e-mail: wangjiawei98@bupt.edu.cn).}


\thanks{H. Xu is with National Space Science Center, Chinese Academy of Sciences, Beijing 100190, China (e-mail: xuhaitao@nssc.ac.cn)}
}






\maketitle

\begin{abstract}
Depth estimation plays a great potential role in obstacle avoidance and navigation for further Mars exploration missions.  Compared to traditional stereo matching, learning-based stereo depth estimation provides a data-driven approach to infer dense and precise depth maps from stereo image pairs. However, these methods always suffer performance degradation in environments with sparse textures and lacking geometric constraints, such as the unstructured terrain of Mars. To address these challenges, we propose M\textsuperscript{3}Depth, a depth estimation model tailored for Mars rovers. Considering the sparse and smooth texture of Martian terrain, which is primarily composed of low-frequency features, our model incorporates a convolutional kernel based on wavelet transform that effectively captures low-frequency response and expands the receptive field. Additionally, we introduce a consistency loss that explicitly models the complementary relationship between depth map and surface normal map, utilizing the surface normal as a geometric constraint to enhance the accuracy of depth estimation. Besides, a pixel-wise refinement module with mutual boosting mechanism is designed to iteratively refine both depth and surface normal predictions. Experimental results on synthetic Mars datasets with depth annotations show that M\textsuperscript{3}Depth achieves a 16\% improvement in depth estimation accuracy compared to other state-of-the-art methods in depth estimation. Furthermore, the model demonstrates strong applicability in real-world Martian scenarios, offering a promising solution for future Mars exploration missions.

\end{abstract}


\section{Introduction}
\IEEEPARstart{L}{imited} scene perception capabilities have become a critical bottleneck in the traveling speed of current Mars rovers \cite{strader2020perception}, which hinders the efficient completion of scientific tasks. For example, the Curiosity Rover encounters delays and slowdowns when navigating around obstacles like rocks, resulting in an average travel distance of only 28.9 meters per sol \cite{rankin2020driving}. Similarly, the Zhurong Rover covers merely 6.2 meters per sol \cite{zhang2022slip}. Depth estimation holds great potential for enhancing scene perception. It provides a more comprehensive understanding of the 3D structure \cite{metric3d_yin} compared to 2D approaches, such as terrain categorization \cite{panambur2022self} and semantic segmentation \cite{Mars_seg_Yao}. Therefore, 3D perception abilities are essential to support further Martian scientific tasks, as they enable critical operations such as robotic navigation and obstacle avoidance\cite{shim2023swindepth,zhao2025m2cs}. 

Depth information not only enhances navigation capabilities of Mars rovers but also complements traditional downstream 2D vision tasks. As demonstrated in \cite{swan2021ai4mars}, the white trapezoid-shaped marker on the Curiosity Rover serves as a scale bar to roughly estimate depth and further refines the distinction between bedrock and rock categorization. Studies, such as those in
\cite{Mars_seg_Yao,xiong2024light4mars,ma2024automated}, have shown that integrating depth information as auxiliary input can significantly improve semantic segmentation precision. Therefore, it is crucial to investigate stereo depth estimation to improve both the 3D scene perception capabilities of Mars rovers and optimize traditional 2D vision tasks.

\begin{figure}[t!] \centering
    \includegraphics[width=0.48\textwidth]{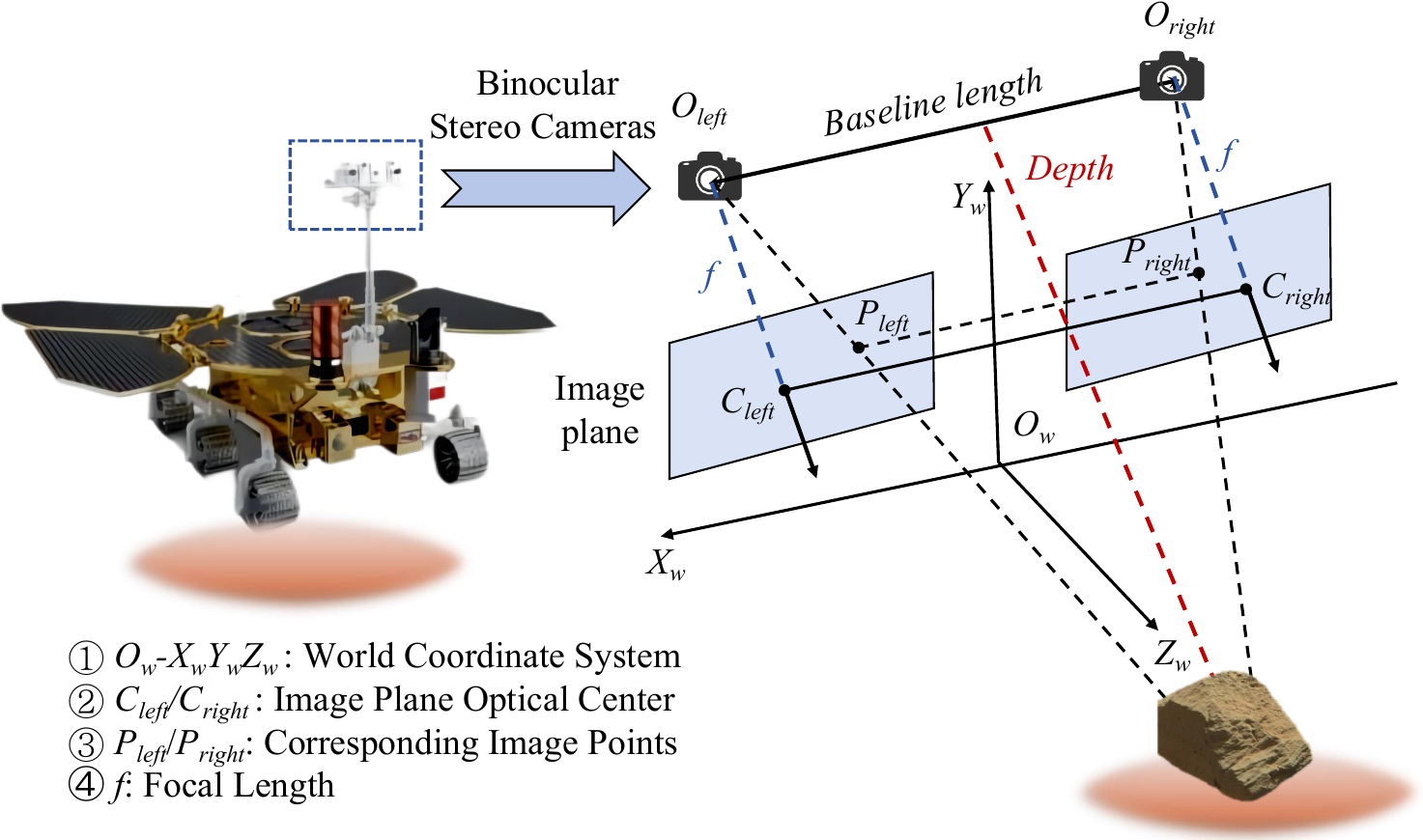}
    \caption{An illustration of China Zhurong Rover stereo camera.} \label{stereo}
\end{figure}

Depth estimation from RGB images has long been explored by the computer vision community. However, this problem is fundamentally ill-posed due to the inherent ambiguity of projecting 3D scenes onto 2D images. One common approach used to tackle this issue is stereo matching, which determines disparity by aligning similar pixel features between rectified left and right image pairs. Mars rovers such as the China Zhurong Rover \cite{liu2022geomorphic} and NASA Perseverance Rover \cite{bell2021mars}, are both equipped with stereo cameras. Leveraging principles of computational photography, these stereo cameras unlock the potential for depth estimation from captured stereo image pairs. As depicted in \autoref{stereo}, with the camera focal length and baseline length provided, the horizontal difference in pixel location (\textit{i.e.} disparity) can be converted into depth based on the geometric principles of parallel binocular vision. 

Mars presents an unstructured environment dominated by smooth, texture-less surfaces and a lack of prominent edges \cite{liu2023hybrid,li2024unsupervised,li2025dustnet}, which are primarily composed of low-frequency signals. Conventional learning-based stereo depth methods struggle to extract meaningful features due to the absence of distinguishable textures when applied in Mars. At the same time, existing neural network architectures exhibit a notable bias toward high-frequency features, as analyzed in \cite{finder2025wavelet,gavrikov2024can}. This bias means that current networks often prioritize sharp and distinct details while being prone to neglecting low-frequency information, further exacerbating the difficulty of feature extraction in the Martian environment.

Traditional stereo matching methods primarily focus on matching pixels across two images captured from slightly different viewpoints, while relying on manual feature descriptors \cite{rao2024cascaded}. 
By utilizing the architectural topology of neural networks, these models establish an effective mapping between the input data and corresponding depth labels in an end-to-end manner, eliminating the reliance on manual feature design \cite{luo2016efficient,vzbontar2016stereo}. Despite these advancements, these models highly depend on distinctive textures for feature extraction, limiting their performance in environments with sparse or ambiguous visual features. Furthermore, relying solely on feature matching constraint between image pairs proves insufficient for precise correspondence estimation \cite{li2021revisiting}. Without explicit geometric constraints (such as epipolar geometry \cite{huang2022h} and planar assumptions \cite{chuah2021deep}), these models often fail to capture the intricate three-dimensional structures of the scene, leading to suboptimal depth estimations. \textbf{\textit{Surface normal map}}, as another modality, is visually discriminative and geometrically informative, offering localized geometric properties compared to the broader geometric attributes of depth \cite{yan2018monocular}. Unlike stereo depth estimation, surface normal maps are not influenced by scale ambiguity and
operate within a compact output space (a unit sphere \textit{vs.} positive real value for depth). Compared to existing geometric constraints, such as epipolar geometry and planar surface assumptions, which typically depend on large-scale training datasets, surface normal estimation methods exhibit stronger generalization capabilities when applied to out-of-distribution images \cite{bae2024rethinking}. Most importantly, as a critical 3D representation form, surface normal map and depth exhibit a strong geometric relationship and are highly complementary \cite{kusupati2020normal}, which makes it feasible to leverage the surface normal as a constraint to enhance Martian depth estimation accuracy.
To summarize, due to Martian unique terrain characteristics, accurate depth estimation on Mars is non-trivial. To address this, two fundamental issues must be considered: 1) \textit{How to eliminate the issues of many neural networks neglecting low-frequency features, which hampers performance in Mar's low-texture, unstructured terrain}, and 2) \textit{How to leverage surface normals as explicit geometric constraints to enhance the model's ability to capture global spatial relationships and achieve more accurate scene depth estimation.} Considering above issues, we aim to emphasize the importance of low-frequency signals by considering the frequency domain characteristics of Martian terrains. Additionally, we exploit the use of surface normal map as an auxiliary modality, leveraging their complementary geometric information to enhance the precision of depth estimation. Specifically, we propose M\textsuperscript{3}Depth, a wavelet-enhanced Depth estimation model on Mars via Mutual boosting of dual-Modal data, as detailed in \autoref{system_model}. Considering the dominance of low-frequency components and the lack of high-frequency features in Martian stereo images, we introduce a wavelet-enhanced convolutional kernel to better retain the low-frequency components, while expanding the receptive field size to capture global scene context more effectively. Moreover, we incorporate the surface normal as a geometric constraint and introduce a consistency loss to explicitly model the inherent geometric relationship between depth map and surface normal map. To further improve the accuracy of depth estimation, we develop a pixel-wise refinement module that iteratively enhances both depth and surface normal predictions through a mutual-boosting mechanism.

In summary, the key contributions of our work are as follows.
\begin{enumerate}
    \item We propose a learning-based end-to-end stereo depth estimation model for Mars rovers, designed to address the challenges posed by Martian unstructured environment. While Mars serves as our primary case study, the approach is generalizable to other similarly challenging planetary exploration scenarios.
    \item We utilize a learnable wavelet-based frequency domain feature extraction module that effectively preserves critical low-frequency information in Martian stereo images while simultaneously expanding the receptive field size.
    \item 
    We introduce a consistency loss to regularize the geometric relationship between depth maps and surface normal maps. In addition, we employ an iterative refinement module to enable mutual boosting between depth and surface normal, achieving more accurate and continuous predictions.
    \item We perform comprehensive evaluations on a high-fidelity synthetic Mars dataset. To support dual-modal learning, we additionally construct reliable surface normal ground truth dataset. Extensive experiments demonstrate improvements in depth estimation accuracy. Besides, we evaluate its adaptability and generalization on real Martian images. 
\end{enumerate}

The remainder of this paper is organized as follows. \autoref{RelatedWork} surveys existing research on stereo depth estimation. \autoref{Overview} presents the proposed framework for depth estimation, while \autoref{Methodology} details the designed methods and modules. Experimental configurations and results are demonstrated in \autoref{Experiments}. Finally, the conclusions are summarized in \autoref{conclusion}.

\section{Related Work}\label{RelatedWork}
In this section, we review the current works and discuss the key ideas that have inspired the design of our approach.
\subsection{Mars Scene Perception}
Datasets form the cornerstone for Mars scene perception tasks. Among the earliest efforts, Ai4Mars \cite{swan2021ai4mars} introduced a Mars terrain segmentation dataset based on images from the Curiosity Rover, focusing on semantic segmentation of Martian surfaces. Subsequently, Zhang \textit{et al.} \cite{zhang2024smars} released the S\textsuperscript{5}Mars dataset, better addressing the challenges of identifying complex terrains. Based on China TianWen-1 mission, Lv \textit{et al.} \cite{lv2022marsnet} introduced the first Mars rock segmentation dataset leveraging images from the Zhurong Rover. In addition to these efforts, Yao and Xiao \textit{et al.} made significant improvements to semantic segmentation networks \cite{liu2023rockformer,xiong2023marsformer} and contributed new datasets, including semantic segmentation dataset (SynMars-TW \cite{xiong2023marsformer}) and panoramic dataset tailored for Martian landscapes \cite{liu2023marsscapes}. To mitigate the scarcity of high-quality real-world data, Ma \textit{et al.} \cite{ma2024automated} generated highly realistic synthetic datasets (SimMars6K), expanding the data resources available for Martian scene perception tasks.

As for Martian scene perception tasks, they can be broadly categorized into 2D (\textit{e.g.} terrain classification \cite{panambur2022self} and semantic segmentation \cite{zhang2024smars,lv2022marsnet,Mars_seg_Yao,xiong2024light4mars,ma2024automated}) and 3D perspectives (\textit{e.g.} stereo-based distance measurement \cite{zheng2023rover} and scene reconstruction \cite{liu2024high}). In terrain classification, Panambur \textit{et al.} \cite{panambur2022self} proposed a self-supervised method to cluster sedimentary textures, enabling the classification of Martian surface materials. Beyond terrain classification, semantic segmentation has been a predominant task to classify the Martian surface into different categories. Such segmentation techniques \cite{zhang2024smars,lv2022marsnet,Mars_seg_Yao,xiong2024light4mars,ma2024automated} help differentiate between various terrain types, focusing primarily on pixel-level labels. However, above approaches focused mainly on 2D texture or category analysis are inherently limited, as they fail to fully capture the three-dimensional structure of the environment. To address this, researchers have also explored 3D scene reconstruction techniques. Zheng \textit{et al.} \cite{zheng2023rover} enhanced stereo matching algorithms for Mars images, enabling a rough estimation of rock distances, while Liu \textit{et al.} \cite{liu2024high} utilized seamless stitching methods to integrate individual block models into a unified coordinate system, producing spatially continuous 3D scene models. Despite these advancements, existing 3D approaches generally offer only coarse geometric information, which highlights the need for refined 3D scene perception techniques.


\subsection{Stereo Depth Estimation}
Building on the success of deep learning, traditional stereo matching \cite{hosni2012fast} have been largely surpassed by learning-based approaches \cite{xu2020aanet,zhang2019ga}, significantly enhancing stereo vision system performance. Modern CNN-based stereo depth estimation methods mostly construct a cost volume by computing matching scores between stereo image pairs, which is then refined by high-level networks to estimate disparities \cite{survey_Laga}. Recently, Transformer-based models have gained significant attention in the depth estimation domain. By leveraging self-attention mechanisms, the Vision Transformer enhances feature extraction and improves the accuracy of dense predictions, making it a promising alternative for stereo depth estimation tasks \cite{li2021revisiting}. Xu \textit{et al.} \cite{xu2023unifying} propose a Transformer-based model utilizing cross-attention, which facilitates cross-view interactions and enhances feature extraction. Tong \textit{et al.} \cite{tong2024robust} propose a Transformer-based disparity estimation network that enhances feature matching through parallax attention, while using focal loss to improve training in ambiguous regions.

However, stereo depth estimation still remains challenging in low-texture regions, where the absence of distinct features complicates reliable correspondence. Inspired by human depth perception, which leverages prior knowledge or contextual information, enabling a more accurate understanding of scenes \cite{survey_Laga}, many stereo depth models have introduced semantic constraints \cite{guney2015displets,kendall2017end,shaked2017improved} and geometric priors \cite{yin2019enforcing,kusupati2020normal} to enhance model robustness and accuracy. While these strategies can mitigate some difficulties, they are often tailored to specific tasks and heavily dependent on large-scale, high-quality training datasets. Recent studies have explored joint optimization paradigms that incorporate surface normal maps. Surface normals are inherently easier to estimate and perceive from visual data, providing a straightforward and effective geometric constraint for depth estimation \cite{metric3d_yin}. Gui \textit{et al.} \cite{gui2024depthfm} and Hu \textit{et al.} \cite{metric3d_yin} propose joint depth-normal optimization methods that distill prior knowledge from large-scale pre-trained models. Meanwhile, Gwangbin \textit{et al.} \cite{bae2022irondepth} and Shao \textit{et al.} \cite{shao2023nddepth} enforce consistency constraint to refine depth estimations by recurrently aligning them with surface normal maps. However, these methods often rely on domain-specific assumptions or large pre-trained models, limiting their adaptability to Mars, which underscores the need for solutions that can operate effectively in the Martian environment.


\subsection{Scene Depth Estimation}
Compared to object-centric stereo depth estimation methods, scene depth estimation focuses on recovering depth throughout an entire scene, which involves modeling broader spatial contexts and understanding the global layout. This requires a larger receptive field to capture the full extent of the scene.

While stereo depth estimation has achieved great success in well-textured Earth-like scenes, its performance often degrades in more challenging environments like Mars, where large-scale, texture-less regions dominate \cite{xiao2021kernel}. Traditional depth estimation techniques rely heavily on local correspondences and high-frequency cues, which are not reliably present in Martian landscapes. Existing neural network architectures are inherently biased towards high-frequency features, which are crucial for edge detection and texture analysis in conventional images \cite{gavrikov2024can,finder2025wavelet}. This bias hinders the extraction in Martian landscapes, also highlights the importance of leveraging frequency domain information to improve scene depth estimation. Most recent works \cite{wang2024selective,zhao2023high} have emphasized high-frequency signals to preserve edges and textures, achieving more clearer boundary preservation. However, these strategies are less effective in Mars. Mars presents unique challenges due to its predominantly texture-less terrain and low-frequency patterns. In this work, our approach concentrates on enhancing the representation of low-frequency components and exploits the use of surface normal as an additional constraint to achieve high accuracy of Martian depth estimation.
\section{Methodology Overview}\label{Overview}
In this section, we introduce the inspiration behind our study and present the main architecture of the M\textsuperscript{3}Depth model. 
\subsection{Preliminaries and Motivation}\label{inspiration}

\textbf{Preliminaries.} Before introducing the architecture of our model, we first review the traditional stereo matching process, as it forms the foundation of our approach. 

\begin{figure}[thb] \centering
    \includegraphics[width=0.48\textwidth]{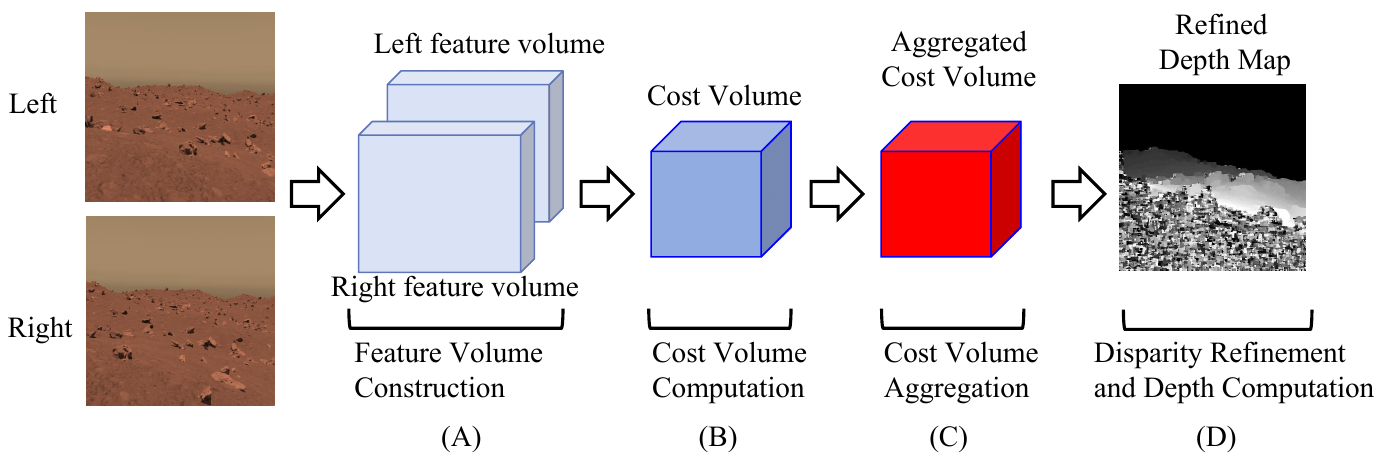}
    \caption{Traditional stereo matching algorithm pipeline.} \label{stereo_pipeline}
\end{figure}

Traditional stereo matching aims to estimate depth by emulating the human binocular vision system \cite{hirschmuller2007stereo}. As illustrated in \autoref{stereo_pipeline}, this process typically involves four key stages: feature volume construction, cost volume computation, cost volume aggregation, and disparity refinement \cite{scharstein2002taxonomy}. Generally, the objective is formulated as an optimization problem that aims to minimize pixel-wise depth discrepancies for more accurate depth estimation \cite{survey_Laga}. Finally, disparity values $d(x,y)$ are converted to depth $D(x,y)$ using the known baseline $B$ and focal length $f$ of the stereo camera setup, given as:
\begin{equation}
{D(x,y)}=\frac{f \times B}{d(x,y)}. \label{Disparity} 
\end{equation}

\begin{figure}[t!] \centering
    \includegraphics[width=0.48\textwidth]{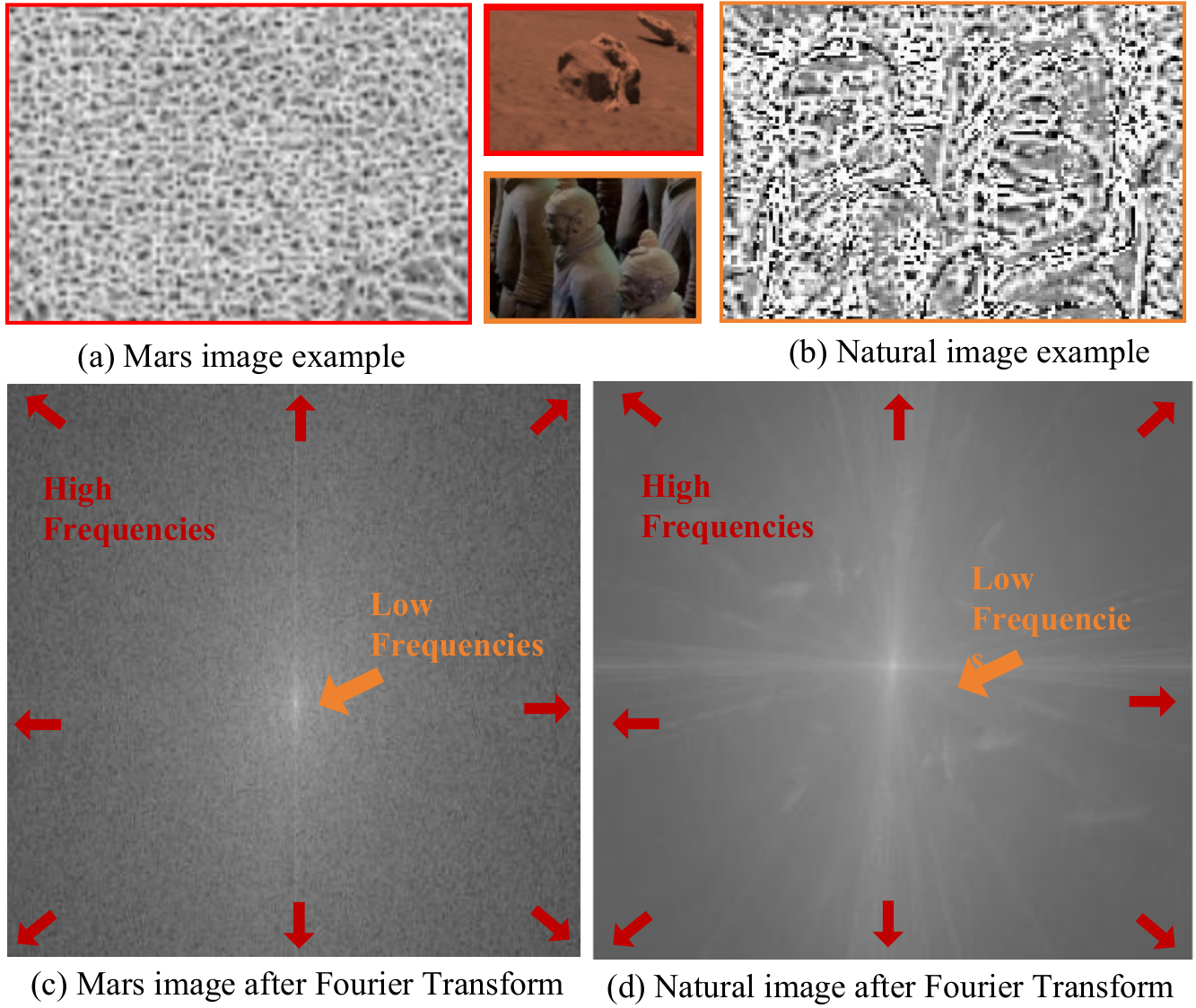}
    \caption{A comparison between a Martian image and a terrestrial image. Subfigures (a) and (b) show the  corresponding LBP grayscale versions. Subfigures (c) and (d) depict the Fourier Transform of original images.} \label{figure1}
\end{figure}

\textbf{Inspiration 1.} Martian surface images exhibit challenging ultra-homogeneous characteristics, similar to Earth's uniform natural terrains (\textit{e.g.}, deserts, salt flats, and ice fields).\footnote{While our work is tailored for Martian terrain, the model’s approach is broadly applicable to other natural environments that exhibit sparse and homogeneous characteristics, where discriminative features are limited.} The sparsity of such features is evidenced by both spatial and spectral analyses:

\begin{itemize}
    \item \textit{\textbf{Spatial Analysis}}: Local Binary Pattern (LBP) visualizations in \autoref{figure1}(a)-(b) reveal fewer edge responses in Martian image compared to natural Earth terrains. The Mars image shows $\sim$19\% lower LBP entropy (3.1 \textit{vs.} 3.7 bits), indicating low-detail surface structure.
    
    \item \textit{\textbf{Frequency Analysis}}: Fourier transforms in \autoref{figure1}(c)-(d) confirm concentration in low-frequency bands ($<$5\% energy above 0.1$f_{\text{max}}$ vs. 32\% in the natural images), signifying dominance of gradual intensity transitions over sharp edges.
\end{itemize}


However, existing neural network architectures, particularly convolutional layers, tend to respond primarily to high-frequency signals \cite{gavrikov2024can,finder2025wavelet}. This tendency can result in an inherent bias that undervalues low-frequency information, which is crucial in Martian surface images. 

These discussions above raise the first inspiration for the model design: \textit{Can we leverage signal-processing techniques to effectively improve the feature extraction of low-frequency regions?}

\begin{figure}[t!] \centering
    \includegraphics[width=0.48\textwidth]{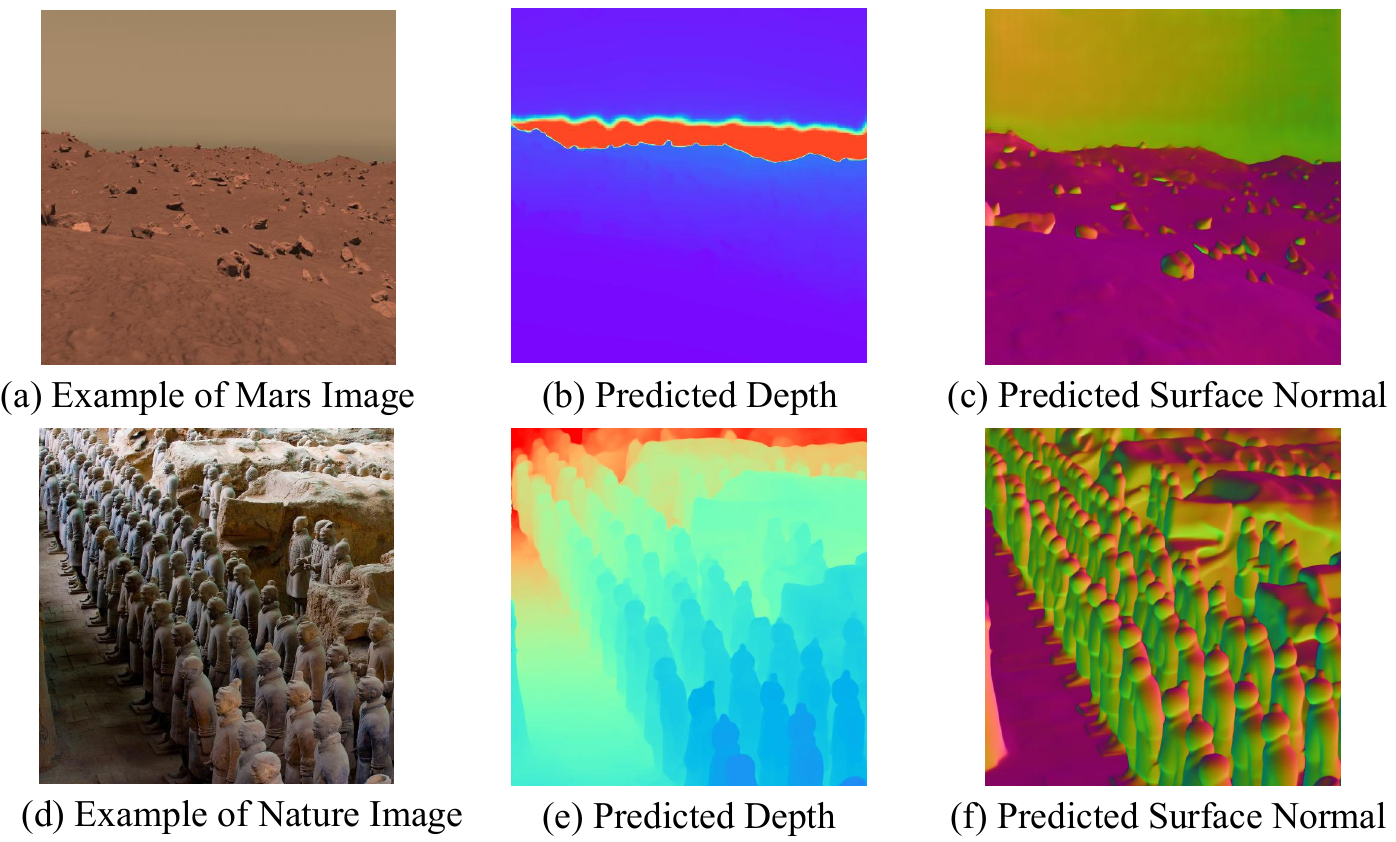}
    \caption{A comparison of depth estimation and surface normal prediction results for Martian (top row) and terrestrial (bottom row) images. (a) and (d) are the original images; (b) and (e) show the predicted depth maps, while (c) and (f) depict the predicted surface normal maps.} \label{figure2}
\end{figure}

\textbf{Inspiration 2.}
In addition to the challenges posed by texture-less regions on Martian surface, we observe that surface normal maps are much easier to infer from visual appearances compared to depth maps. As illustrated in \autoref{figure2}, learning-based methods produce more detailed and accurate predictions for surface normals than for depth maps, as surface normals primarily capture local geometric properties such as orientation and smoothness. In contrast, depth estimation requires integrating global spatial relationships and scene structure, demanding a comprehensive understanding of the broader image context. This phenomenon is especially pronounced on Mars, where homogeneity and the absence of distinct edges make it challenging to infer accurate depth values. Notably, the geometric relationship between surface normals and depth is inherently complementary \cite{qi2018geonet}.



These observations above inspire the second inspiration about the model design motivation: \textit{Can we leverage the surface normal as another modal data to boost the depth estimation in feature-sparse environments?}

\subsection{Model Framework Overview}
\begin{figure*}[thbp] \centering
    \includegraphics[width=\textwidth]{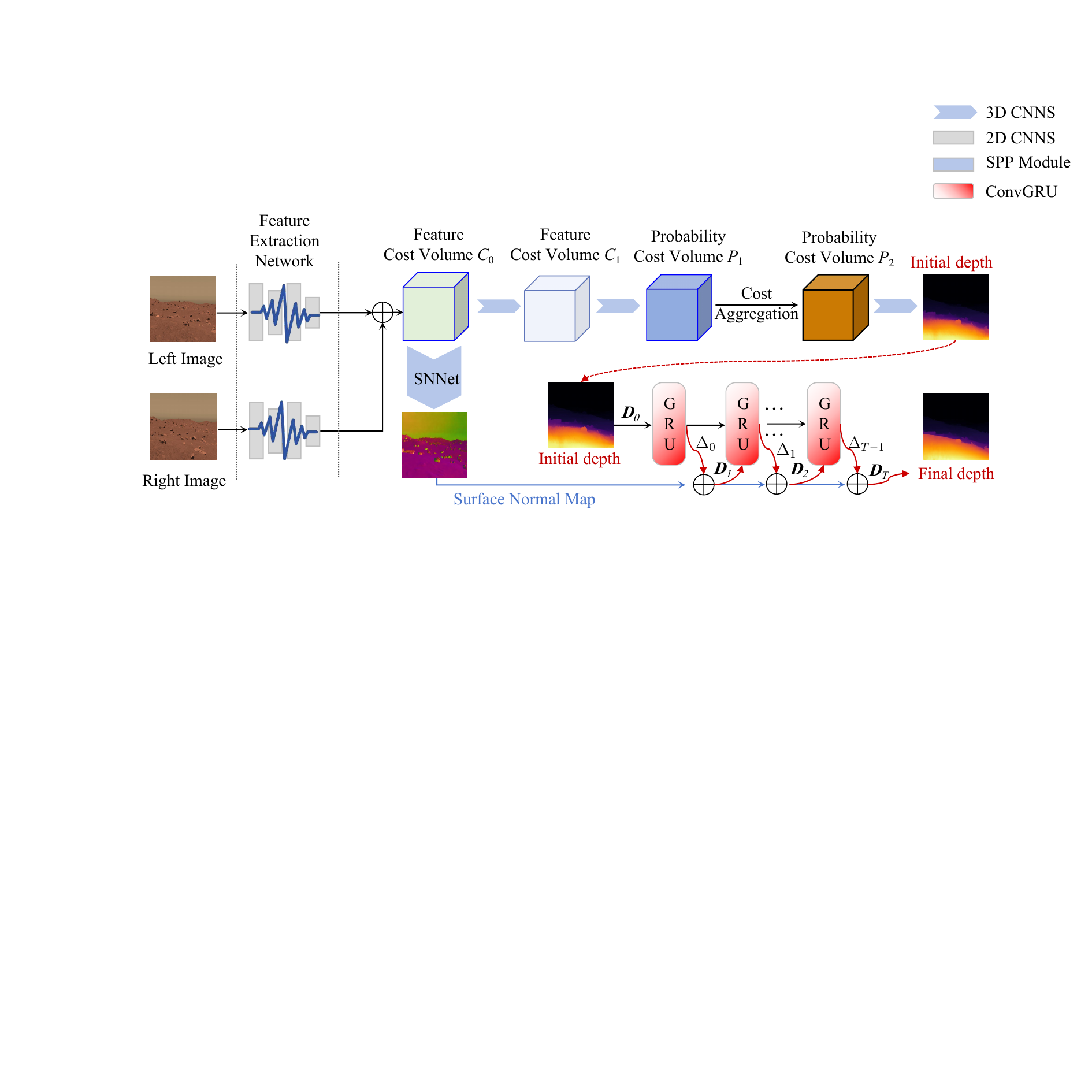}
    \caption{Overview of our M\textsuperscript{3}Depth model. In our model, the left image is served as the reference image and the right image as the target image. The cost volumes are constructed based on the wavelet-enhanced feature extraction network.} \label{system_model}
\end{figure*}

In this section, we present the end-to-end overview pipeline of our proposed M\textsuperscript{3}Depth model, as illustrated in \autoref{system_model}. The network is designed to estimate depth maps from stereo image pairs. Generally, our pipeline consists of multiple key stages, including feature extraction, 4D cost volume construction and cost aggregation, surface normal prediction, and the depth refinement module that incorporates surface normals as geometric constraints. All modules in the proposed architecture are trained in an end-to-end manner. The feature extraction backbone is shared between the depth and normal branches. Gradients are jointly back-propagated to update the network parameters. Notably, our model predicts disparity maps, which are subsequently converted to depth maps using the known stereo camera parameters, as provided in the Eq.\eqref{Disparity}. The camera intrinsics are assumed to be known and are provided as part of the dataset. 

The pipeline begins with a shared CNN-based feature extraction network enhanced by wavelet transforms, applied separately to the left and right images to generate feature representation. The extracted feature maps from the left and right images are then processed to construct an initial 4D feature cost volume ${C_0}$, with dimensions defined as $f \times 
D \times H \times W $, where $f$ denotes the feature dimensions, $D$ is the disparity dimension, and $H$, $W$ represent the image height and width, respectively. This cost volume encodes raw matching costs across all possible disparity levels and serves as the foundation for depth estimation and surface normal map prediction. Subsequently, the initial feature cost volume ${C_0}$ is refined through a series of 3D CNNs to generate feature cost volume ${C_1}$. Then, ${C_1}$ is further processed with additional 3D CNN layers to produce a probability cost volume ${P_1}$, which reflects the disparity probabilities for each pixel. Further cost aggregation is performed, yielding a more precise probability cost volume ${P_2}$. Finally, the final probability cost volume ${P_2}$ is used to compute an initial depth map $\boldsymbol{D}_{0}$ via a \textit{soft argmin} operation over the disparity probabilities.

\begin{figure*}[htb] \centering
\includegraphics[width=\textwidth]{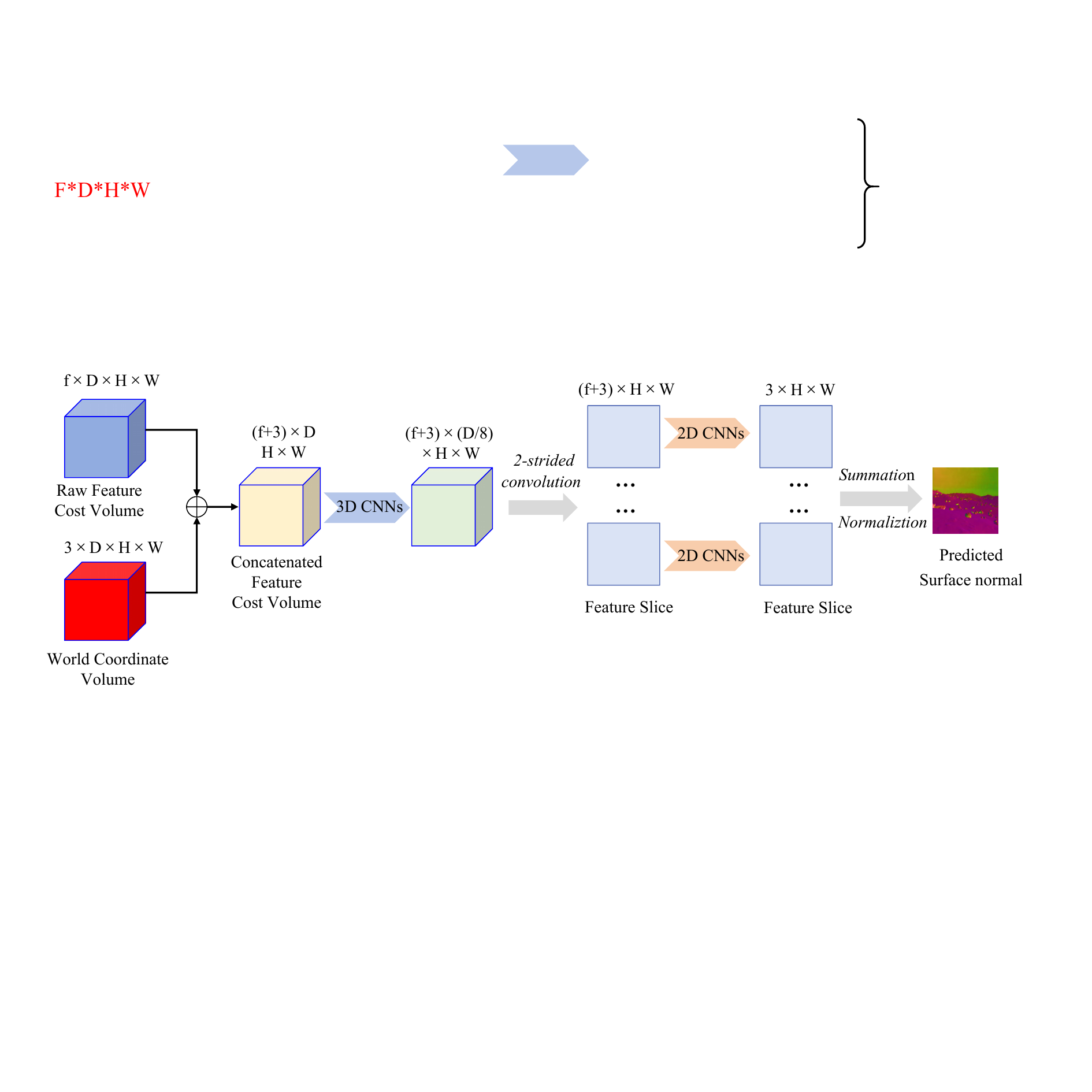}
    \caption{Overview of the surface normal prediction network. The outputs are summed and normalized to generate the final predicted surface normal map.} \label{nnet}
\end{figure*}

To incorporate geometric priors, we learn a surface normal estimation network based on the feature cost volume, which is jointly optimized with the depth estimation pipeline. The relationship between depth and surface normals is explicitly modeled through a consistency constraint loss formulation. The surface normal map is estimated through a dedicated Surface Normal Network (SNNet), as illustrated in \autoref{nnet}. The surface normal map predicted by SNNet is a three-channel output, representing the $n_x$, $n_y$, and $n_z$ components of the unit normal vector at each pixel. While depth and normal maps are initially predicted independently, we introduce the mutual-boosting mechanism in the iterative refinement module. In this stage, the initial depth map and the surface normal map are jointly fed into a ConvGRU-based update module to incrementally update the predicted depth map. During each iteration, each ConvGRU unit computes a residual depth map, which is then added to the previous depth estimate, updating the predicted depth map and achieving higher depth estimation accuracy. 

In our framework, the left image is used as the reference view and the right image as the target view. The model predicts disparity maps aligned with the left image view. We assume all the input stereo image pairs are rectified, such that corresponding points lie on the same horizontal scanline. A more detailed introduction is presented in \autoref{Methodology}, where we systematically decompose our approach into several key components. Detailed training losses and optimization procedures are described in \autoref{lossfunction}.



\section{Module Design Detail }\label{Methodology}
In this section, we present main parts of our M\textsuperscript{3}Depth in detail, including the wavelet-enhanced feature extractor (WEFE), cost volume construction, surface normal map prediction, consistency constraint formulation and iterative refinement module (IRM).

\subsection{Wavelet-Enhanced Feature Extractor}
In Martian depth perception, multiple objects are present compared to conventional object-centric depth estimation. In such scenarios, relying on small receptive fields or local texture cues often leads to ambiguous or unstable depth predictions. A larger receptive field allows the model to integrate contextual signals across multiple objects and structures \cite{ji2021rdrf}. Additionally, conventional convolutional layers prioritize high-frequency information, such as edges and textures, but often neglect low-frequency components that are critical for smooth regions and global context \cite{gavrikov2024can}. 

Taking the inspiration from \cite{finder2025wavelet}, we integrate a wavelet-based operator into our feature extraction module, replacing traditional convolution with Wavelet Transform (WT) operations. The wavelet transform decomposes input signals into spatial-frequency sub-bands, allowing the network to simultaneously extract multi-scale information and implicitly enlarge the receptive field without increasing kernel size \cite{zhou2025wavelet,guo2022hyperspectral}.

\subsubsection{Wavelet-Enhanced Convolution with Cascading Decomposition}

Following \cite{finder2025wavelet}, we utilize the 2-D Haar WT to decompose the input image along both spatial dimensions. This is implemented as a depth-wise convolution operation with a stride of 2 using a filter set $\mathcal{F}_{WT} = [f_{A}, f_{H}, f_{V}, f_{D}]$:
\begin{equation}
\begin{aligned}
f_{A} = \frac{1}{2}\begin{bmatrix}1&1\\1&1\end{bmatrix},& \quad f_{H} = \frac{1}{2}\begin{bmatrix}1&-1\\1&-1\end{bmatrix},\\
f_{V} = \frac{1}{2}\begin{bmatrix}1&1\\-1&-1\end{bmatrix},& \quad f_{D} = \frac{1}{2}\begin{bmatrix}1&-1\\-1&1\end{bmatrix}.
\end{aligned}
\end{equation}
Here, $f_A$ captures the low-frequency (approximation) component, while $f_H$, $f_V$, and $f_D$ extract high-frequency details in the horizontal, vertical, and diagonal directions, respectively. Applying $\mathcal{F}_{WT}$ to an input channel ${X}$ produces four output channels with half the spatial resolution: ${X_A}$ (low-frequency) and ${X_H}, {X_V}, {X_D}$ (high-frequency details). We denote the high-frequency components collectively as $\mathcal{X}_{HF} = [{X_H}, {X_V}, {X_D}]$.

Due to the orthonormality of $\mathcal{F}_{WT}$, the decomposed signals can be perfectly reconstructed via the Inverse Wavelet Transform (IWT), implemented as a transposed convolution:

\begin{equation}
X = \text{IWT}({X_A}, \mathcal{X}_{HF}).
\end{equation}

\subsubsection{Cascading Decomposition and Feature Extraction}

To capture multi-scale information efficiently, we recursively apply the WT only to the low-frequency component at each level. Starting with the input image $X^{(0)}_A = X$:

\begin{equation}
X^{(i)}_A, \mathcal{X}^{(i)}_{HF} = \text{WT}\left(X^{(i-1)}_A\right), \quad i = 1, \dots, \ell
\label{eq:decomp}
\end{equation}
where $i$ denotes the decomposition level, $X^{(i)}_A$ is the level-$i$ low-frequency component, and $\mathcal{X}^{(i)}_{HF} = [X^{(i)}_H, X^{(i)}_V, X^{(i)}_D]$ are the level-$i$ high-frequency components. $\ell$ is the total number of decomposition levels.

At each level $i$ (including level $0$), a $k \times k$ depth-wise convolution (with weights $W^{(i)}$) is applied to all frequency components present at that level to extract features:

\begin{equation}
\begin{aligned}
    \mathbf{Y}^{(i)} &= \left[Y^{(i)}_A, Y^{(i)}_H, Y^{(i)}_V, Y^{(i)}_D\right] \\
     & = \text{Conv}\left(W^{(i)}, \left[X^{(i)}_A, \mathcal{X}^{(i)}_{HF}\right]\right), \quad i = 0, \dots, \ell
\end{aligned}
\label{eq:conv}
\end{equation}

Specifically, for $i=0$, $X^{(0)}_A = X$, $\mathcal{X}^{(0)}_{HF}$ is empty. Convolution operates solely on $X^{(0)}_A$, yielding $Y^{(0)}_A$. For $i \ge 1$,  convolution operates on all four components $X^{(i)}_A, X^{(i)}_H, X^{(i)}_V, X^{(i)}_D$ independently. We denote the extracted low-frequency feature at level $i$ as $Y^{(i)}_A$ and the extracted high-frequency features collectively as $\mathcal{Y}^{(i)}_{HF} = [Y^{(i)}_H, Y^{(i)}_V, Y^{(i)}_D]$.

\subsubsection{Hierarchical Feature Integration}

To integrate features across scales while emphasizing low-frequency information propagation, we reconstruct from the deepest level upwards. We initialize the deepest aggregated feature as zero ($Z^{(\ell+1)} = \mathbf{0}$). For levels $i = \ell, \ell-1, \dots, 1$:

\begin{equation}
Z^{(i)} = \text{IWT}\left( \beta^{(i)} \cdot (Y^{(i)}_A + Z^{(i+1)}), \mathcal{Y}^{(i)}_{HF} \right),
\label{eq:fusion}
\end{equation}
here, $Z^{(i)}$ denotes the aggregated feature reconstructed to the spatial resolution of level $i-1$. $\beta^{(i)}$ is the adaptive amplification coefficient for low-frequency components. $Y^{(i)}_A$ represents the low-frequency features extracted at level $i$. $Z^{(i+1)}$ denotes the aggregated feature from the deeper level $i+1$. $\mathcal{Y}^{(i)}_{HF}$ is the high-frequency features extracted at level $i$.

The key step $Y^{(i)}_A + Z^{(i+1)}$ combines features extracted from the current coarse scale ($Y^{(i)}_A$) with progressively refined features integrated from finer scales ($Z^{(i+1)}$). This sum, along with the current high-frequency features ($\mathcal{Y}^{(i)}_{HF}$), is then upsampled via IWT to generate $Z^{(i)}$ at the previous level's resolution.

Finally, the level $0$ output is combined with the first integrated feature:
\begin{equation}
Z^{(0)} = Y^{(0)}_A + Z^{(1)},
\label{eq:output}
\end{equation}
this $Z^{(0)}$ is the final output feature map at the original input resolution.

\subsubsection{Receptive Field Expansion}

\begin{figure}[t!] \centering
\includegraphics[width=0.48\textwidth]{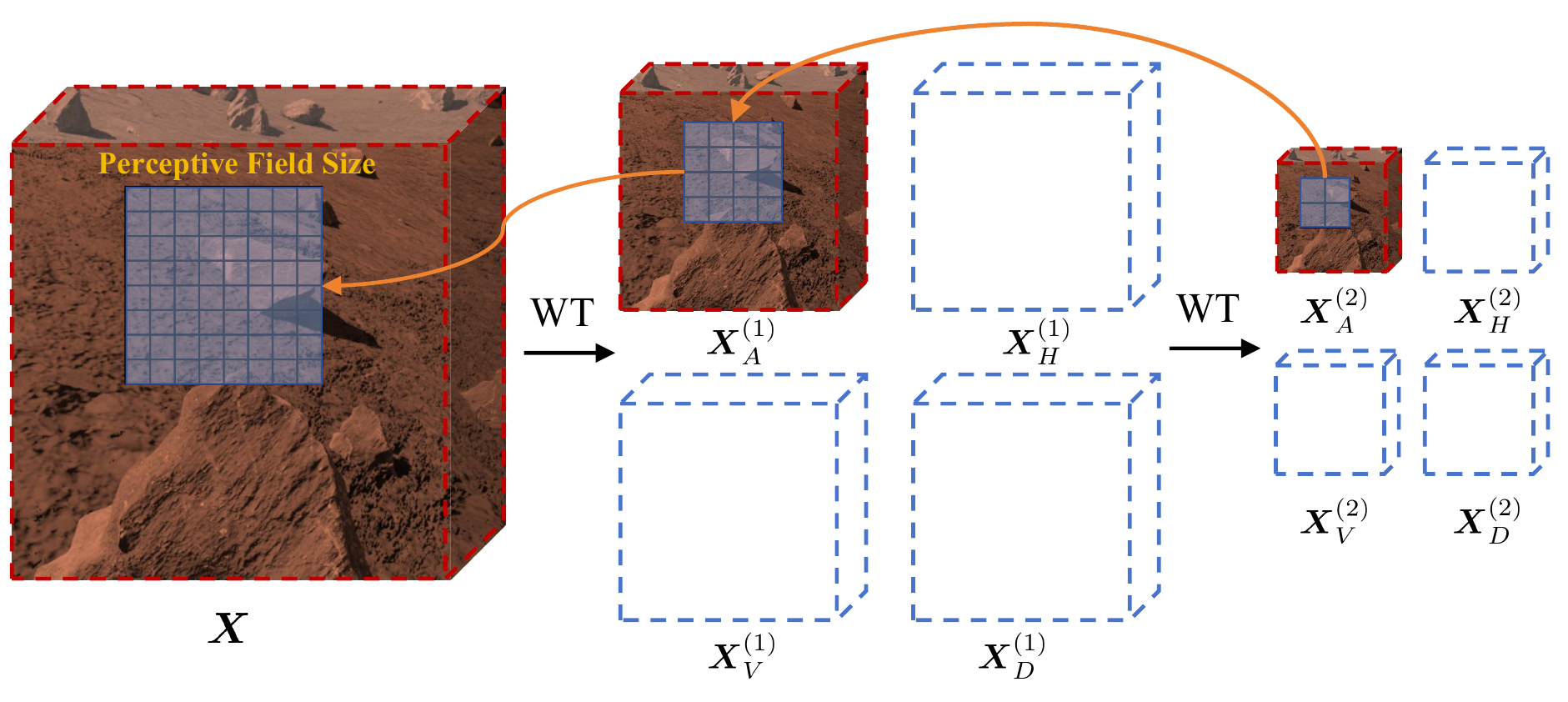}
    \caption{Illustration of hierarchical decomposition using wavelet transform. The diagram demonstrates the process of dividing an image across multiple levels, emphasizing the progressive increase in receptive field size.} 
    \label{erf}
\end{figure}

Each level of decomposition reduces the spatial resolution by half. For example, in \autoref{erf} performing a $2 \times 2$ convolution on the second-level low-frequency component $X_A^{(2)}$ corresponds to an effective receptive field of $8 \times 8$ in the original input space. Consequently, applying convolution ($k \times k$ kernels) at level $i$ operates on a feature map of size $(H / 2^i) \times (W / 2^i)$. Crucially, each element in this map corresponds to a region of size $(2^i \times 2^i)$ in the original input. Therefore, the effective receptive field (ERF) of a convolution operation at level $i$ in the original input space is:

\begin{equation}
\text{ERF}_{\text{level } i} = (2^i \times k) \times (2^i \times k).
\label{eq:erf}
\end{equation}

The hierarchical decomposition achieves exponential receptive field growth with decomposition level $i$, as convolutions are performed at reduced resolutions. Deeper layers ($i > 1$) can thus capture extensive global context using relatively small kernels ($k$). The overall operation
details are illustrated in Algorithm \ref{alg:wtconv}.

\begin{algorithm}[t]
\caption{Wavelet-Enhanced Convolution}
\label{alg:wtconv}
\begin{algorithmic}[1]
\Require{Input image $X \in \mathbb{R}^{C \times H \times W}$, number of decomposition levels $\ell$, convolution kernel size $k$.}
\Ensure{Enhanced feature map $Z^{(0)} \in \mathbb{R}^{C' \times H \times W}$ \Comment{$C'$ depends on $W^{(i)}$ channel settings}}
\State Initialize: $X^{(0)}_A \gets X$
\State $Y^{(0)}_A \gets \text{Conv}(W^{(0)}, X^{(0)}_A)$ \Comment{Level 0 convolution (only low-pass)}
\For{$i = 1$ \textbf{to} $\ell$} 
    \Comment{Forward decomposition \& feature extraction}
    \State $X^{(i)}_A, X^{(i)}_H, X^{(i)}_V, X^{(i)}_D \gets \text{WT}(X^{(i-1)}_A)$ \Comment{Eq. \eqref{eq:decomp}}
    \State $\begin{aligned}[t]
    &[Y^{(i)}_A, Y^{(i)}_H, Y^{(i)}_V, Y^{(i)}_D] \\
    &\quad= \text{Conv}(W^{(i)}, [X^{(i)}_A, X^{(i)}_H, X^{(i)}_V, X^{(i)}_D])
\end{aligned}$ \Comment{Eq. \eqref{eq:conv}}
    \State Set $\mathcal{Y}^{(i)}_{HF} \gets [Y^{(i)}_H, Y^{(i)}_V, Y^{(i)}_D]$
\EndFor
\State Initialize: $Z^{(\ell+1)} \gets \mathbf{0}$ \Comment{Start integration at deepest level}
\For{$i = \ell$ \textbf{to} $1$} 
    \Comment{Backward feature integration (coarse to fine)}
    \State $Z^{(i)} \gets \text{IWT}( \beta^{(i)} \cdot (Y^{(i)}_A + Z^{(i+1)}), \mathcal{Y}^{(i)}_{HF} )$ \Comment{Eq. \eqref{eq:fusion}}
\EndFor
\State $Z^{(0)} \gets Y^{(0)}_A + Z^{(1)}$ \Comment{Eq. \eqref{eq:output}} - Final combination
\State \Return $Z^{(0)}$
\end{algorithmic}
\end{algorithm}

\subsection{Cost Volume Construction and Surface Normal Map Prediction}
Existing stereo depth approaches can be categorized into 3D and 4D convolution-based frameworks, distinguished by how the cost volume is constructed \cite{im2019dpsnet}. 3D convolution-based methods aggregate costs across height, width, and disparity \cite{xu2022attention,shen2021cfnet}. 4D convolution-based methods retain an additional feature dimension for enhanced depth estimation \cite{nie2019multi,wu2019semantic}. Given rectified stereo image pairs, we extract deep features from the left (reference) and right (target) images using a shared CNN-based backbone enhanced with wavelet transforms. The details of this feature extraction module are shown in \autoref{feature-extractor}. Specifically, the feature extractor begins with an initial 7$\times$7 convolutional layer, followed by 5 additional 3$\times$3 convolutional layers. Each convolutional layer is followed by a non-linear activation and normalization. Wavelet transforms are applied at two intermediate stages in the backbone, where they substitute standard convolutions to explicitly decompose features into low- and high-frequency components. 


\begin{table*}[ht]
\centering
\caption{Architecture of Feature Extraction and Cost Volume Construction ($f=64$)}
\label{feature-extractor}
\begin{tabular}{l|c|c|c|c|c}
\toprule
\multirow{2}{*}{Layer} & \multirow{2}{*}{Type} & Kernel & \multirow{2}{*}{Stride} & \multirow{2}{*}{Padding} & Output \\
 & & Size & & & Shape \\
\midrule
Input & RGB Image & -- & -- & -- & $3 \times 512 \times 512$ \\
\midrule
Conv1 & Conv2D & $7 \times 7$ & 2 & 3 & $64 \times 256 \times 256$ \\
Conv2 & Conv2D & $3 \times 3$ & 1 & 1 & $64 \times 256 \times 256$ \\
\midrule
Conv3 & Conv2D  & $3 \times 3$ & 2 & 1 & $128 \times 128 \times 128$ \\
Conv4 & Conv2D & $3 \times 3$ & 1 & 1 & $128 \times 128 \times 128$ \\
\midrule
Conv5 & Conv2D & $3 \times 3$ & 2 & 1 & $256 \times 64 \times 64$ \\
Conv6 & Conv2D & $3 \times 3$ & 1 & 1 & $256 \times 64 \times 64$ \\
\midrule
\multicolumn{6}{c}{\textbf{Cost Volume Construction}} \\
\midrule
\textit{Left Features} & Feature Map & -- & -- & -- & $256 \times 64 \times 64$ \\
\textit{Right Features} & Feature Map & -- & -- & -- & $256 \times 64 \times 64$ \\
Cost Volume & Unfold \& Compare & -- & -- & -- & $64 \times D \times 64 \times 64$ \\
\midrule
\multicolumn{6}{c}{\textbf{Cost Volume Processing}} \\
\midrule
Conv7 & Conv3D & $3 \times 3 \times 3$ & 1 & 1 & $128 \times D \times 64 \times 64$ \\
Conv8 & Conv3D & $3 \times 3 \times 3$ & 1 & 1 & $64 \times D \times 64 \times 64$ \\
\midrule
\multicolumn{6}{c}{\textbf{Output Head}} \\
\midrule
Upsample1 & Transposed Conv & $4 \times 4$ & 2 & 1 & $64 \times D \times 128 \times 128$ \\
Upsample2 & Transposed Conv & $4 \times 4$ & 2 & 1 & $64 \times D \times 256 \times 256$ \\
Upsample3 & Transposed Conv & $4 \times 4$ & 2 & 1 & $64 \times D \times 512 \times 512$ \\
\bottomrule
\end{tabular}
\end{table*}

For each candidate disparity level, we warp the target image’s feature map to the coordinate frame of the reference image. The warped features and the corresponding reference features are then used to compute a similarity measure, yielding a set of cost maps, each corresponding to a specific disparity hypothesis. Stacking these across all disparity levels results in a 4D cost volume $C_0\in\mathbb{R}^{f \times D \times H \times W }$, where $f$ is the number of feature channels, $H$, $W$ are spatial dimensions and $D$ is the number of disparity levels. Following DPSNet \cite{im2019dpsnet}, we interpret the cost volume as a stack of such slices and process them using a series of 3D convolutions layers. The initial feature cost volume $C_0$ is refined through a series of 3D CNNs to generate
feature cost volume $C_1$. Further 3D convolutions are applied to $C_1$ to produce a probability volume $P_1\in\mathbb{R}^{D\times H\times W}$, where each pixel has a softmax-normalized distribution over all disparity levels. Inspired by cost aggregation techniques in traditional stereo and context-aware refinement in DPSNet, we further introduce a residual cost aggregation module to process $P_1$, resulting in a more robust cost volume $P_2$. After aggregation, we apply a softmax operation along the disparity axis to produce a per-pixel disparity probability distribution. The initial depth map $\boldsymbol{D}_0$ is obtained via soft argmin over this distribution.

Beyond the depth estimation branch, we introduce the surface normal prediction network (SNNet). It outputs a three-channel normal vector per pixel, which lies on the unit sphere. Since the surface normal at a pixel is determined by the orientation of the surface at its true depth, the accuracy of normal prediction relies on the alignment between depth hypotheses and the true depth of the surface. To address the inherent ambiguity in depth estimation, the depth range is discretized into multiple disparity level (\textit{i.e.}, depth slices), each corresponding to a specific depth hypothesis. Slices closer to the true depth contribute more accurate local geometric information, while slices farther away contribute less and are effectively suppressed. Therefore, we employ the slice-based surface normal prediction network, denoted as $\text{SNNet}(S_i)$, which utilizes the feature cost volume $C_0$ to predict the surface normal map. As shown in \autoref{nnet}, the cost volume encapsulates both spatial and visual feature information, forming the foundation for surface normal estimation. The concatenated feature volume $(f+3)  \times D \times H \times W$ is processed by 3D convolutions with a stride of 2 along the depth dimension, reducing its size and generating a compact representation while preserving key spatial features. Following this, the processed feature volume is sliced along the depth dimension into $\frac{D}{8}$ depth slices. Each slice-specific 2D CNN predicts a preliminary surface normal map for its corresponding depth hypothesis. The details of the SNNet are shown in \autoref{ssnet-arch}.
The final normal map, $\vec{n}$
is computed by summing the normal estimates from all slices and normalizing the result, as shown by:

\begin{equation}\vec{n}=\frac{\sum_{i=1}^{D/8}\text{SNNet}(S_{i})}{\left\|\sum_{i=1}^{D/8}\text{SNNet}(S_{i})\right\|_{2}}.\end{equation}


\begin{table*}[htbp]
\centering
\caption{Architecture of SNNet for surface normal prediction ($f=64$)}
\label{ssnet-arch}
\begin{tabular}{@{}lccc@{}}
\toprule
\textbf{Stage} & \textbf{Input Size} & \textbf{Operation} & \textbf{Output Size}\\
\midrule
\multirow{2}{*}{0} & $64 \times D \times 512 \times 512 $ & \multirow{2}{*}{Concatenation} & \multirow{2}{*}{$67 \times D \times 512 \times 512$} \\
& $3 \times D \times 512 \times 512 $ & &  \\
1 & $67 \times D \times 512 \times 512 $ & 3D Conv + BN + ReLU × 3 & $67 \times D' \times 512 \times 512$ \newline ($D' = D/8$)  \\
2 & $67 \times D' \times 512 \times 512 $ & Depth-wise slicing & $D'$ slices of $67 \times 512 \times 512$ \\
3 & $67 \times 512 \times 512$ (each slice) & 2D Conv + BN + ReLU × 2 & $3 \times 512 \times 512$ (each slice) \\
4 & $D'$ feature slices & Summation + L2 normalization & $3 \times 512 \times 512$ \\
\bottomrule
\end{tabular}
\end{table*}


\subsection{Consistency Constraint
Formulation}
To jointly optimize the depth and surface normal predictions within our model framework, we introduce a consistency constraint to model the explicit geometric alignment between the two maps. Leveraging the pinhole camera model, we first establish the geometric correlation between depth and surface normal and then derive the consistency constraint from their intrinsic geometric correlation.

\begin{figure}[thb] \centering
    \includegraphics[width=0.48\textwidth]{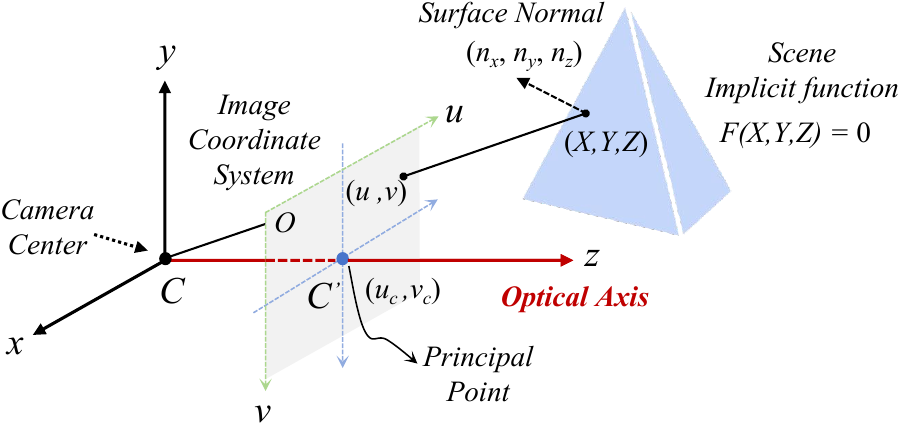}
    \caption{Pinhole camera model and the scene implicit function.} \label{pinhole_camera_o}
\end{figure}

Under the pinhole camera model, the spatial gradient of the depth map in the pixel coordinate system is computed by integrating information from both the depth and surface normal maps. As illustrated in \autoref{pinhole_camera_o}, $(u,v)$ represents the pixel coordinate corresponding to the 3D point $(X,Y,Z)$, and $(u_c,u_v)$ denotes the pixel coordinate of the camera's optical center. Additionally, $f_x$ and $f_y$ represent the focal lengths along the X-axis and Y-axis, respectively.

Based on fundamental principles of photogrammetry, the pixel coordinates $(u,v)$ can be transformed into three-dimensional spatial coordinates $(X,Y,Z)$, as Eq.\eqref{eq1_o} depicts. 

\begin{equation}
\begin{bmatrix} u \\ v \\ 1 \end{bmatrix} = \frac{1}{Z} \begin{bmatrix} f_x & 0 & u_c \\ 0 & f_y & v_c \\ 0 & 0 & 1 \end{bmatrix} \begin{bmatrix} X \\ Y \\ Z \end{bmatrix}.
\label{eq1_o}
\end{equation}

Based on the pinhole camera model and the chain rule, we can yield equation below:

\begin{equation}
\left\{
\begin{aligned}
\frac{\partial X}{\partial v} &= \frac{u-u_c}{f_x}\frac{\partial Z}{\partial v}. \\
\frac{\partial Y}{\partial v} &= \frac{v-v_c}{f_y}\frac{\partial Z}{\partial v} + \frac{Z}{f_y}.
\end{aligned}
\right.
\label{eq3_o}
\end{equation}

Depth and surface normal information are inherently related through the geometry of the 3D scene. To ensure consistency between these two modalities, we consider two types of constraints as follows. Due to space limitations, the detailed derivation of the geometric relationship and constraint formulation is provided in the \autoref{appendix}.

\textit{Constraints 1: Geometric Mathematical Constraint.} 

The first constraint is derived from the mathematical relationship between depth gradients and surface normals. The surface normal can be expressed as the gradient of an implicit function, and this relationship provides a direct linkage between depth gradients and normal vectors.

Under the surface function assumption, the scene can be represented as an implicit function $F(X,Y,Z)=0$. The surface normal map, denoted as $\vec{n} = (n_x,n_y,n_z)$, can be interpreted as the gradient of this function:

\begin{equation}
\nabla F=(\frac{\partial F}{\partial X},\frac{\partial F}{\partial Y},\frac{\partial F}{\partial Z}) = (n_x,n_y,n_z).
\label{eq4_o}
\end{equation}




Then, we can yield equation below:

\begin{equation}
\left(\frac{\partial Z}{\partial u}\right)_1=\frac{\left(\frac{-n_xZ}{n_zf_x}\right)}{1+\frac{n_x}{n_z}\frac{u-u_c}{f_x}+\frac{n_y}{n_z}\frac{y-u_c}{f_y}}.
\end{equation}

\begin{equation}\left(\frac{\partial Z}{\partial v}\right)_1=\frac{(\frac{-n_yZ}{n_zf_y})}{1+\frac{n_x}{n_z}\frac{u-u_c}{f_x}+\frac{n_y}{n_z}\frac{v-v_c}{f_y}}\end{equation}

\textit{Constraint 2: Spatial Gradient Constraint.}

The spatial gradient of the depth map can be efficiently estimated using standard image processing techniques, such as the Sobel filter, which approximates local depth variations in the image. This gradient captures changes in depth along the horizontal and vertical directions, providing a data-driven estimate of the surface geometry structure. The spatial gradient of the depth map can be computed by using a Sobel filter:

\begin{equation}
\begin{aligned}
\left(\frac{\partial Z}{\partial u},\frac{\partial Z}{\partial v}\right)_2=\left(\frac{\Delta Z}{\Delta u},\frac{\Delta Z}{\Delta v}\right). \\
\end{aligned}
\end{equation}

The consistency constraint is formulated as deviation between the two constraints of $\left(\frac{\partial Z}{\partial u}, \frac{\partial Z}{\partial v}\right)_1$ and $\left(\frac{\partial Z}{\partial u}, \frac{\partial Z}{\partial v}\right)_2$. The consistency constraint is further formulated as a consistency loss function, enabling it to guide model training by aligning the two gradient estimates. This loss is defined as the Huber norm of their deviation, as shown in Eq.\eqref{eq9_o}. 

\begin{equation}
\mathcal{L}_c=\left|\left(\frac{\partial Z}{\partial u},\frac{\partial Z}{\partial v}\right)_1-\left(\frac{\partial Z}{\partial u},\frac{\partial Z}{\partial v}\right)_2\right|_{\boldsymbol{H}}.
\label{eq9_o}
\end{equation}

\subsection{Iterative Refinement Module}
To facilitate the mutual boosting across depth and surface normal predictions, we implement a learning-based optimization with recurrent refinement blocks, as illustrated in \autoref{system_model} and \autoref{gru}. Unlike previous methods \cite{kusupati2020normal}, our approach iteratively updates both depth and surface normals through a unified refinement process. $\hat{\boldsymbol{D}}_{t}$ and $\hat{\boldsymbol{N}}_{t}$ represent the depth map and surface normal map refined at iteration step $t$, where  $t\in[0,1,...,T]$ denotes the iteration index. Inspired by \cite{metric3d_yin,lipson2021raft}, we adopt a convolutional gated recurrent unit
(ConvGRU) to yield these updates. ConvGRU serves as an efficient recurrent mechanism that maintains spatial structure while modeling temporal refinement dependencies, making it particularly suitable for multi-step geometric optimization.



Grounded on the initial predictions, $\hat{\boldsymbol{D}}_0$ and $\hat{\boldsymbol{N}}_0$, the depth and surface normal maps are iteratively refined using updates $\Delta\hat{\boldsymbol{D}}_{t+1}$ and $\Delta\hat{\boldsymbol{N}}_{t+1}$ at each step. In general, the refinement process is governed by a recurrent block $\Phi$, which computes updates for both depth and surface normal predictions, as follows:

\begin{equation}
\Delta\hat{\boldsymbol{D}}_{t+1},\Delta\hat{\boldsymbol{N}}_{t+1},\boldsymbol{H}_{t+1}=\Phi(\hat{\boldsymbol{D}}_{t},\hat{\boldsymbol{N}}_{t},\boldsymbol{H}_{t},\boldsymbol{H}_0),
\end{equation}
where $\boldsymbol{H}_t$ denotes the hidden feature state at iteration $t$. The depth and surface normals are updated iteratively as:

\begin{equation}
\hat{\boldsymbol{D}}_{t+1}\leftarrow\hat{\boldsymbol{D}}_t+\Delta\hat{\boldsymbol{D}}_t,\hat{\boldsymbol{N}}_{t+1}\leftarrow\hat{\boldsymbol{N}_t}+\Delta\hat{\boldsymbol{N}}_t,
\end{equation}
where $\Delta\hat{\boldsymbol{D}}_t$ and $\Delta\hat{\boldsymbol{N}}_t$ represent the respective updates at iteration $t$.

In our IRM design, the recurrent block $\Phi$ consists of a ConvGRU sub-block and two projection heads, $\Phi_d$ and $\Phi_n$, which predict updates $\Delta\hat{\boldsymbol{D}}_{t+1}$ and $\Delta\hat{\boldsymbol{N}}_{t+1}$ respectively. Each projection head takes the updated hidden state and current predictions as input and outputs residual updates. Specifically:

\begin{equation}
\left\{
\begin{aligned}
    \Delta\hat{\boldsymbol{D}}_{t+1} &=\Phi_d(\hat{\boldsymbol{D}}_t,\boldsymbol{H}_{t},\boldsymbol{H}_{0}),\\
    \Delta\hat{\boldsymbol{N}}_{t+1}&=\Phi_n(\hat{\boldsymbol{N}}_t,\boldsymbol{H}_{t},\boldsymbol{H}_{0}),
\end{aligned}
\right.
\end{equation}
where $\boldsymbol{H}_{t+1}$ is the updated hidden state computed by the ConvGRU, combining the information from the current depth and surface normal maps:

\begin{figure}[thb] \centering
    \includegraphics[width=0.48\textwidth]{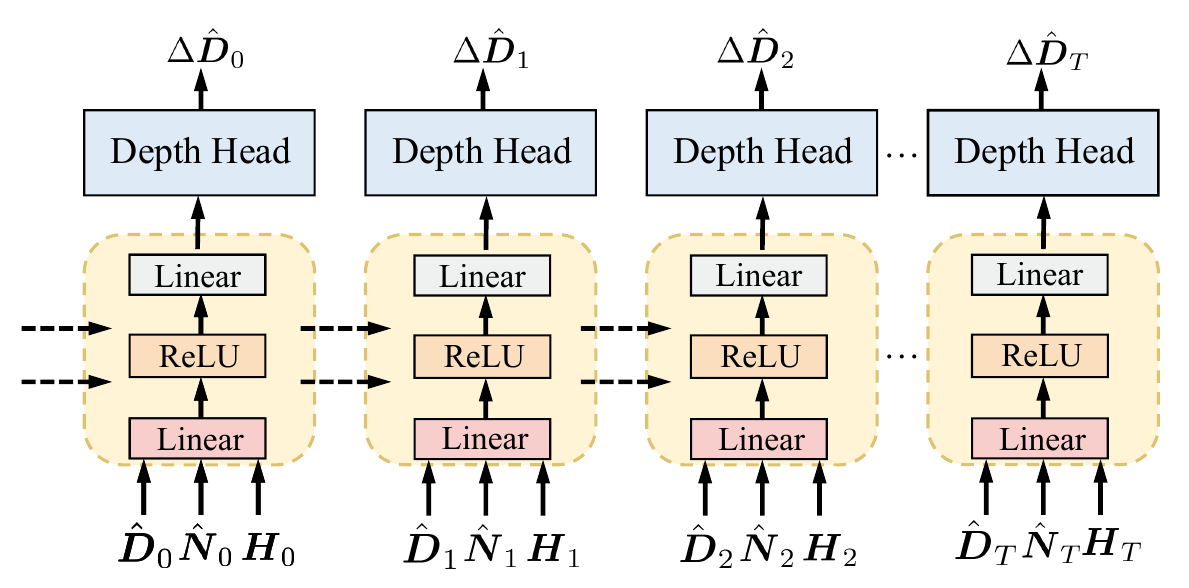}
    \caption{Structure of the ConvGRU update block, taking the depth head as an example.} \label{gru}
\end{figure}

\begin{equation}
\boldsymbol{H}_{t+1} = \text{ConvGRU}(\hat{\boldsymbol{D}}_t,\hat{\boldsymbol{N}}_t,\boldsymbol{H}_0,\boldsymbol{H}_t).
\end{equation}

After $T$ refinement steps, the process yields the final optimized depth map $\hat{\boldsymbol{D}}_T$ and surface normal map $\hat{\boldsymbol{N}}_T$, capturing the synergistic relationship between the two modalities for improved accuracy.

\section{Experiments}\label{Experiments}
We describe the supervised learning strategy in our framework, including the training objective, the design of loss functions. Then, we perform comprehensive experiments using Mars datasets, including both a synthetic Mars dataset and real-world scenario captured by Zhurong Rover. We present both quantitative and qualitative comparisons with current state-of-the-art (SOTA) competitors. To validate the efficacy of each key component, we conduct ablation studies both quantitative and qualitative. Finally, we discuss potential applications and highlight the future works. In addition, we include supplementary experiments in the \autoref{appendix}. 
\subsection{Supervision Formulation}\label{lossfunction}
To improve the accuracy and smoothness of the depth estimation, we integrate multiple complementary loss functions into the training process. These losses are designed to leverage both pixel-wise supervision and geometric consistency between depth and surface normals. Altogether, our approach incorporates the following three key loss functions during model training:

\begin{enumerate}
    \item \textit{Pixel-Wise Depth Loss} $\mathcal{L}_{d}$: measures the discrepancy between the predicted depth map $\hat{\boldsymbol{D}}$ and the ground truth depth map $\boldsymbol{D}_{gt}$, aiming to minimize the pixel-wise depth error.
    \item \textit{Pixel-Wise Surface Normal Loss} $\mathcal{L}_{n}$: quantifies the difference between the predicted surface normal map $\hat{\boldsymbol{N}}$ and the ground truth normals $\boldsymbol{N}_{gt}$, ensuring accurate prediction of surface normal.
    \item \textit{Consistency Constraint Loss} $\mathcal{L}_{c}$: enforces geometric consistency between the depth map and surface normal predictions, as Eq.\eqref{eq9_o} shows.
\end{enumerate}

The depth loss $\mathcal{L}_{d}$ and surface normal loss $\mathcal{L}_{n}$ are both defined using the Huber norm, which provides robust handling of outliers by balancing between the squared error for small deviations and the absolute error for larger deviations. Taking depth as an example, the per-pixel error $e(i,j)$ is computed as:

\begin{equation}
e(i,j) = \hat{\boldsymbol{D}}(i,j) - \boldsymbol{D}_{gt}(i,j),
\end{equation}
where $\hat{\boldsymbol{D}}(i,j)$ represents the predicted depth at pixel $(i,j)$, and $\boldsymbol{D}_{gt}(i,j)$ denotes the corresponding ground truth depth value. The Huber loss is then applied to this error:

\begin{equation}
\mathcal{L}_\delta(e(i,j)) =
\begin{cases}
\frac{1}{2}e(i,j)^2 & \textit{if } |e(i,j)| \leq \delta, \\
\delta \cdot \left(|e(i,j)| - \frac{1}{2}\delta\right) & \textit{if } |e(i,j)| > \delta.
\end{cases} 
\end{equation}

For depth, the total loss is computed as the average over all pixels:
\begin{equation}
\mathcal{L}_{d}=\frac1N\sum_{i,j}\mathcal{L}_\delta(e(i,j)),
\end{equation}
where $N$ is the total number of pixels.

The final loss function incorporates all three components:

\begin{equation}
\mathcal{L}_{overall}=\lambda_d\mathcal{L}_d+\lambda_n\mathcal{L}_n + \lambda_c\mathcal{L}_c,
\end{equation}
where $\lambda_d$, $\lambda_n$, and $\lambda_c$ are weights balancing the contributions of the depth loss, normal loss, and consistency loss.







Our M\textsuperscript{3}Depth model involves training a unified predictor $\mathcal{N}_{d-n}$ to predict depth map $\hat{\boldsymbol{D}}$ and surface normal map $\hat{\boldsymbol{N}}$ from a given set of images $\boldsymbol{I} = \{(\boldsymbol{I}_L^i,\boldsymbol{I}_R^i),i\in[1,2,\dots,M]\}$, where $\boldsymbol{I}_L^i$ and $\boldsymbol{I}_R^i$ represent the left and right images of the $i$-th stereo pair, respectively, and $M$ denotes the total number of stereo pairs in the dataset.

The primary objective is to ensure that the predicted depth map $\hat{\boldsymbol{D}}$ aligns closely with the true depth map $\boldsymbol{D}_{gt}$. Formally, the end-to-end prediction process can be formulated as: 

\begin{equation}
\hat{\boldsymbol{D}},\hat{\boldsymbol{N}}=\mathcal{N}_{d\text{-}n}(\boldsymbol{I},\boldsymbol{\theta}),
\end{equation}
where $\boldsymbol{\theta}$ represents the network parameters. The training objective is to identify the optimal parameters $\boldsymbol{\theta}^*$ which minimize the overall loss function $\mathcal{L}_{overall}$, defined as:
\begin{equation}
\boldsymbol{\theta}^* = \arg\min_{\boldsymbol{\theta}} \left\{\mathcal{L}_{overall}(\mathcal{N}_{d\text{-}n}(\boldsymbol{I};\boldsymbol{\theta}),\boldsymbol{D}_{gt},\boldsymbol{N}_{gt})\right\}.
\end{equation}
\subsection{Experimental Implementation}
\subsubsection{Datasets}
Our study leverages the SimMars6K dataset \cite{ma2024automated} to train and evaluate our proposed model, which comprises 6325 pairs of simulated stereo images with depth labels, generated using the OAISYS simulator. This dataset is designed to closely replicate Martian terrain and aligns with the configurations of the Navigation and Terrain Camera (NaTeCam) on China’s Zhurong Rover. 

The stereo camera setup mirrors the real-world specifications of NaTeCam used on Zhurong Rover. It is mounted 1.2 meters above the ground surface, with a baseline of 270 mm, a focal length of 595.90 pixels, and a field of view (FoV) measuring 46.5$^{\circ}$ $\times$ 46.5$^{\circ}$. The captured images are all with a resolution of 512 $ \times $ 512 pixels, ensuring consistency with the rover’s real-world imaging capabilities.

\begin{figure}[thb] \centering
    \includegraphics[width=0.45\textwidth]{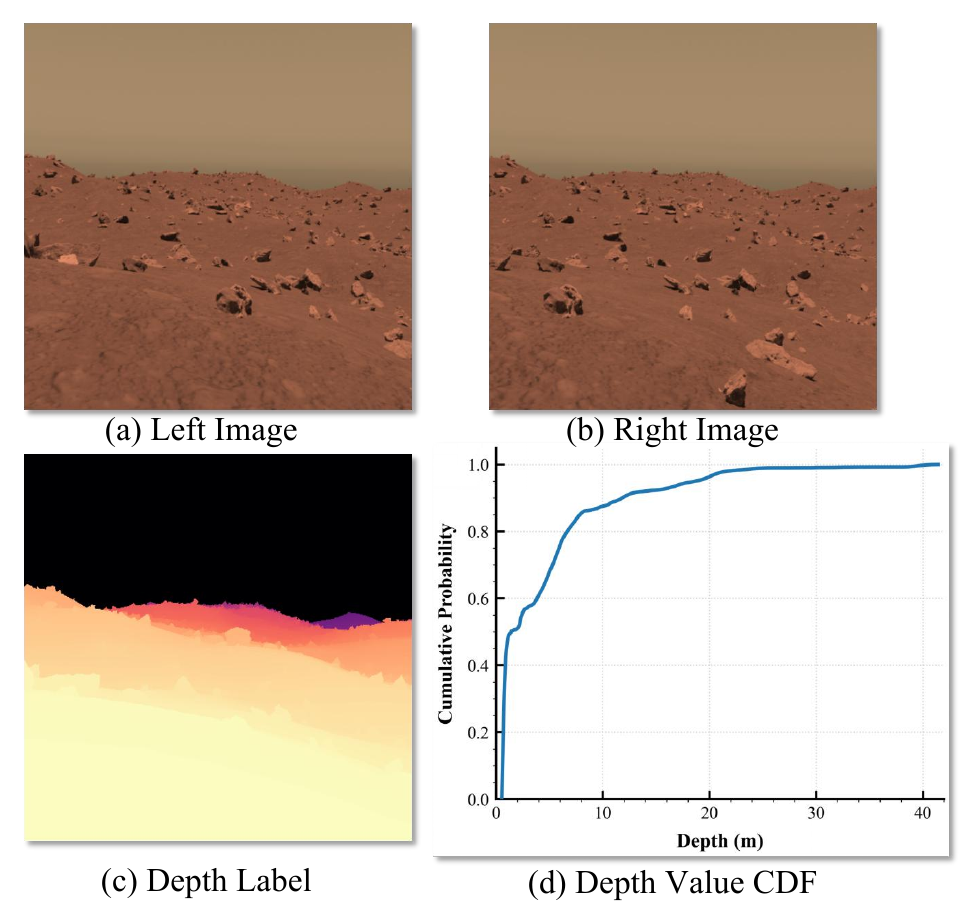}
    \caption{Visualization of dataset examples and depth distribution characteristics.} 
    \label{dataset}
\end{figure}

Following the guidelines in \cite{liu2024high}, the camera can capture a depth field ranging from 0.5 meters to infinity, with optimal focus at 1 meters. To ensure the practicality in real-world scenarios, only areas within a 15 meters capture radius are considered in this study, excluding regions beyond this range or at infinite depth \cite{zeng2022pan}. As shown in \autoref{dataset}, each stereo image pair consists of left and right perspective images. The left image is used as the reference image, and its depth ground truth is processed accordingly. The cumulative distribution function of the depth values is illustrated in \autoref{dataset}.(d), revealing that approximately 90\% of the pixel values are within a range of 15 meters. Additionally, we follow these established methods described in \cite{kusupati2020normal,fouhey2013data} to generate ground truth surface normal for training  \footnote{The processed dataset, including depth and surface normal labels, is publicly available at \url{https://zenodo.org/records/14286916}.}.


\subsubsection{Comparison Models}
To verify the advantages of our proposed depth estimation model, we use following SOTA methods for comparison:
\begin{itemize}
    \item Raft-Stereo \cite{lipson2021raft}: employs multi-level convolutional GRUs to iteratively propagate and aggregate contextual information across the image, facilitating precise stereo correspondence estimation.
    \item HitNet-Stereo \cite{tankovich2021hitnet}: replaces traditional cost volume construction with a multi-resolution initialization strategy. It employs differentiable 2D geometric propagation and warping mechanisms, while leveraging slanted plane hypotheses for accurate disparity estimation.
    \item CRE-Stereo \cite{li2022practical}: employs a hierarchical coarse-to-fine disparity refinement with a cascaded recurrent architecture and an adaptive group correlation layer to address rectification errors, enabling robust stereo matching.
    \item UniMVSNet-Stereo \cite{peng2022rethinking}: introduces a unified depth representation combining regression and classification advantages, along with a unified focal loss to address sample imbalance, enabling fine-grained, sub-pixel depth estimation without additional computational overhead.
    \item Selective-Stereo \cite{wang2024selective}: utilizes selective recurrent unit, an iterative update operator that adaptively fuses multi-frequency disparity information for edges and smooth regions using a contextual spatial attention module.
\end{itemize}
\subsubsection{Evaluation Metrics}

\begin{table*}[!th]
\centering
\caption{Quantitative Results Comparison on depth estimation among SOTA algorithms. All results are evaluated on SimMars6K dataset. The best results are marked in \textbf{Bold}.}
\label{results_comparison}
\begin{tabular}{c||cccc||ccc}
\toprule
\multirow{2}{*}{\textbf{Method}} & \multicolumn{4}{c||}{\textbf{Depth Estimation Error}} & \multicolumn{3}{c}{\textbf{Depth Estimation Accuracy}} \\ 
& {Abs Rel $\downarrow$} & {Sq Rel $\downarrow$} & {RMSE $\downarrow$} & {log$_{10}$ $\downarrow$} & $\delta_1 < 1.25$ $\uparrow$ & $\delta_2 < 1.25^2$ $\uparrow$ & $\delta_3 < 1.25^3$ $\uparrow$ \\
\midrule
Raft-Stereo \cite{lipson2021raft}  & 0.136 & 0.103 & 0.581 & 0.061 & 0.814 & 0.953 & 0.988  \\
HitNet-Stereo \cite{tankovich2021hitnet} & 0.128 & 0.085 & 0.562 & 0.057 & 0.839 & 0.966 & 0.992 \\
CRE-Stereo \cite{li2022practical} & 0.103 & 0.063 & 0.367 & 0.047 & 0.887 & 0.976 & 0.996 \\
UniMVSNet-Stereo \cite{peng2022rethinking} & 0.148 & 0.126 & 0.653 & 0.063 & 0.783 & 0.948 & 0.981 \\ 
Selective-Stereo \cite{wang2024selective} & 0.116 & 0.074 & 0.416 & 0.051 & 0.856 & 0.973 & 0.994 \\
\rowcolor{gray!25}
\textbf{M\textsuperscript{3}Depth (Ours)}  & \textbf{0.089} & \textbf{0.058} & \textbf{0.314} & \textbf{0.038} & \textbf{0.905} & \textbf{0.991} & \textbf{0.998} \\
\bottomrule
\end{tabular}
\label{comparsion}
\end{table*}

To quantitatively evaluate the
performance of the proposed framework and comparative approaches from various aspects, our study evaluates the models using widely adopted metrics in depth estimation \cite{survey_Laga}. The commonly used evaluation metrics are defined as follows:
\begin{itemize}
    \item Absolute Relative Error (Abs Rel): computes the mean of relative errors between the predicted depth and the ground truth depth:
    \begin{equation}
       \text{Abs Rel} = \frac{1}{N} \sum_{j}^{N} \frac{\left|\hat{d}_{j}-d_{j}\right|}{d_j}.
    \end{equation}
    \item Squared Relative Error (Sq Rel): defined as the average squared error ($L_2$ distance) between the ground-truth depth and the predicted depth, normalized by the ground-truth depth value for each pixel. Sq Rel penalizes larger errors more than Abs Rel, which makes it more sensitive to large discrepancies:
    \begin{equation}
    \text{Sq Rel} =\frac{1}{N}\sum_{j}^{N}\frac{\left|\hat{d}_{j}-d_{j}\right|^2}{d_{j}}.
    \end{equation}
    \item Root Mean Squared Error (RMSE): measures the square root of the mean squared differences between predicted and ground truth depths:
    \begin{equation}
    \text{RMSE} = \sqrt{\frac{1}{N} \sum_{j}^{N} \left|\hat{d}_{j}-d_{j}\right|^2},
    \end{equation}
    \item log$_{10}$: computes the mean of the logarithmic differences between the predicted depth and the ground truth depth. The log$_{10}$ metric is used to reduces the impact of large disparities and is less sensitive to large-scale errors in depth:
    \begin{equation}
    \text{log}_{10} = \frac{1}{N} \sum_{j=1}^{N} \left| \log_{10} \hat{d}_{j} - \log_{10} d_{j} \right|.
    \end{equation}
    \item Threshold Accuracy ($\delta_i$): calculates the percentage of pixels that satisfy the following condition:
    \begin{equation}
    \max \left(\frac{d_{j}}{\hat{d}_{j}}, \frac{\hat{d}_{j}}{d_{j}}\right)=\delta < \delta_i = 1.25^i, i \in [1,2,3].
    \end{equation}

\end{itemize}

All above, $\hat{d}_{j}$ is the predicted depth of pixel $j$, ${d}_{j}$ is the ground truth of pixel $j$ in depth label map, and $N$ is the total number of pixels.

\subsubsection{Training Settings}

All our models are implemented using the PyTorch framework and trained on Ubuntu 18.04 system equipped with 4 NVIDIA RTX 4090 GPUs. The dataset is split into training, validation, and testing subsets with a ratio of 7:2:1. The training epochs is uniformly set to 100. The batch size is set to 12. The Adam optimizer is used with an initial learning rate of $1 \times 10^{-4}$ by default. In our network, the trainable parameters are initialized with a normal distribution. The ground-truth depth values of all the depth images are normalized to the range $\left[0,1\right]$, which is a common preprocessing step in depth estimation tasks. The hyper-parameters, loss function weights $\lambda_d$, $\lambda_n$ and $\lambda_c$ are set to 2, 1 and 3 respectively. Unless otherwise stated, our experiments use 5 iterations in the IRM, as we observed this provides a balance between accuracy and efficiency. The number of decomposition levels in the WEFE is set to 3 by default, based on the analysis in \autoref{appendix}.

\subsection{Experimental Results}

\subsubsection{Quantitative Results}

To compare on the SimMars6K datasets, we choose several state-of-the-art (SOTA) approaches of a diverse kind. These include iterative optimization-based methods like Raft-Stereo \cite{lipson2021raft}, hierarchical and geometric propagation approaches such as HitNet-Stereo \cite{tankovich2021hitnet}, coarse-to-fine refinement strategies exemplified by CRE-Stereo \cite{li2022practical}, unified depth representation models like UniMVSNet-Stereo \cite{peng2022rethinking}, and frequency domain-based designs such as Selective-Stereo \cite{wang2024selective}. The complete comparison on all the metrics is presented in \autoref{results_comparison}, where Metric with $\uparrow$ indicates that higher value is preferred, and $\downarrow$ denotes the opposite. 

\begin{figure}[thb]
    \centering
    \includegraphics[width=0.46\textwidth]{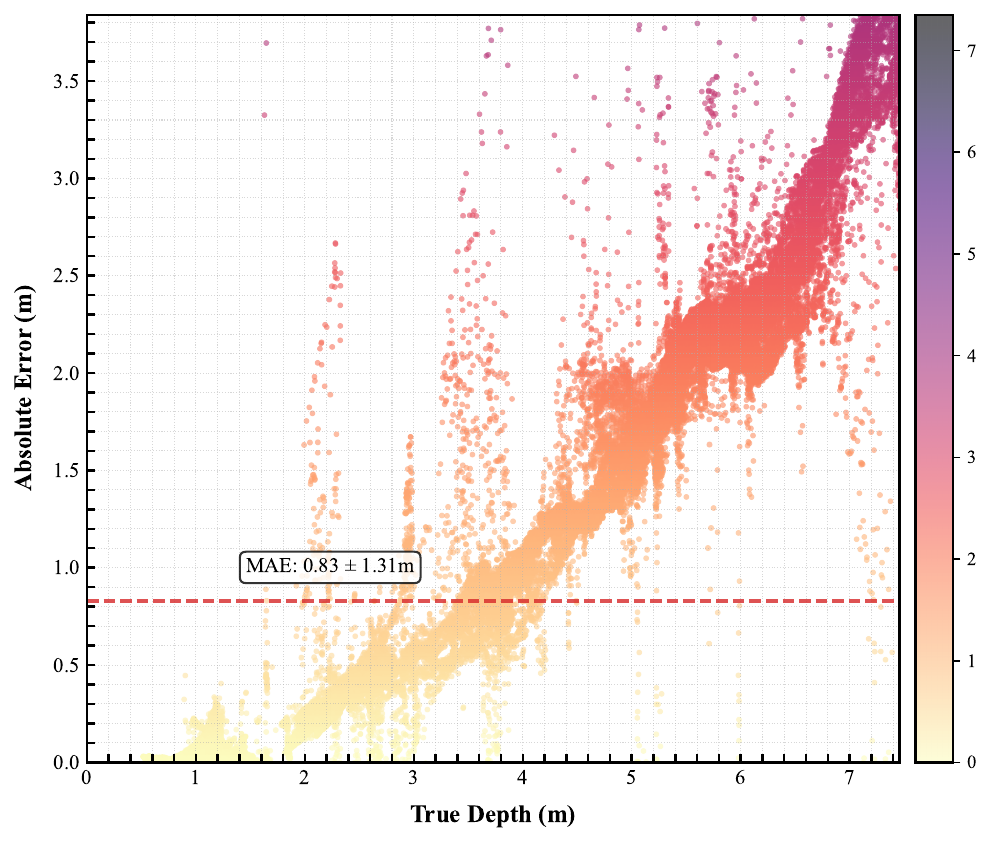}\\
    \makebox[0.46\textwidth]{\footnotesize \rmfamily (a) Mean absolute error across depth ground truth}
    \vspace{0.2cm}
    \includegraphics[width=0.46\textwidth]{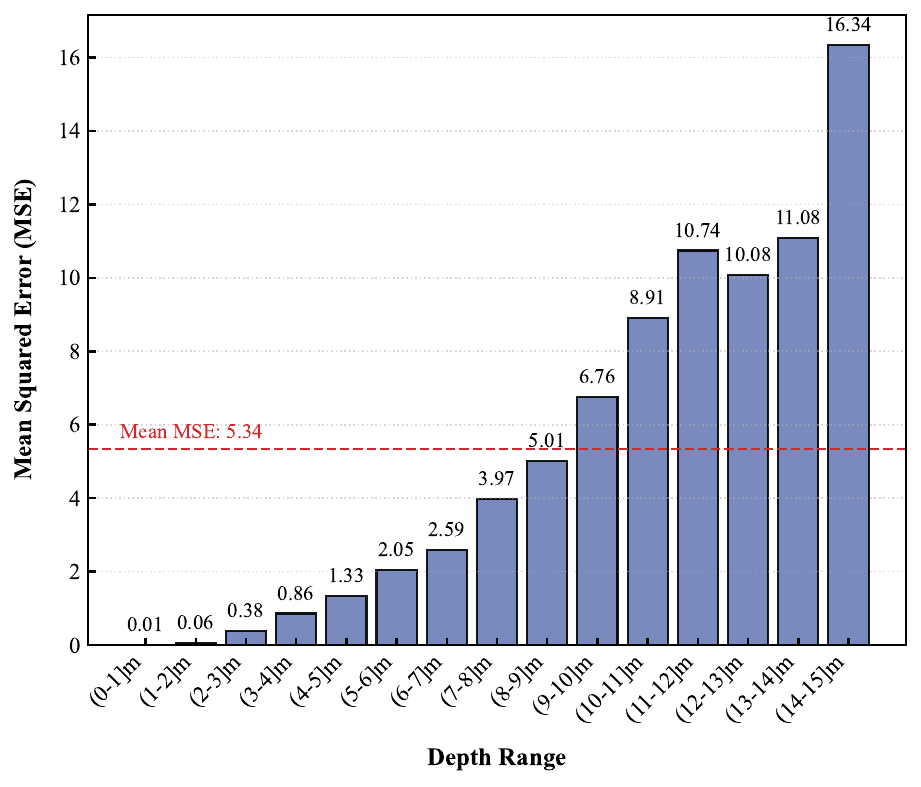}\\
    \makebox[0.46\textwidth]{\footnotesize \rmfamily (b) Mean squared error across depth interval}
    
    \caption{Comparative analysis of depth estimation performance across different depth values.}
    \label{depth_error}
\end{figure}

As summarized in \autoref{results_comparison}, our proposed method achieves the best performance across all evaluated metrics, demonstrating its effectiveness in handling the unique challenges of Mars terrains. Specifically, our method achieves Abs Rel of 0.089 and RMSE of 0.314, representing improvements of 13.6\% and 14.5\%, respectively, compared to the next-best method (CRE-Stereo). Additionally, under $\delta_i$ thresholds, our model achieves the highest accuracy across all levels.

\begin{table}[thb]
\centering
\caption{Metrics for Surface Normal Evaluation with Formulations.}
\label{normal_metrics}
\resizebox{0.48\textwidth}{!}{
\begin{tabular}{ll}
\toprule
\textbf{Metric} & \textbf{Formulation} \\
\midrule
\vspace{0.2em}
Mean & $\frac{1}{N} \sum_{i=1}^N \boldsymbol{\theta}_{i}$, where $\boldsymbol{\theta}_{i}$ is the angular error. \\
\vspace{0.2em}
Median & $\text{Median}(\boldsymbol{\theta}_{1}, \boldsymbol{\theta}_{2}, \dots, \boldsymbol{\theta}_{N})$. \\
\vspace{0.2em}
$\text{RMSE}_n$ & $\sqrt{\frac{1}{N} \sum_{i=1}^N \boldsymbol{\theta}_{i}^2}$. \\
\vspace{0.2em}
$11.25^\circ$ & $\frac{1}{N} \sum_{i=1}^N \boldsymbol{1}(\boldsymbol{\theta}_{i} < 11.25^\circ)$. \\
\vspace{0.2em}
$22.5^\circ$ & $\frac{1}{N} \sum_{i=1}^N \boldsymbol{1}(\boldsymbol{\theta}_{i} < 22.5^\circ)$. \\
\vspace{0.2em}
$30^\circ$ & $\frac{1}{N} \sum_{i=1}^N \boldsymbol{1}(\boldsymbol{\theta}_{i} < 30^\circ)$. \\
\bottomrule
\end{tabular}
}
\end{table}

In comparison, UniMVSNet-Stereo demonstrates lower performance due to its reliance on a unified depth representation that combines regression and classification. It is effective for general tasks, yet struggles with fine-grained depth estimation in complex Mars terrains. Similarly, Raft-Stereo and HitNet-Stereo achieve moderate performance due to their iterative refinement and geometric propagation strategies. However, their reliance on traditional matching mechanisms limits their adaptability in texture-less regions, leading to suboptimal results. Selective-Stereo and CRE-Stereo exhibit better performance, with CRE-Stereo particularly excelling due to its hierarchical coarse-to-fine refinement and adaptive group correlation layer, which effectively addresses rectification errors and improves matching robustness. Selective-Stereo leverages frequency domain analysis and places emphasis on high-frequency signals, allowing it to represent texture-rich edges and detailed boundaries with greater clarity. However, these methods still fall short in highly challenging regions where fine-grained precision is required.

To further investigate the relationship between depth estimation error and true depth values, we conduct a depth-wise error analysis as shown in \autoref{depth_error}. The \autoref{depth_error}(a) presents a point-wise distribution of absolute error against the ground truth depth for all pixels. The color gradient represents the magnitude of squared error, with warmer colors indicating lower error regions. We observe that the absolute error tends to increase with depth, which is consistent with the common trend in stereo-based methods where disparity resolution decreases with increasing depth. The overall mean absolute error (MAE) is 0.83 $\pm$ 1.31 meters. The \autoref{depth_error}(b) plot provides a binned MSE analysis across 15 depth intervals. It shows that error remains relatively low and stable for near-field depth ranges (\textit{e.g.}, (0–4m]), but increases in far-field regions, peaking at 16.34 for the (14–15m] range.

Surface normal plays a critical role in our framework, serving as an auxiliary geometric constraint to enhance depth estimation accuracy. By leveraging the intrinsic relationship between surface normals and depth gradients, our method incorporates normal predictions to refine depth estimation. Given the importance of surface normals in our approach, it is essential to ensure the accuracy and reliability of the predicted normal maps, as errors in normal predictions could directly affect the depth estimation process. To evaluate the quality of surface normal predictions, we adopt a set of standard metrics commonly used in surface normal estimation tasks, as summarized in \autoref{normal_metrics}.
In this context, $N$ represents the total number of evaluated pixels, and $\boldsymbol{\theta}_{i}$ is the angular error between the predicted normal vector and the ground-truth normal vector at pixel $i$. This angular error is computed as:

\begin{equation}\boldsymbol{\theta}_{i}=\arccos\left(\frac{\hat{\boldsymbol{n}}_{i}\cdot\boldsymbol{n}_{i}}{\|\hat{\boldsymbol{n}}_{i}\|\|\boldsymbol{n}_{i}\|}\right),\end{equation}
where $\hat{\boldsymbol{n}}_i$ is the predicted normal vector, and $\boldsymbol{n}_i$ is the ground-truth normal vector. The indicator function $\boldsymbol{1}(\cdot)$ outputs 1 if the condition is true and 0 otherwise.


\autoref{normal_pre} presents the quantitative evaluation of surface normal prediction across multiple methods. Notably, Fusion \cite{zeng2019deep} achieves the best results across most metrics due to its reliance on origin depth information. While our approach is not specifically tailored for surface normal estimation, it still achieves competitive performance, with a mean angular error of $20.6^\circ$ and a median angular error of $12.3^\circ$. These results confirm that the predicted surface normals are sufficiently accurate to serve as a reliable auxiliary modality for improving depth estimation. The performance also demonstrates the robustness of our framework in extracting geometric features without requiring explicit ground truth depth or specific optimization for surface normal estimation.


\begin{table}[thb]\centering
    \caption{Quantitative evaluation of Surface Normal Prediction Results.}
    \label{normal_pre}
    \resizebox{0.48\textwidth}{!}{
    \large
    \begin{tabular}{c||ccc||ccc}
    \toprule
      \multirow{2}{*}{\textbf{Method}}  & \multicolumn{3}{c||}{\textbf{Normal Error}} & \multicolumn{3}{c}{\textbf{Normal Accuracy}} \\ 
       & Mean $\downarrow$ & Median $\downarrow$ & $\text{RMSE}_n$ $\downarrow$ & $11.25^{\circ}$ $\uparrow$ &  $22.5^{\circ}$ $\uparrow$ & $30^{\circ}$ $\uparrow$ \\
        \midrule
        Fusion \cite{zeng2019deep} & 15.7 & 10.59 & 30.4 & 0.513 & 0.672 & 0.798 \\
        GeoNet \cite{qi2018geonet} & 23.1 & 15.8 & 38.9 & 0.414 & 0.611 & 0.736 \\
        \rowcolor{gray!25}
        \textbf{M\textsuperscript{3}Depth} & 20.6 & 12.3 & 34.2 & 0.445 & 0.638 & 0.759\\
        \bottomrule
    \end{tabular}
    }
\end{table}



\subsubsection{Qualitative Results}
In addition to the quantitative comparison, we also perform qualitative depth comparisons to further validate the effectiveness of our proposed method. The qualitative analysis is conducted using the SimMars6K dataset, which provides a controlled environment for visualizing depth estimation performance under Martian-like conditions. Moreover, to assess the generalization ability of different methods in real-world scenarios, we extend the evaluation to real Mars imagery captured by Zhurong Rover.

\begin{figure*}[t] \centering
\includegraphics[width=0.137\textwidth]{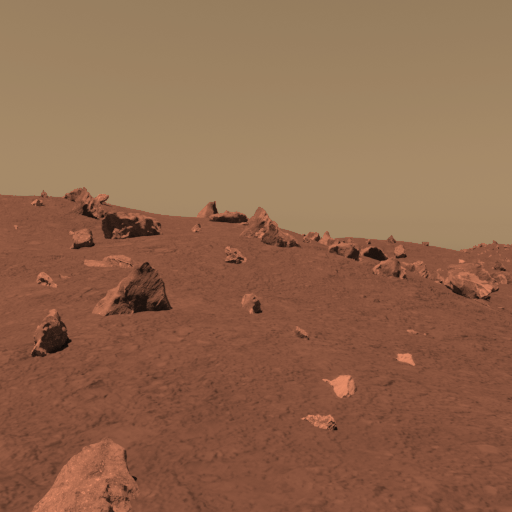}
    \includegraphics[width=0.137\textwidth]{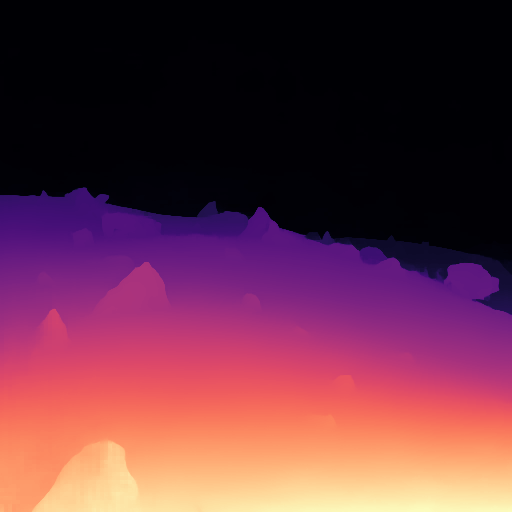}
    \includegraphics[width=0.137\textwidth]{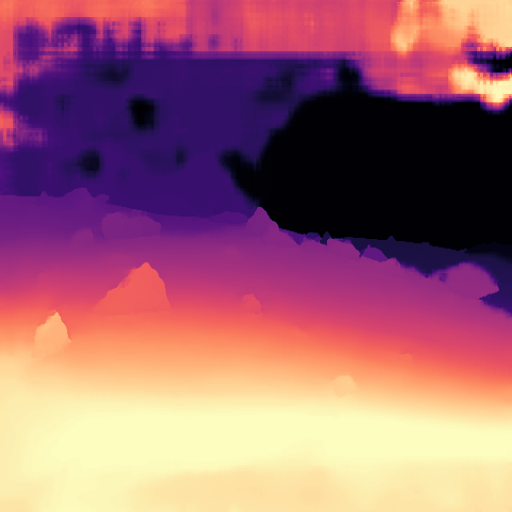}
    \includegraphics[width=0.137\textwidth]{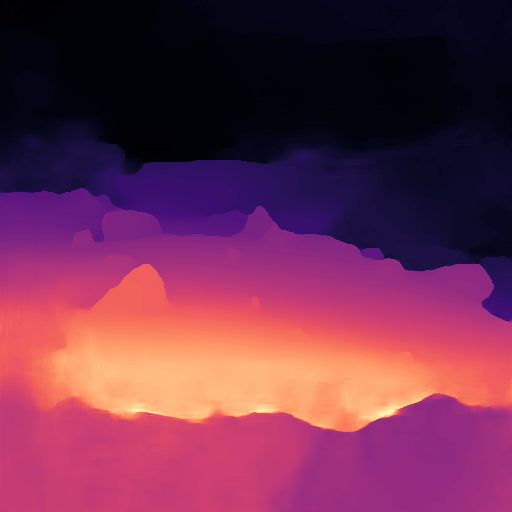}
    \includegraphics[width=0.137\textwidth]{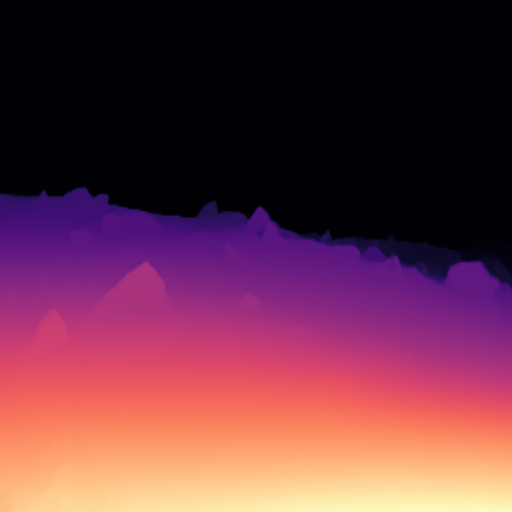}
    \includegraphics[width=0.137\textwidth]{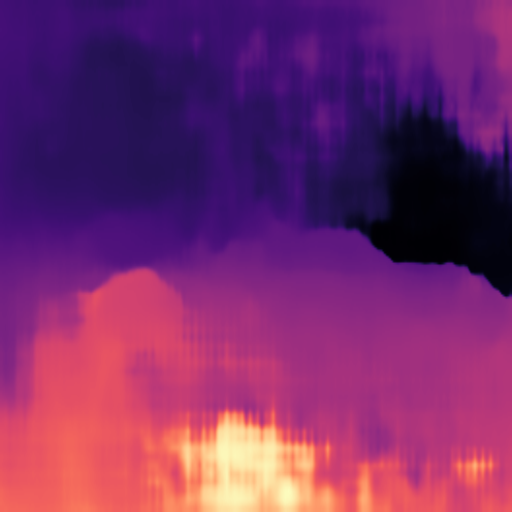}
    \includegraphics[width=0.137\textwidth]{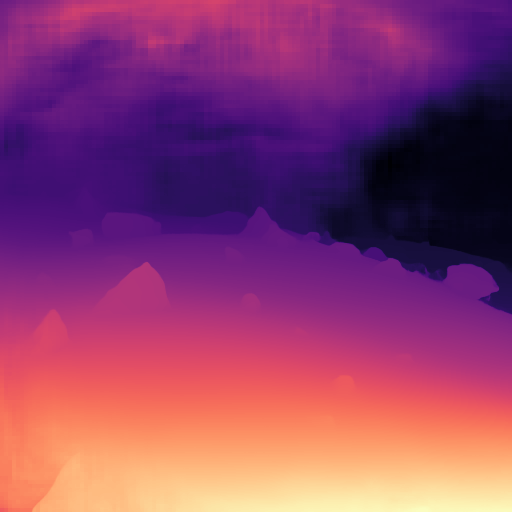}
    \\
    \vspace{0.2em}
    \includegraphics[width=0.137\textwidth]{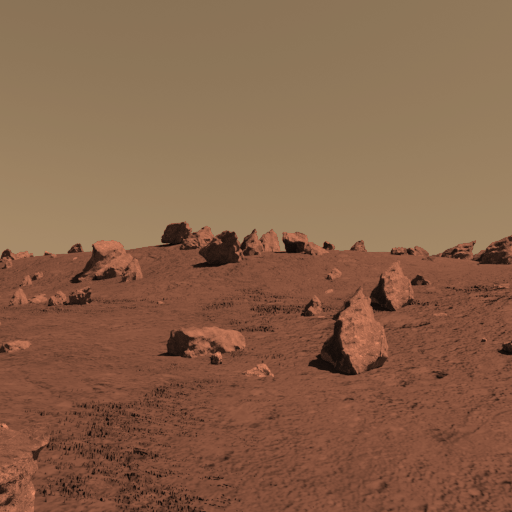}
    \includegraphics[width=0.137\textwidth]{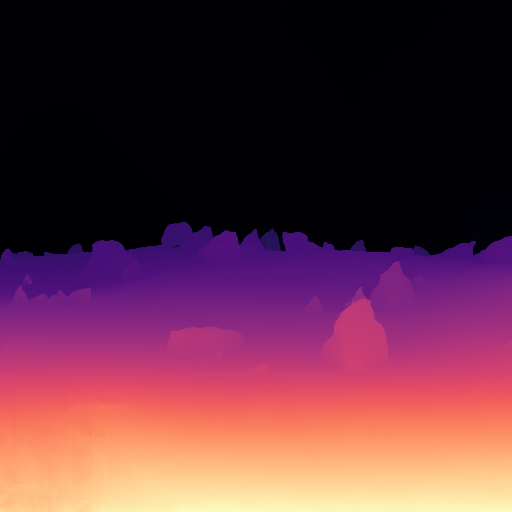}
    \includegraphics[width=0.137\textwidth]{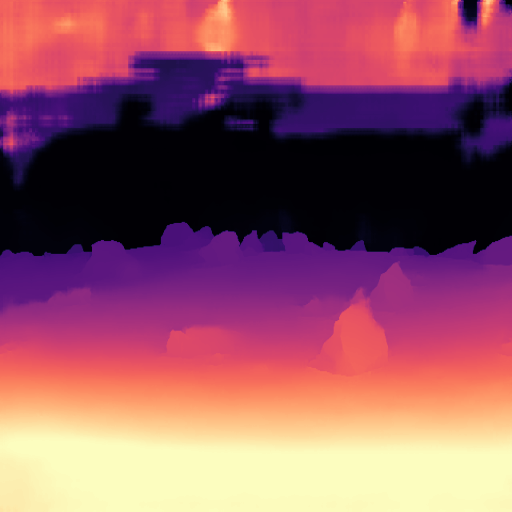}
    \includegraphics[width=0.137\textwidth]{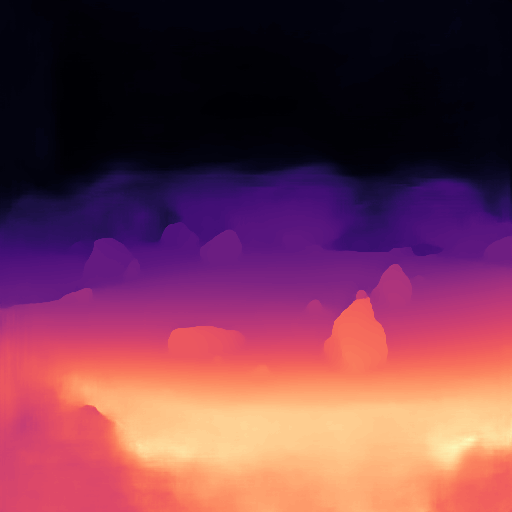}
    \includegraphics[width=0.137\textwidth]{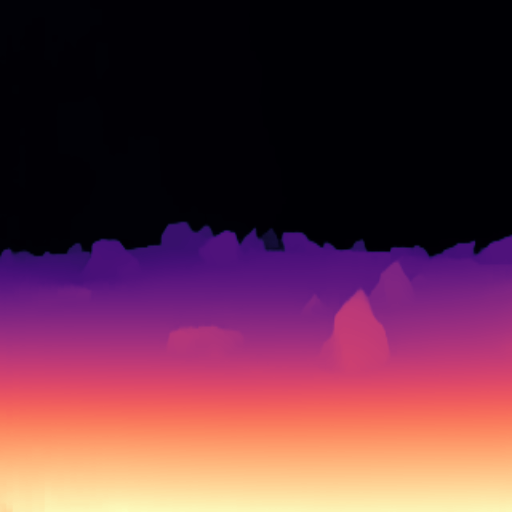}
    \includegraphics[width=0.137\textwidth]{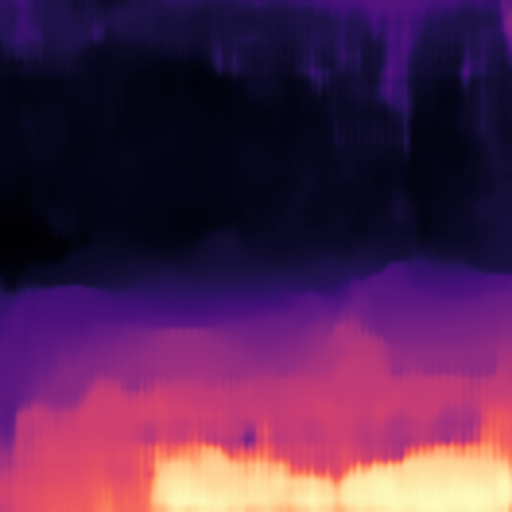}
    \includegraphics[width=0.137\textwidth]{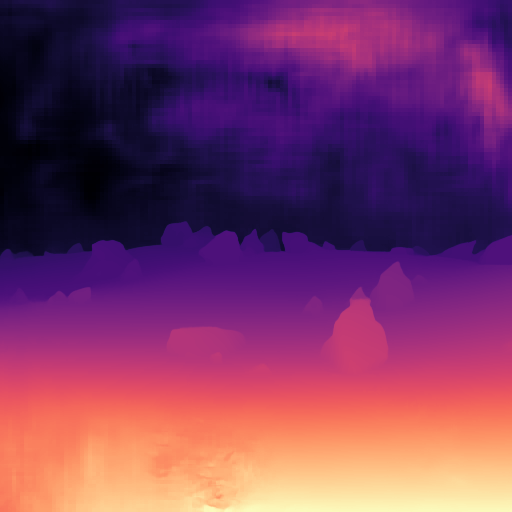}
    \\
    \vspace{0.2em}
    \includegraphics[width=0.137\textwidth]{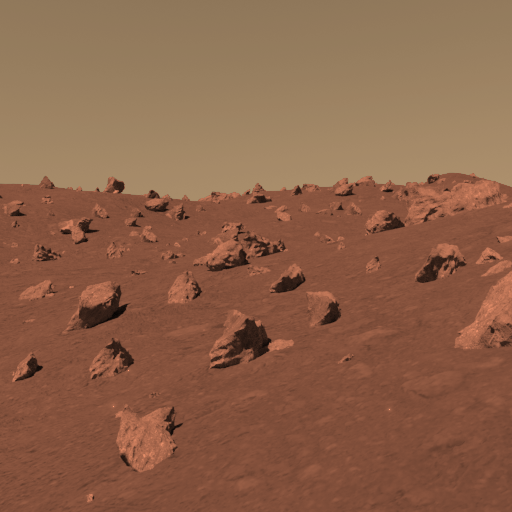}
    \includegraphics[width=0.137\textwidth]{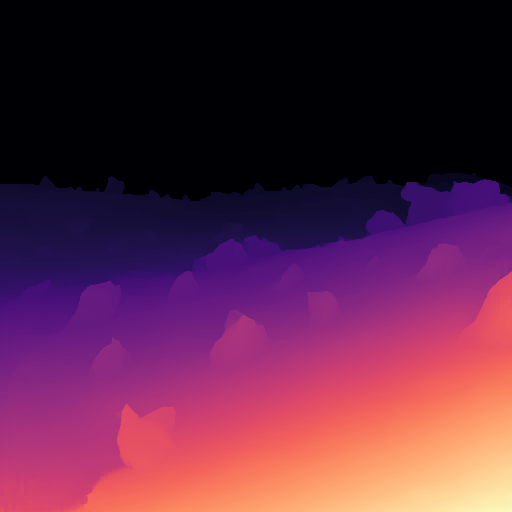}
    \includegraphics[width=0.137\textwidth]{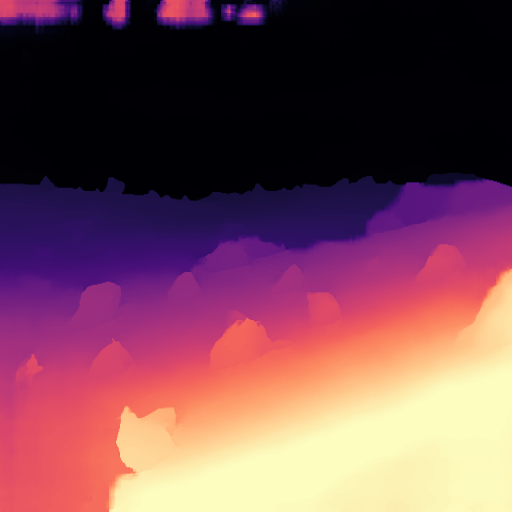}
    \includegraphics[width=0.137\textwidth]{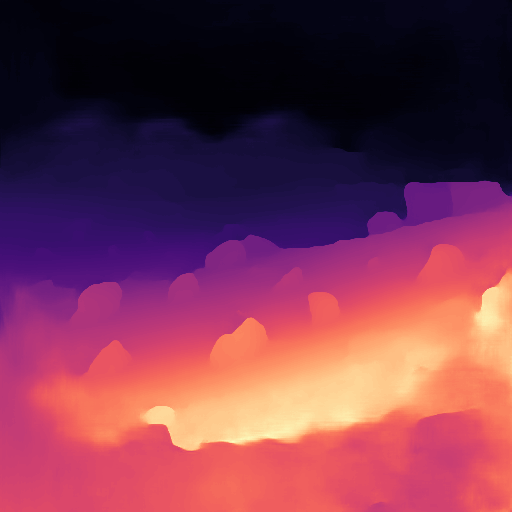}
    \includegraphics[width=0.137\textwidth]{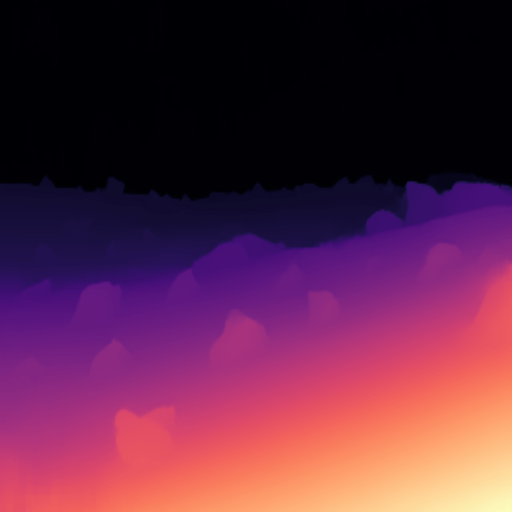}
    \includegraphics[width=0.137\textwidth]{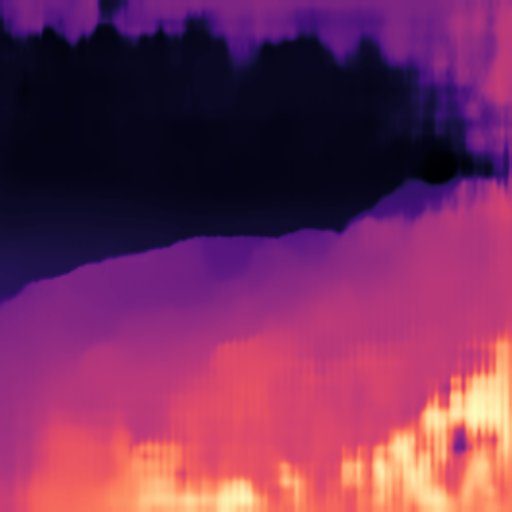}
    \includegraphics[width=0.137\textwidth]{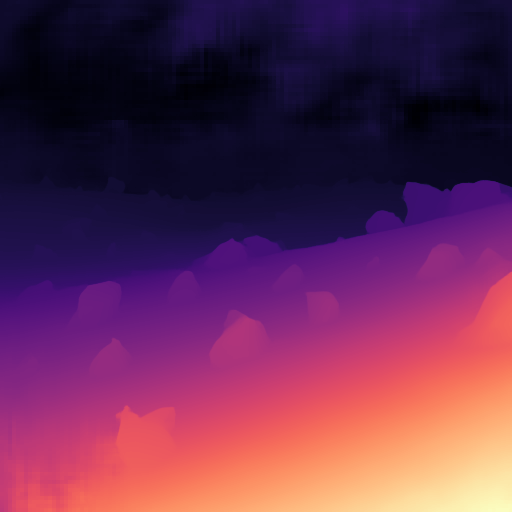}
    \\
    \vspace{0.2em}
    \includegraphics[width=0.137\textwidth]{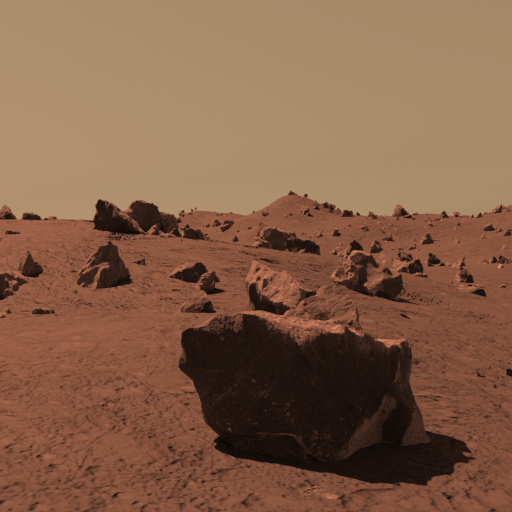}
    \includegraphics[width=0.137\textwidth]{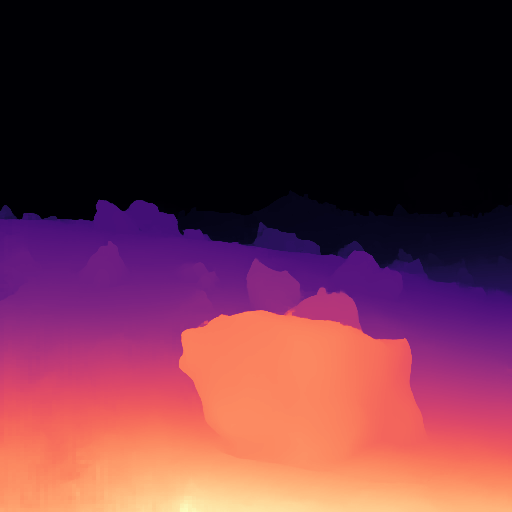}
    \includegraphics[width=0.137\textwidth]{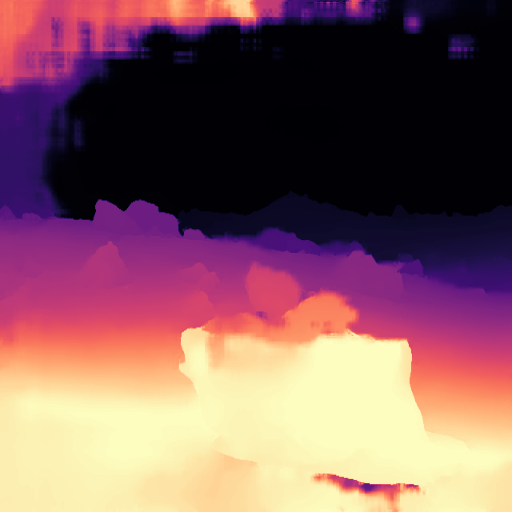}
    \includegraphics[width=0.137\textwidth]{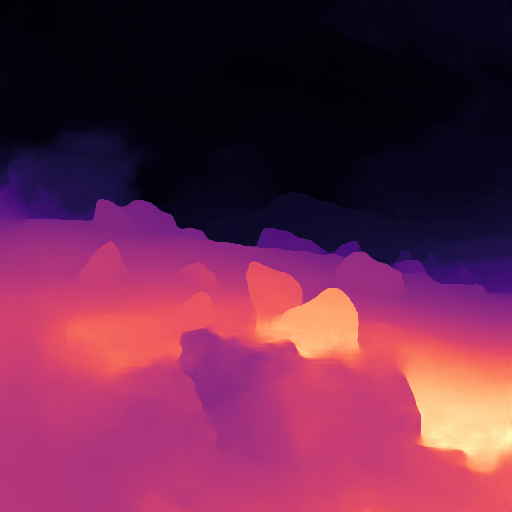}
    \includegraphics[width=0.137\textwidth]{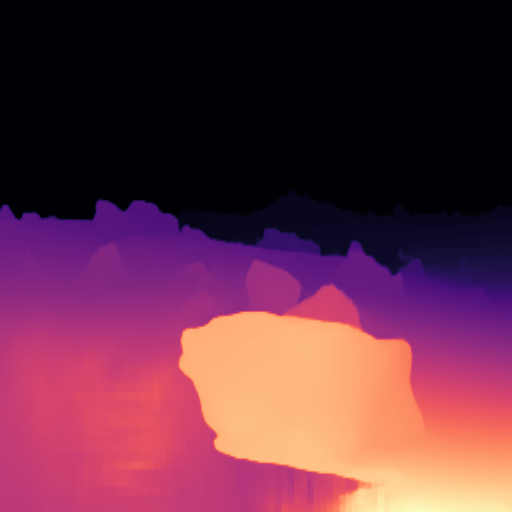}
    \includegraphics[width=0.137\textwidth]{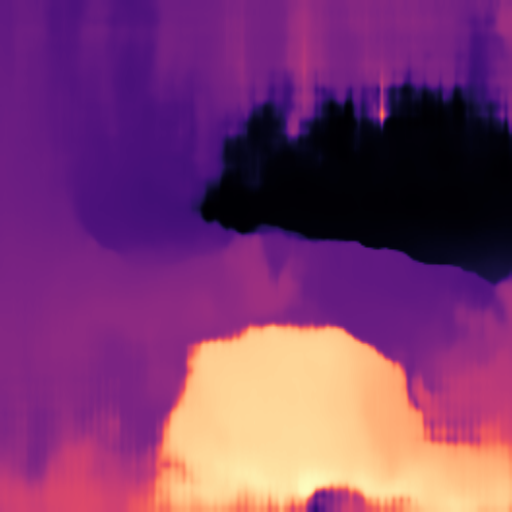}
    \includegraphics[width=0.137\textwidth]{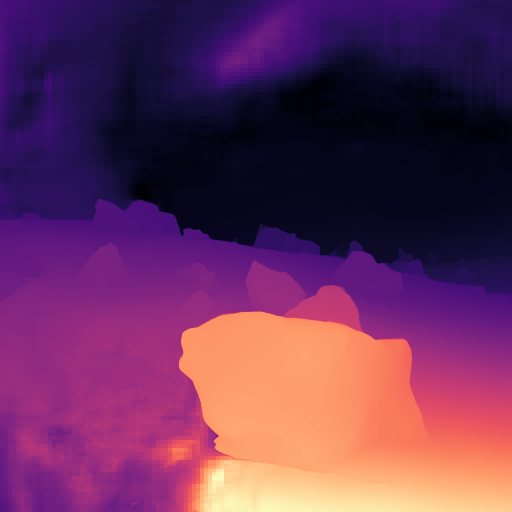}
    \\
    \makebox[0.137\textwidth]{\footnotesize Origin}
    \makebox[0.137\textwidth]{\footnotesize \textbf{M\textsuperscript{3}Depth (Ours)}}
    \makebox[0.137\textwidth]{\footnotesize Raft-Stereo}
    \makebox[0.137\textwidth]{\footnotesize HitNet-Stereo}
    \makebox[0.137\textwidth]{\footnotesize CRE-Stereo}
    \makebox[0.137\textwidth]{\footnotesize UniMVSNet-Stereo}
    \makebox[0.137\textwidth]{\footnotesize Selective-Stereo}
    \\
    \vspace{0.2em}
    \caption{Qualitative comparison of depth estimation results on Martian images. The first column shows the original reference images (left image), while the remaining columns compare the depth estimation outputs of our proposed M\textsuperscript{3}Depth model with other competitive baselines.} 
    \label{com_qualitative}
\end{figure*}

As illustrated in \autoref{com_qualitative}, our proposed M\textsuperscript{3}Depth method demonstrates superior performance compared to state-of-the-art approaches across a variety of challenging Martian scenes. Specifically, our method excels in delineating planar regions while preserving fine-grained local details, such as edges and boundaries around rocks and terrain structures. Compared to baseline methods such as Raft-Stereo and UniMVSNet-Stereo, which exhibit artifacts and noise in texture-less regions, our M\textsuperscript{3}Depth effectively eliminates inconsistencies induced by texture-less regions and maintains geometric continuity. It is more evident in large texture-less areas and thin structures (\textit{e.g.}, rock edges and smooth terrain surfaces), where competing methods tend to either over-smooth the depth or introduce estimation errors.

In addition, methods like HitNet-Stereo and Selective-Stereo display a degree of improvement in handling texture variations but still struggle with depth discontinuities, especially near object boundaries. On the other hand, our method achieves sharper and more accurate transitions, demonstrating its ability to balance global structure and local details. CRE-Stereo, while performing better for coarse-level depth reconstruction, struggles to capture fine details in areas with complex geometry or intricate surface patterns, leading to suboptimal performance in high-detail regions.

\begin{figure*}[htb] \centering
\includegraphics[width=0.9\textwidth]{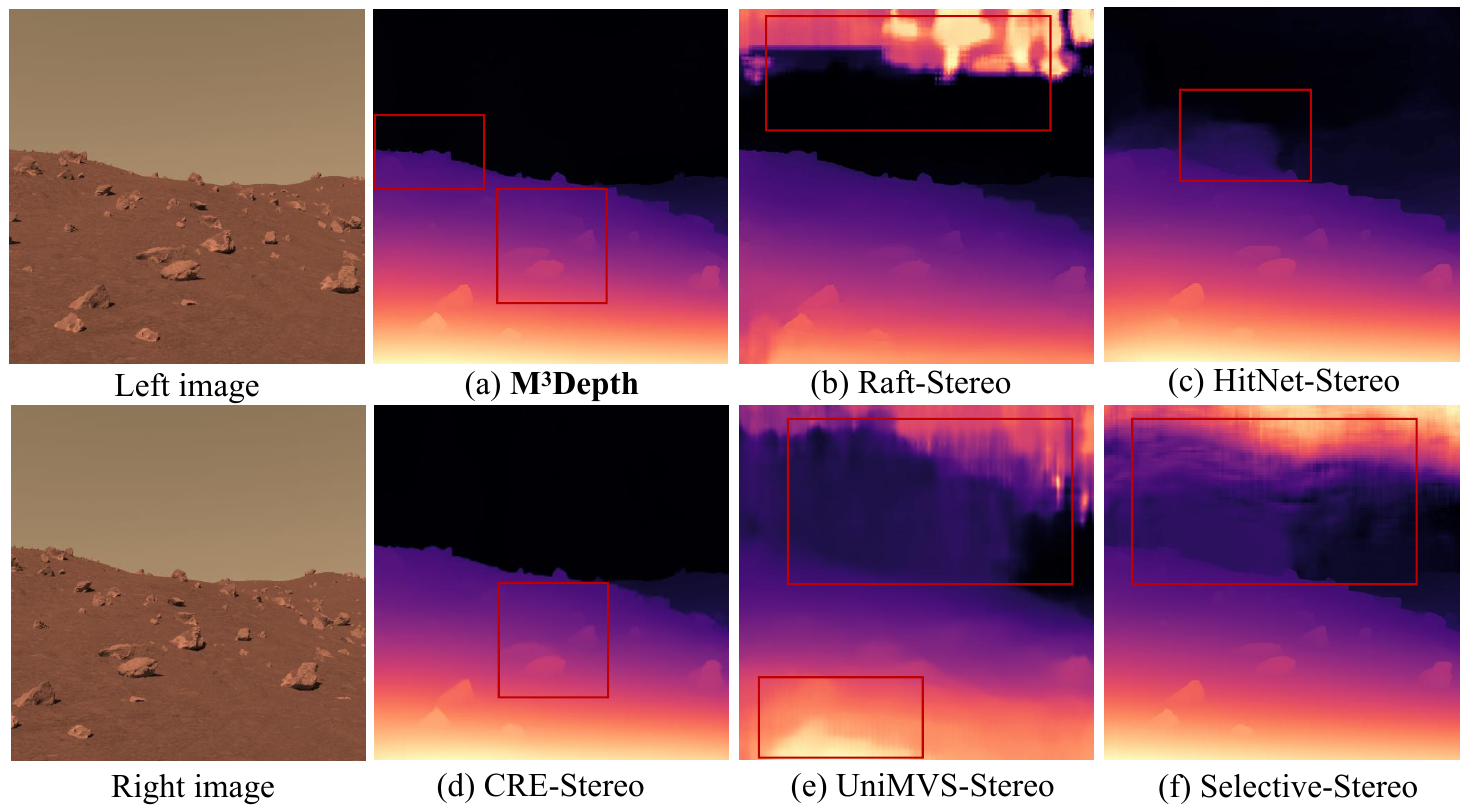}
    \caption{Qualitative depth estimation results on the SimMars6K dataset (Case-I). The red boxes highlight the regions to emphasize.} \label{zoom}
\end{figure*}

\begin{figure*}[!ht] \centering
\includegraphics[width=0.9\textwidth]{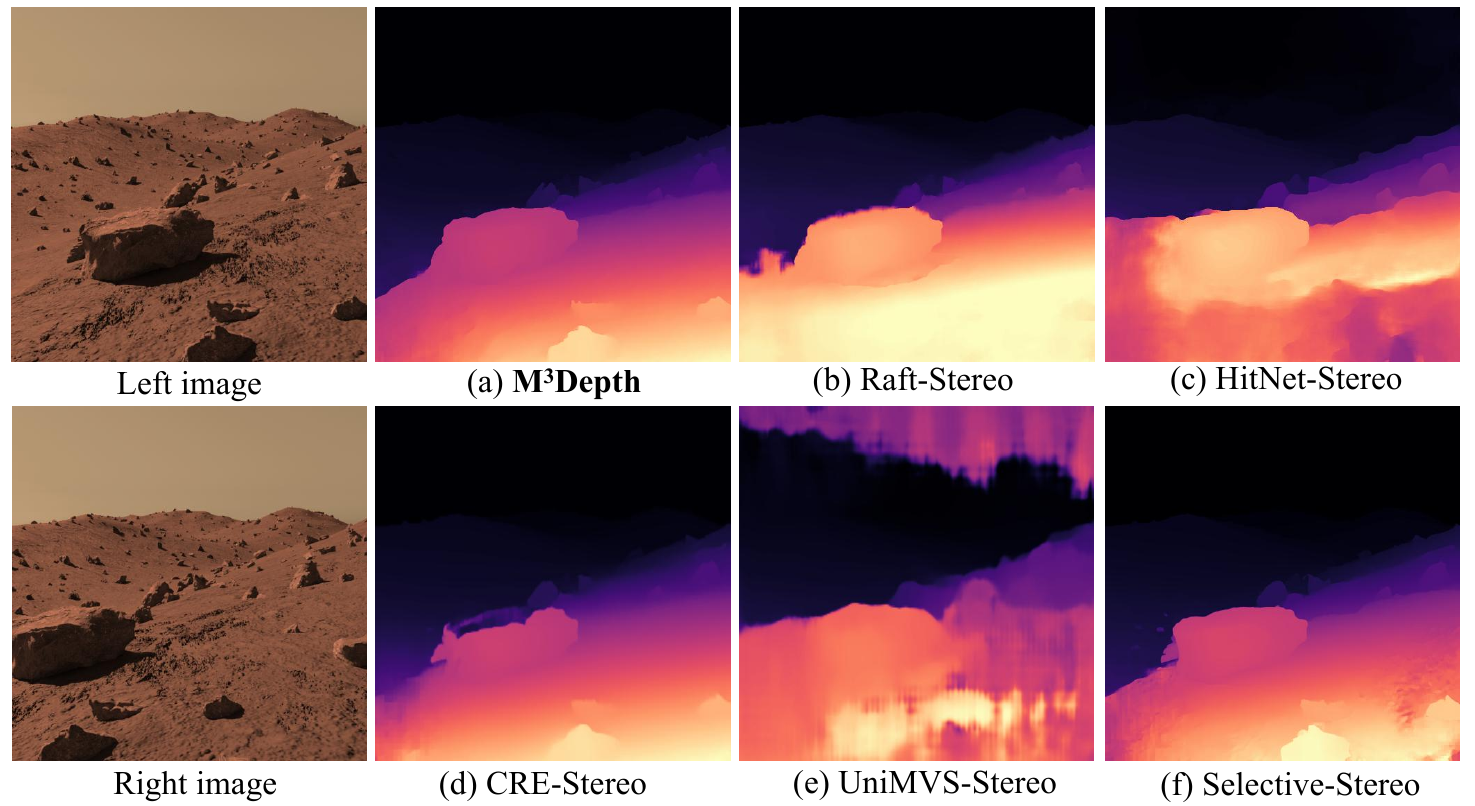}
    \caption{Qualitative depth estimation results on the SimMars6K dataset (Case-II).} \label{zoom2}
\end{figure*}

To further emphasize the strengths of our proposed M\textsuperscript{3}Depth method, we present a case study using an example from the SimMars6K dataset. As illustrated in \autoref{zoom}, specific regions of interest are highlighted with red boxes to focus on critical depth estimation challenges, including texture-less regions, edge preservation, and anti-artifact. Additionally, we provide other case studies in \autoref{zoom2}, which includes regions with complex shadows. This choice was made because the presence of shadows presents a more challenging environment for depth estimation, where several baseline methods show noticeable performance degradation due to the difficulties in accurately capturing depth in such regions.

\begin{itemize}
    \item \textit{Texture-less Regions.} In the flat terrain regions, methods like Raft-Stereo and UniMVS-Stereo exhibit noticeable noise and inconsistencies, with uneven or incorrect depth estimations. In contrast, M\textsuperscript{3}Depth produces smooth and coherent depth estimations, effectively handling the lack of texture by leveraging its wavelet-transformed convolutional kernels to capture low-frequency features.
    \item \textit{Edge Preservation.} In the regions with fine geometric details and boundaries, Raft-Stereo, HitNet-Stereo, and Selective-Stereo suffer from depth bleeding and blurring, failing to delineate object boundaries accurately. CRE-Stereo, while appearing to achieve better results in such regions, struggles with sharp object edges and fine transitions. It often exhibits slight over-smoothing or loss of precision, particularly where the terrain meets the horizon or along complex geometric structures. M\textsuperscript{3}Depth, on the other hand, accurately preserves boundary integrity and generates clear depth transitions, which can be attributed to the mutual boosting mechanism between depth and surface normal predictions.
    \item \textit{Anti-Artifacts.} UniMVS-Stereo introduce more artifacts and structural distortions, especially in areas with limited visual cues. These artifacts disrupt the geometric continuity of the predicted depth maps. Our method effectively suppresses such artifacts, producing consistent and realistic depth estimations across the scene.
\end{itemize}

\begin{figure*}[t] \centering
    
    \includegraphics[width=0.137\textwidth]{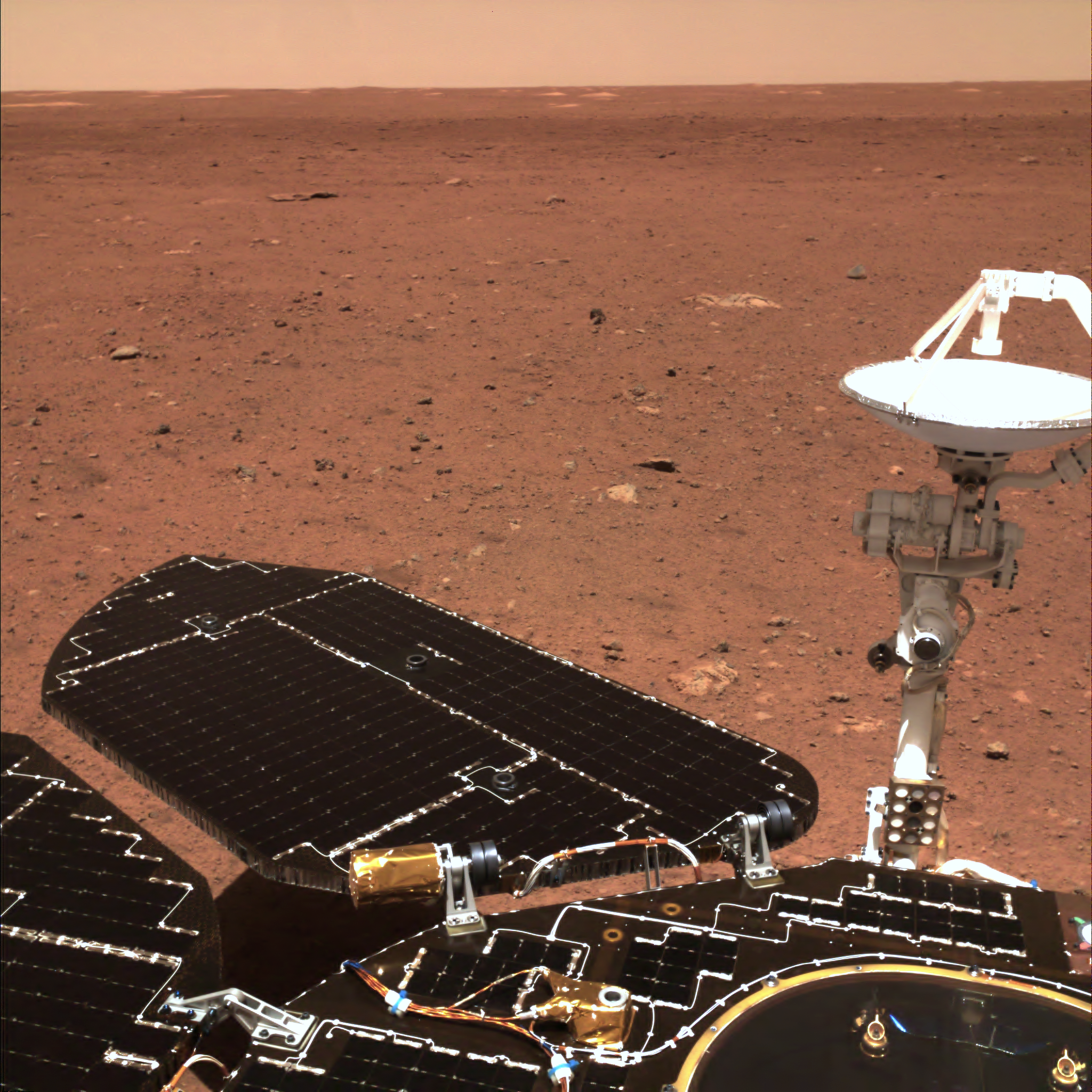}
    \includegraphics[width=0.137\textwidth]{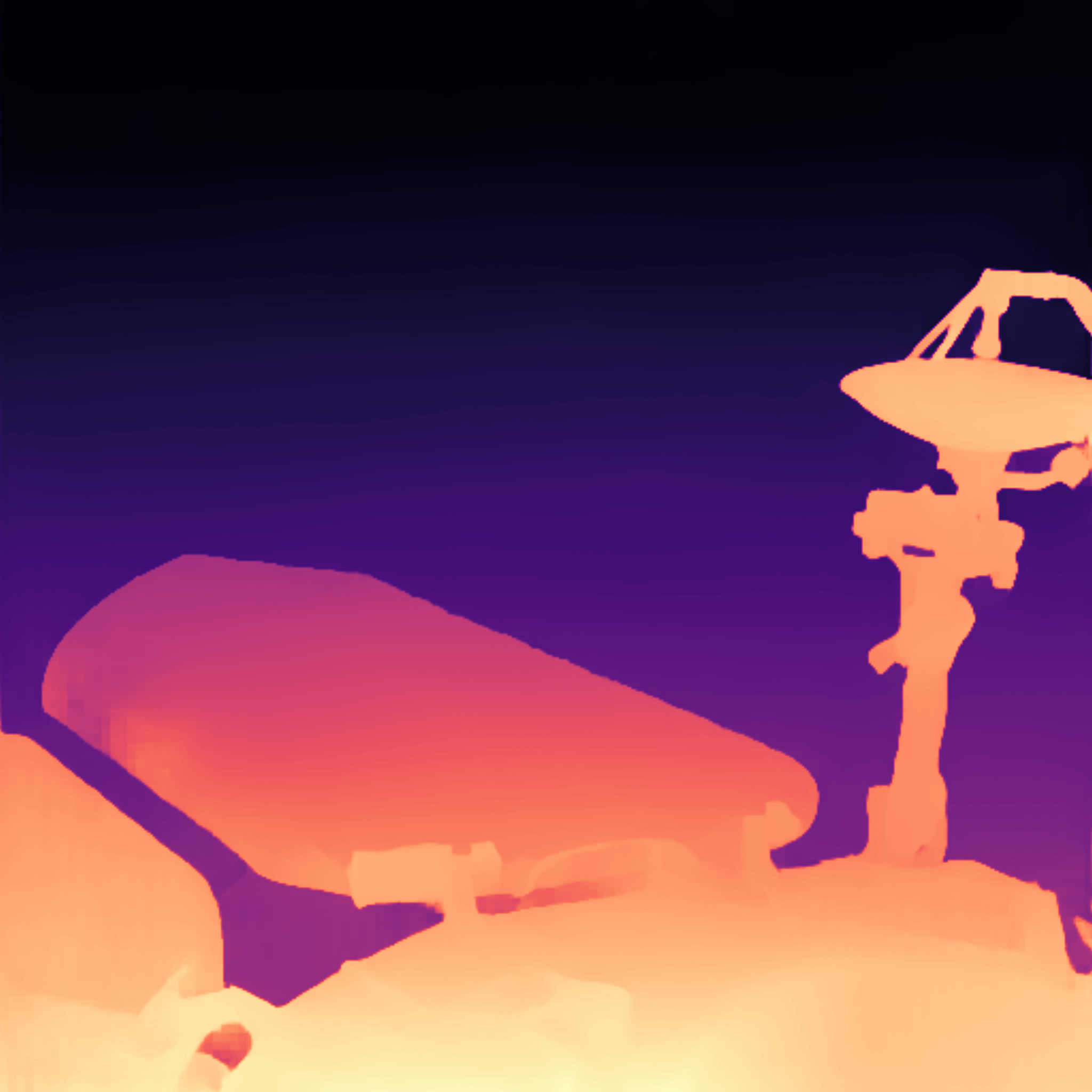}
    \includegraphics[width=0.137\textwidth]{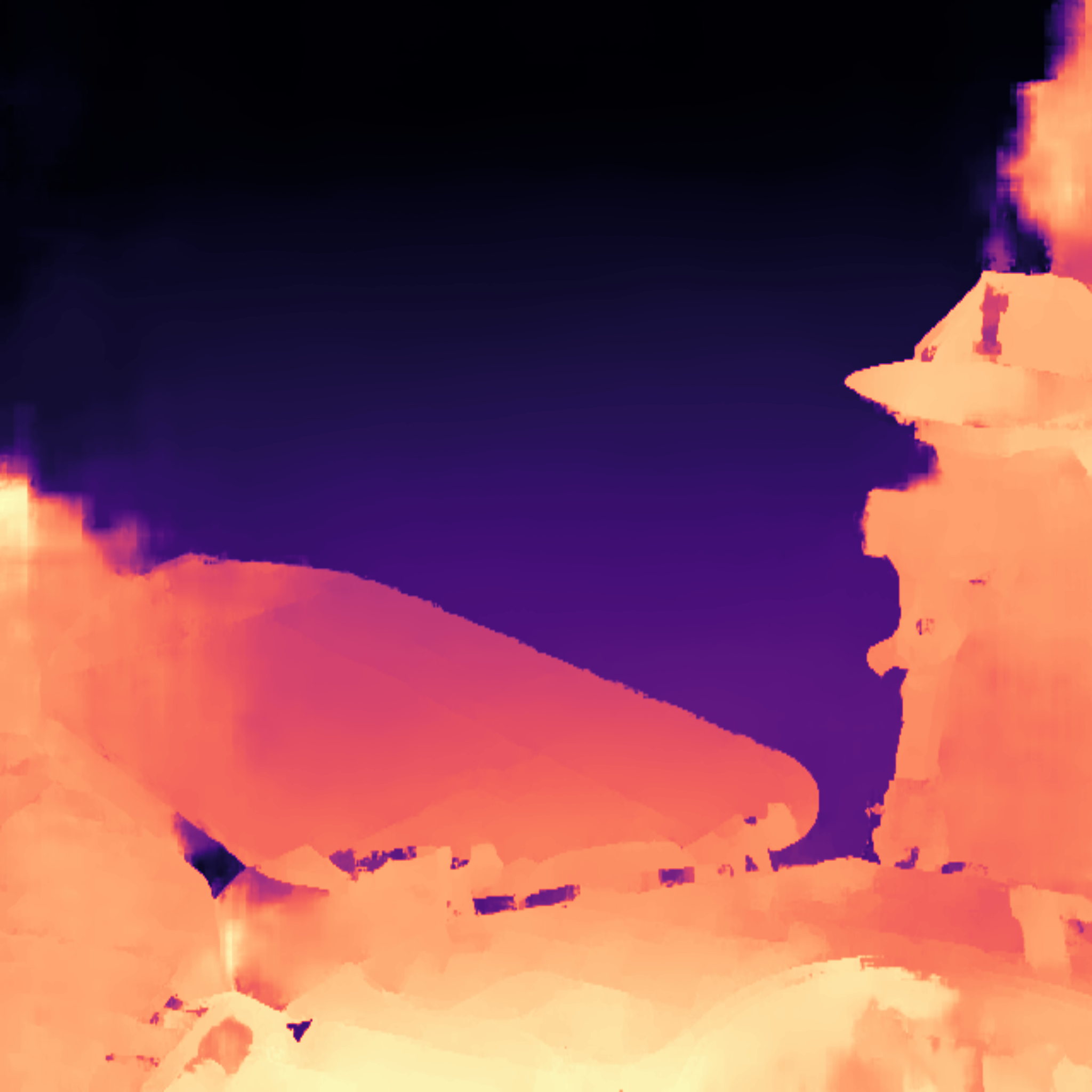}
    \includegraphics[width=0.137\textwidth]{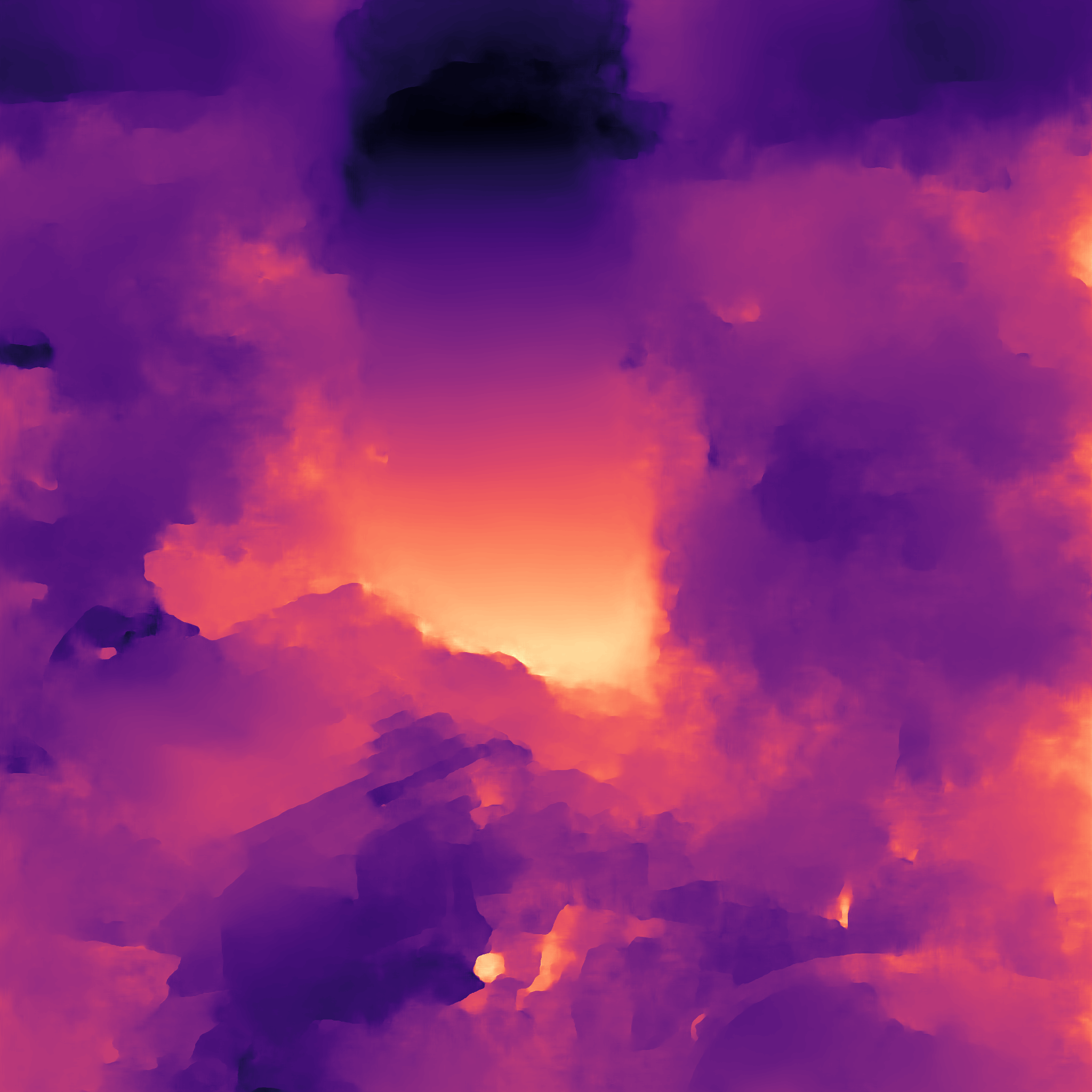}
    \includegraphics[width=0.137\textwidth]{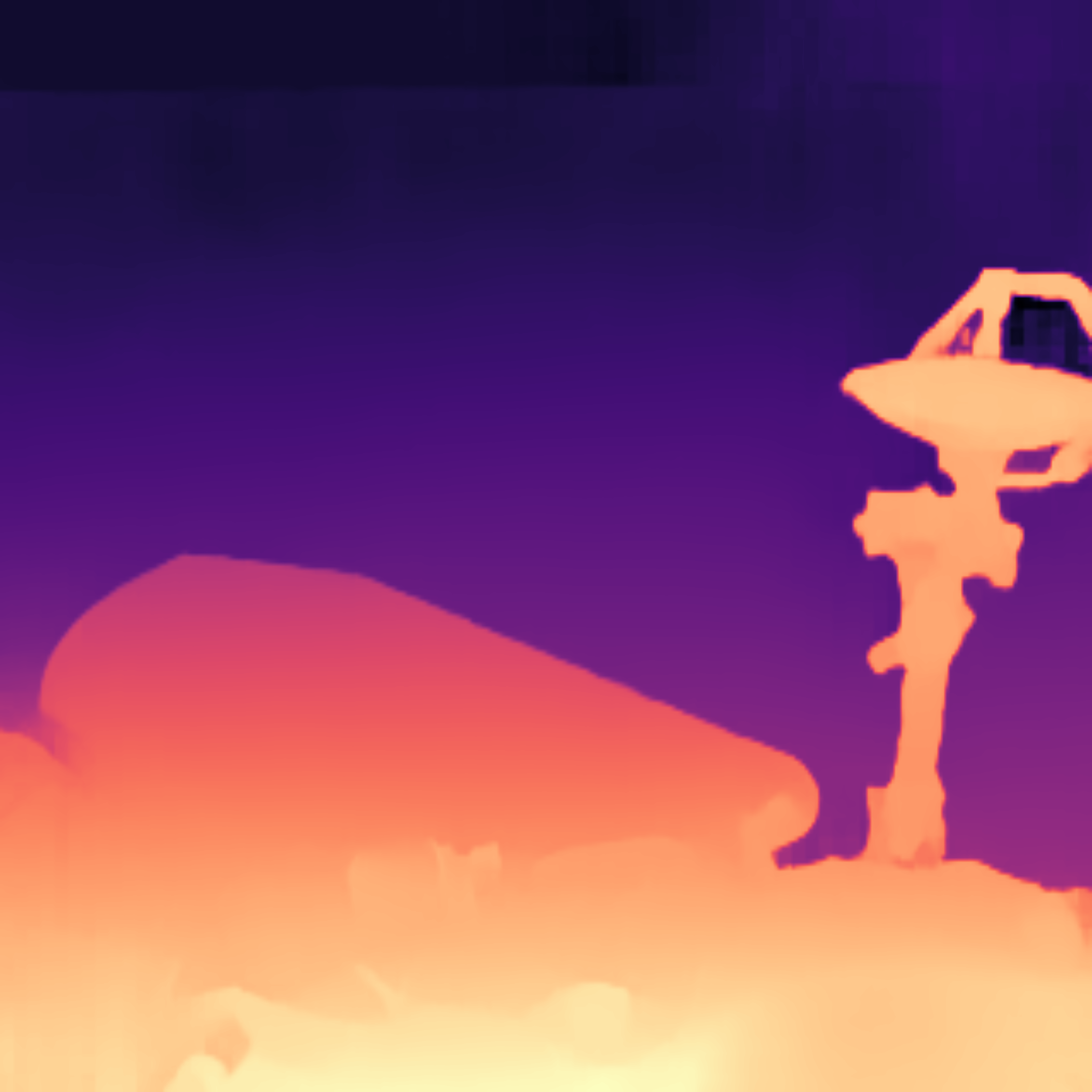}
    \includegraphics[width=0.137\textwidth]{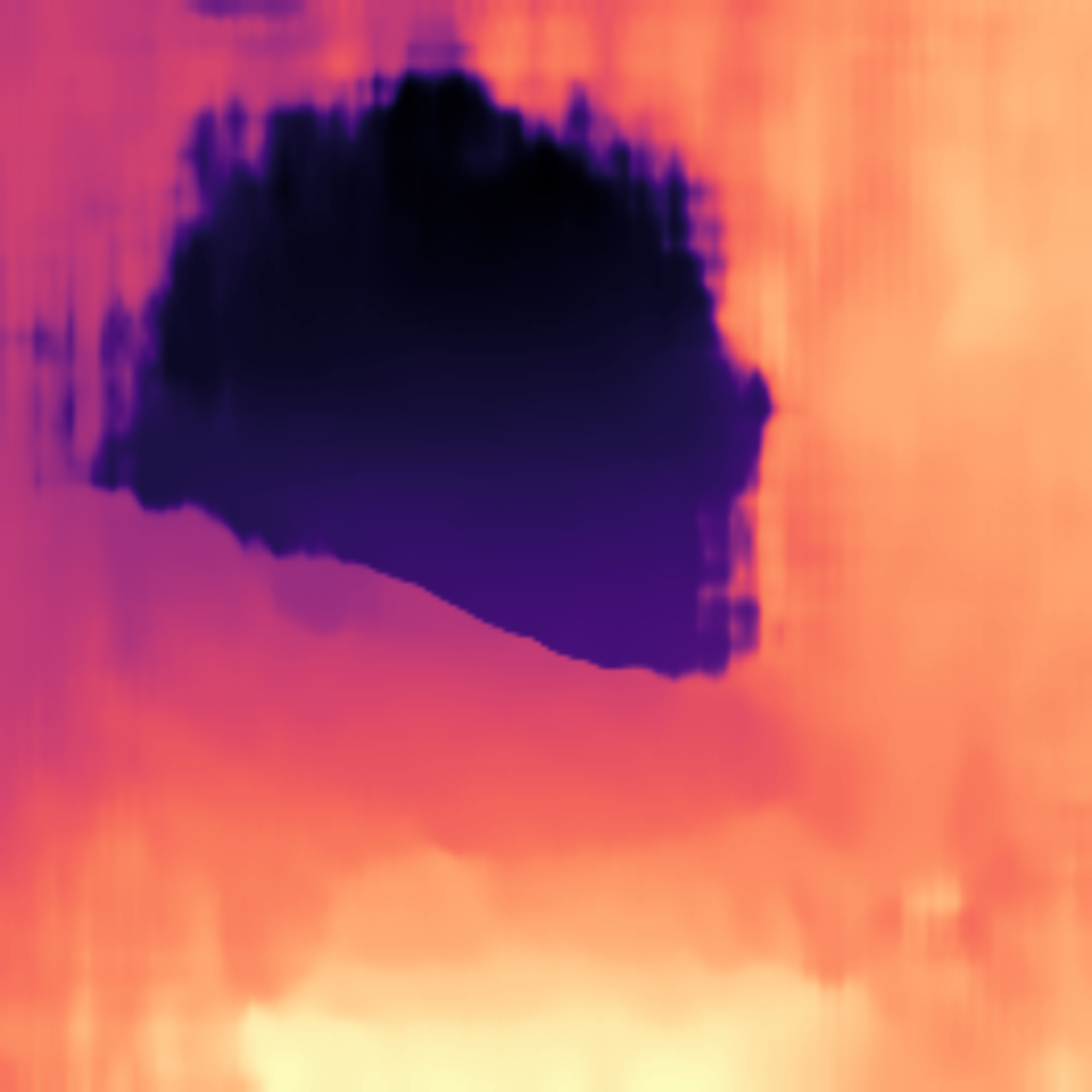}
    \includegraphics[width=0.137\textwidth]{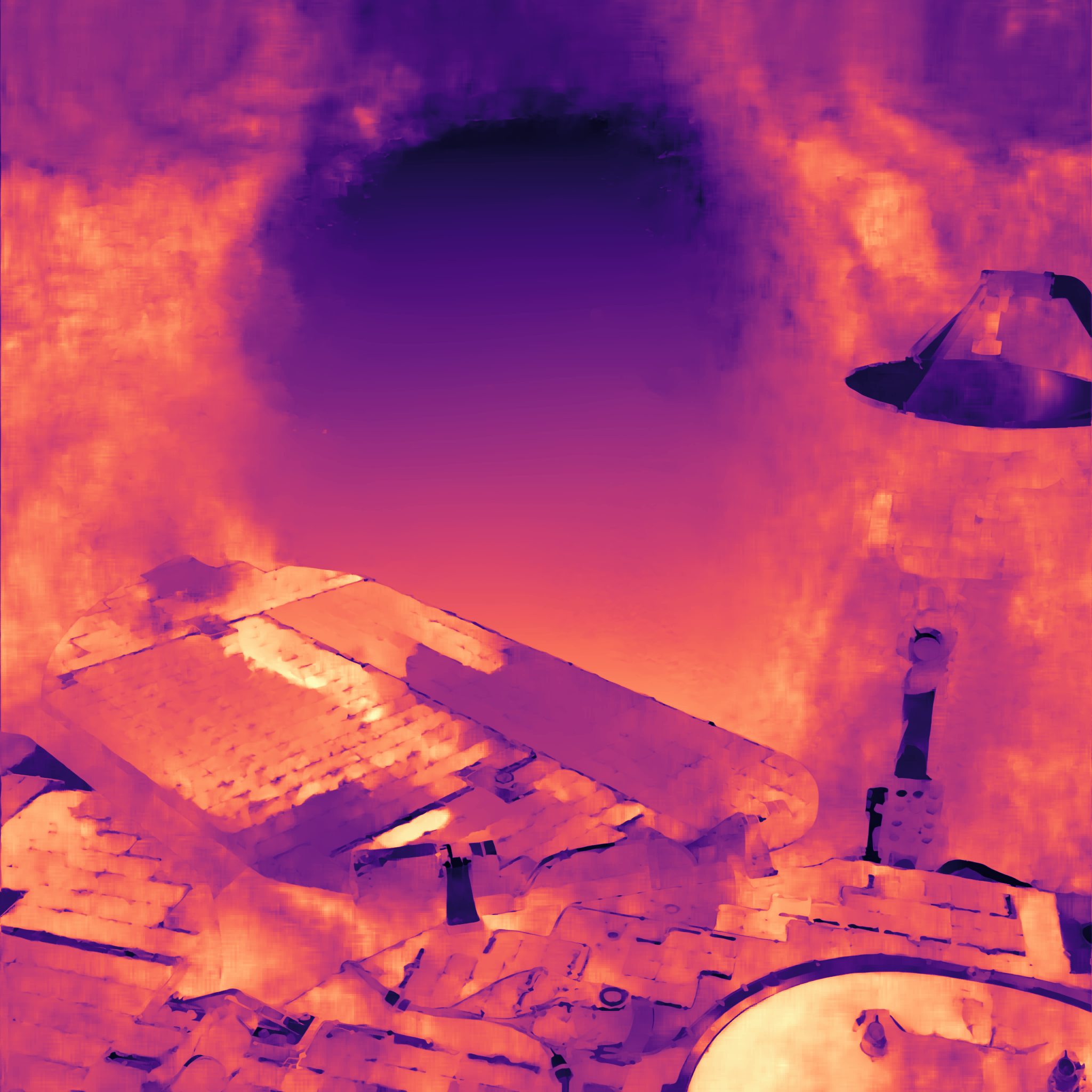}
    \\
    \vspace{0.2em}
    \includegraphics[width=0.137\textwidth]{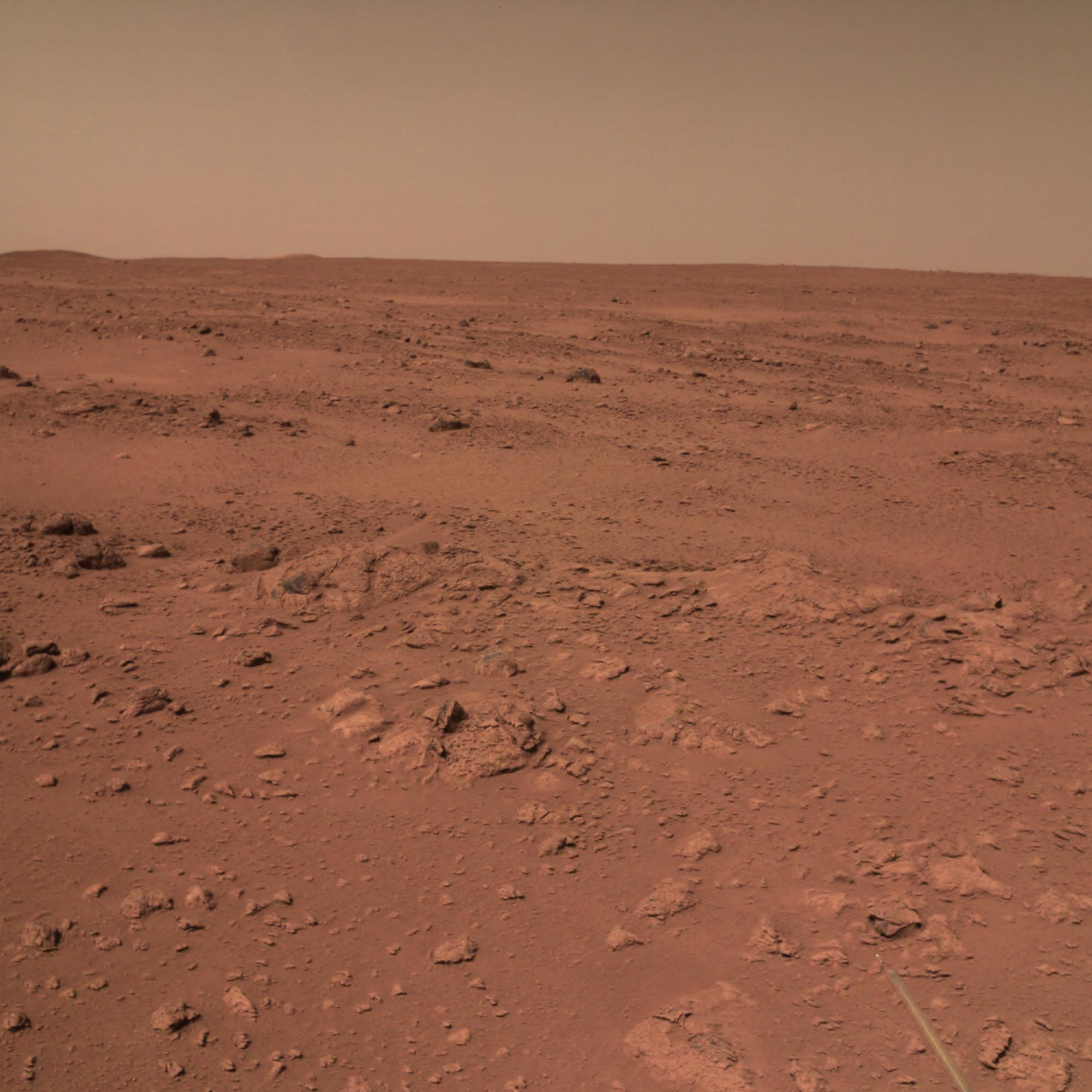}
    \includegraphics[width=0.137\textwidth]{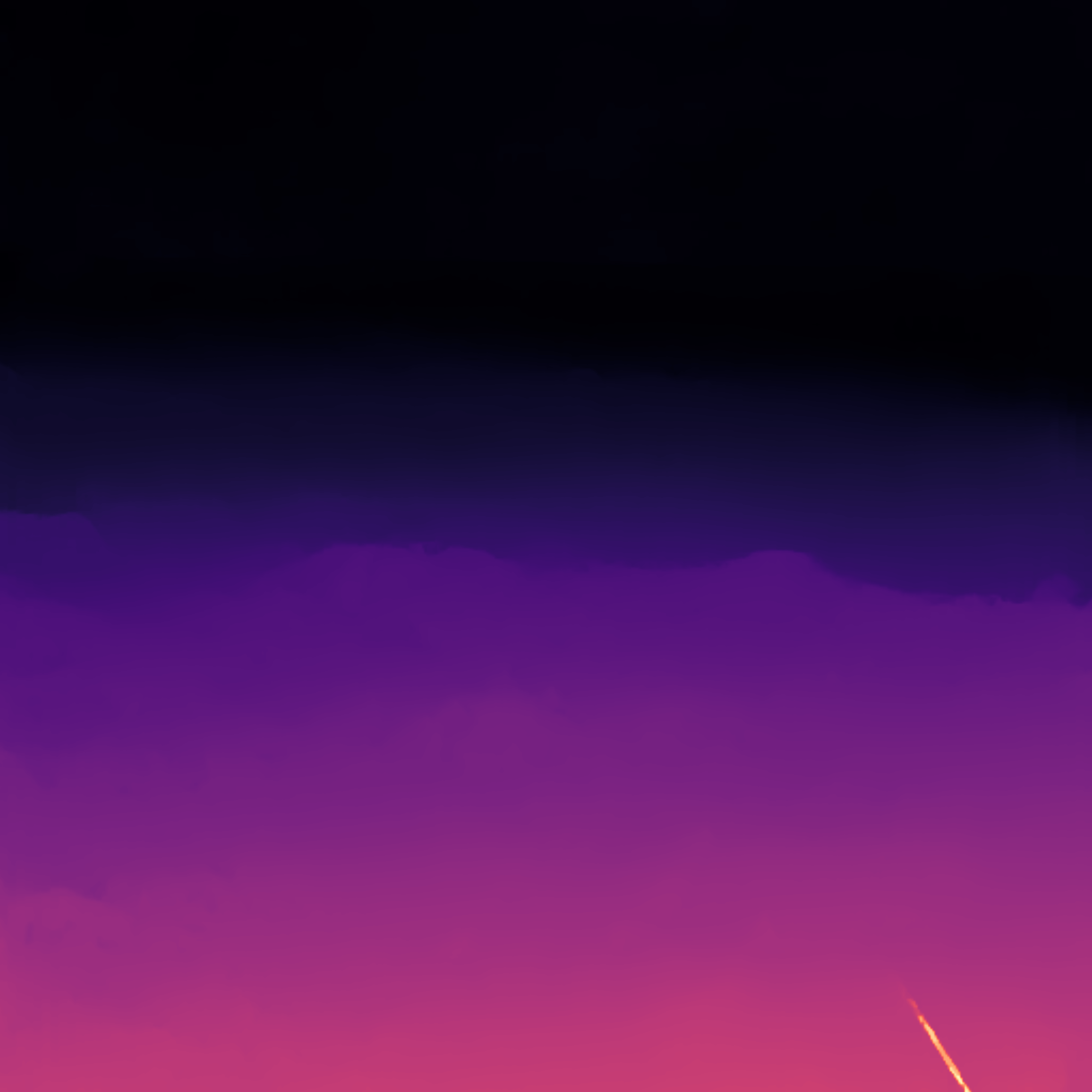}
    \includegraphics[width=0.137\textwidth]{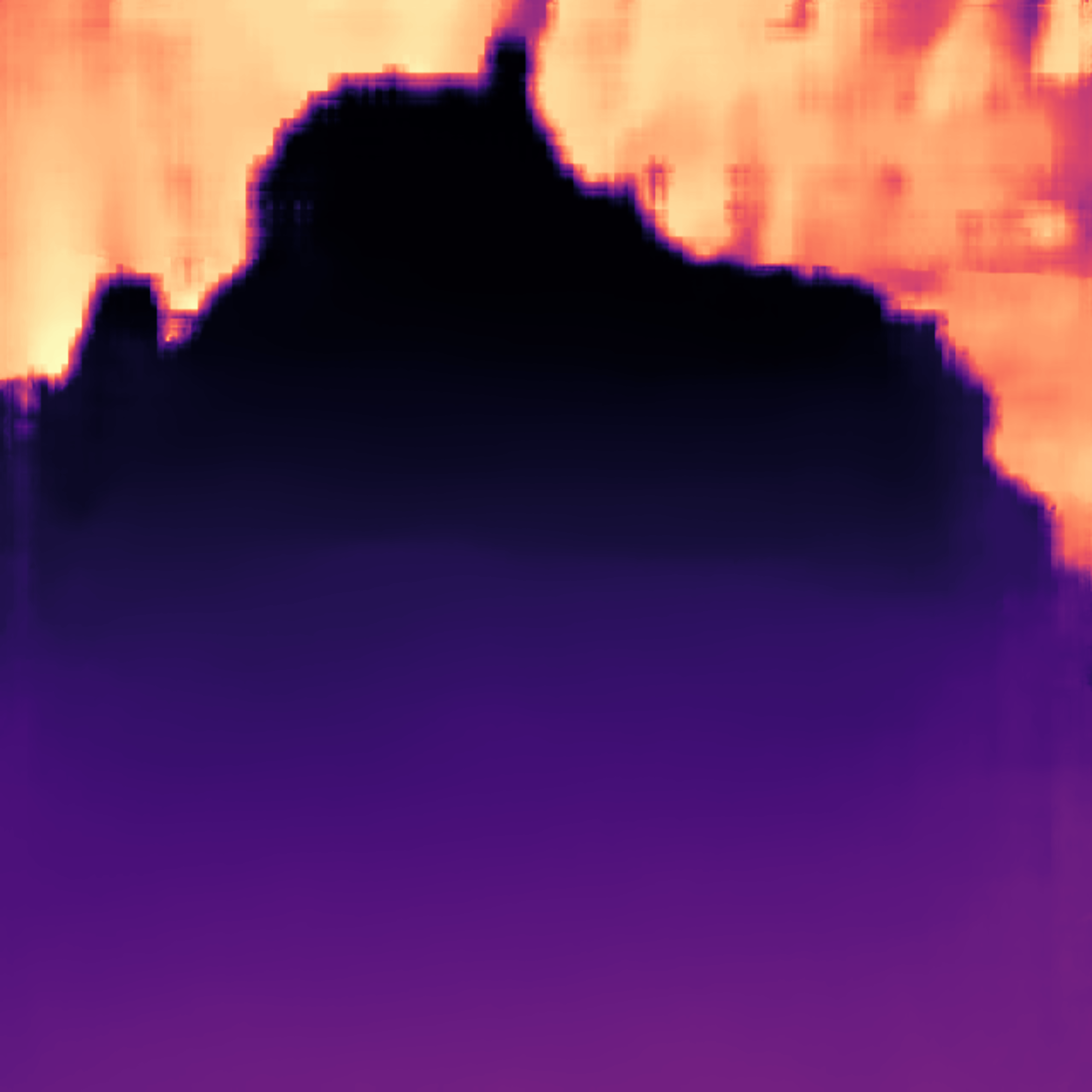}
    \includegraphics[width=0.137\textwidth]{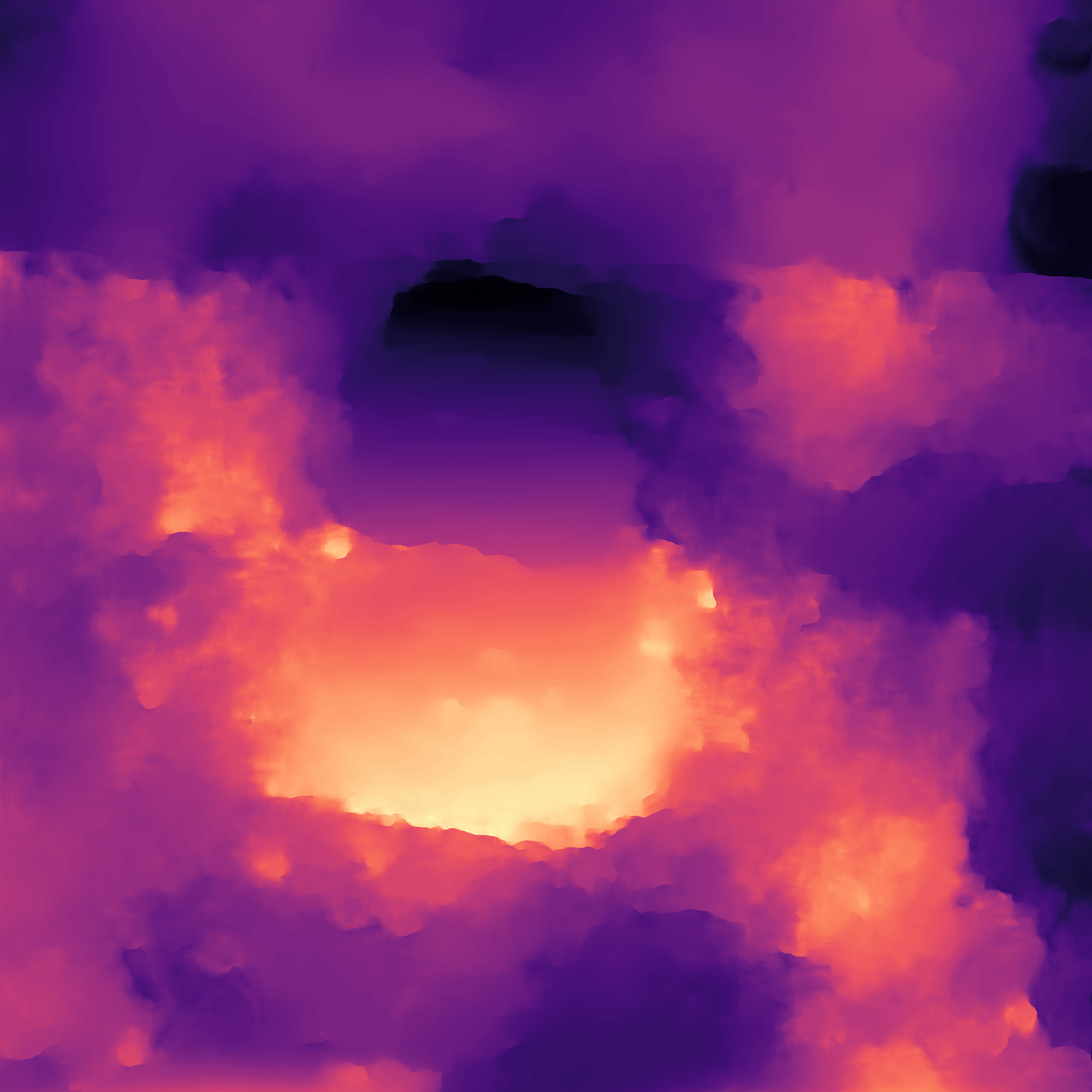}
    \includegraphics[width=0.137\textwidth]{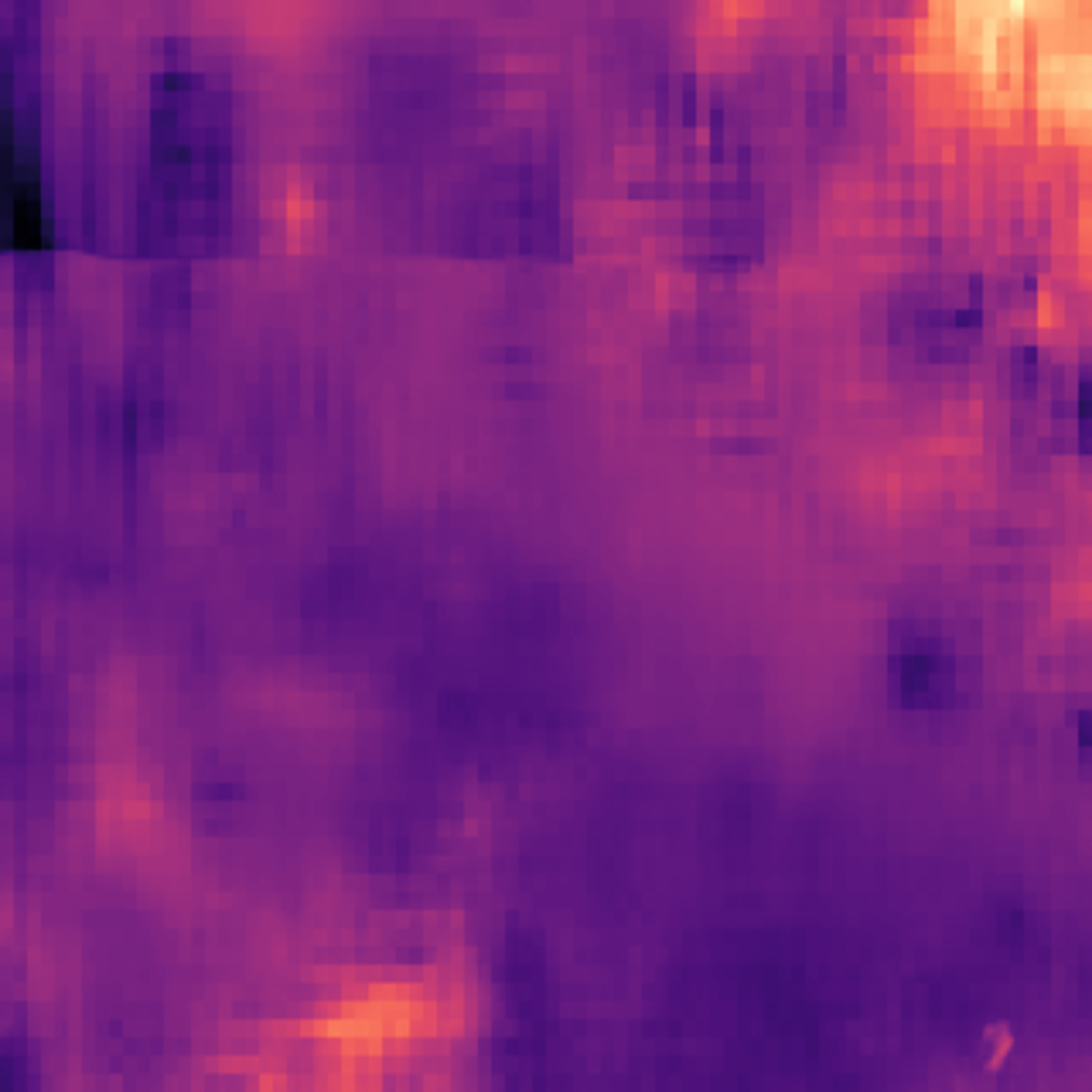}
    \includegraphics[width=0.137\textwidth]{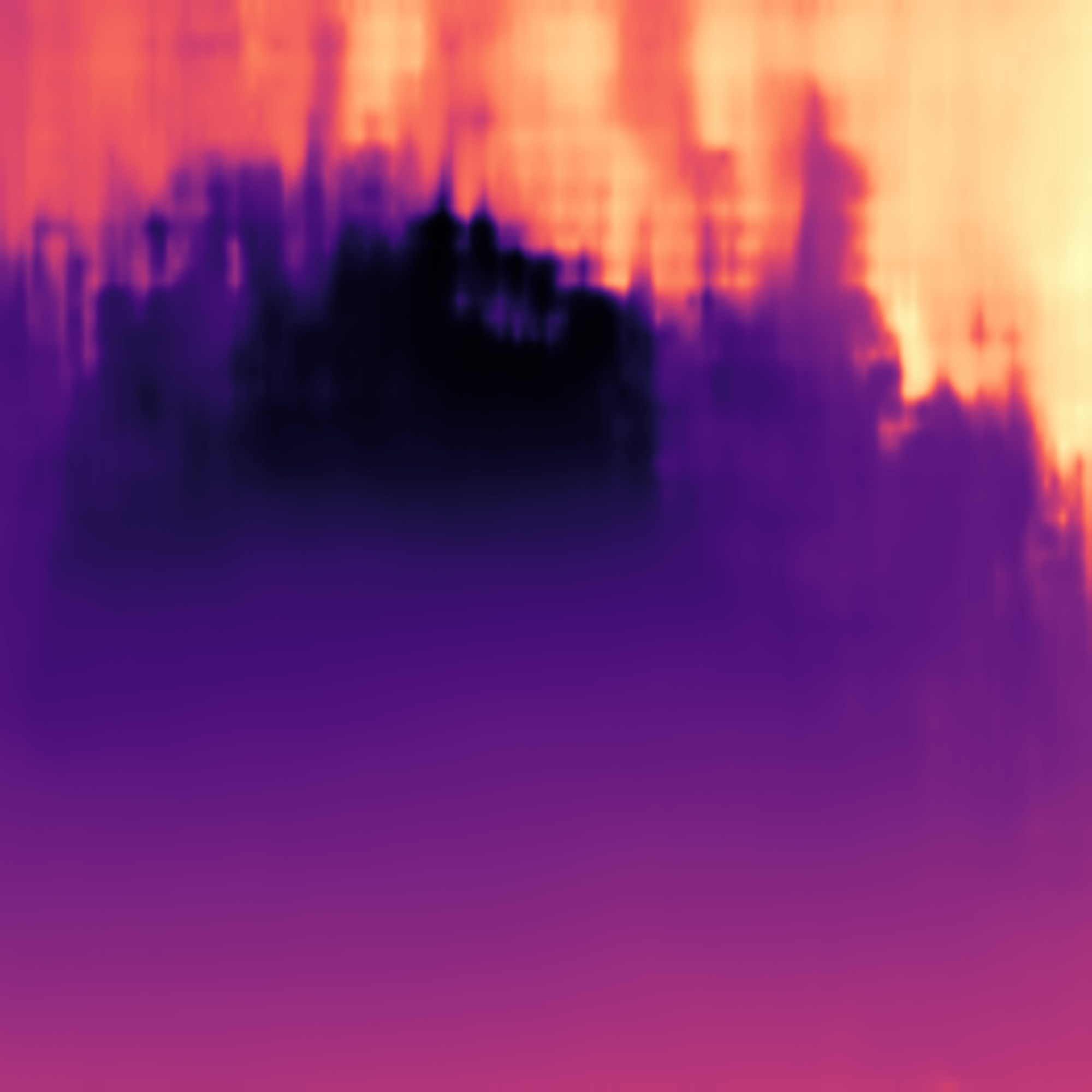}
    \includegraphics[width=0.137\textwidth]{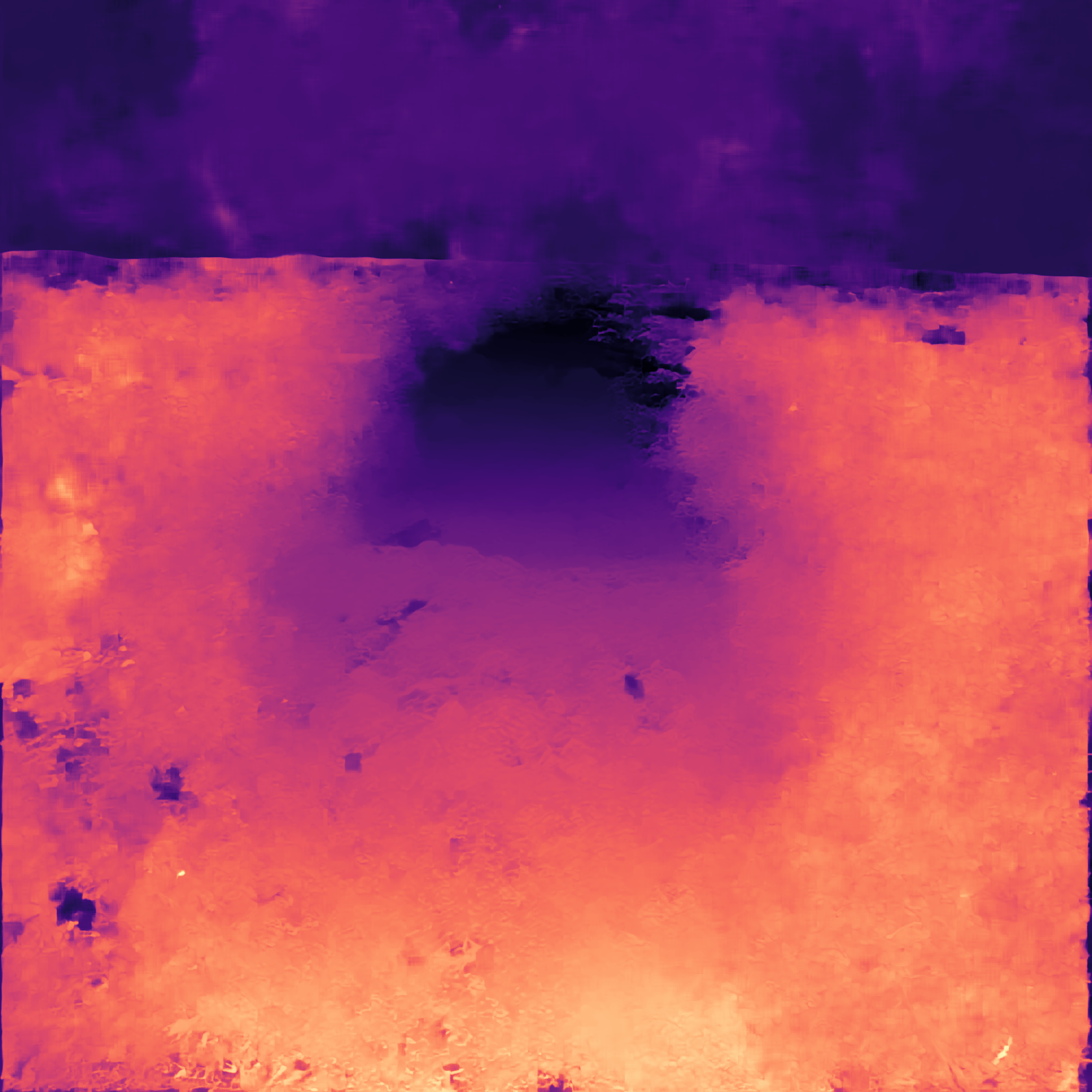}
    \\
    \vspace{0.2em}
    \includegraphics[width=0.137\textwidth]{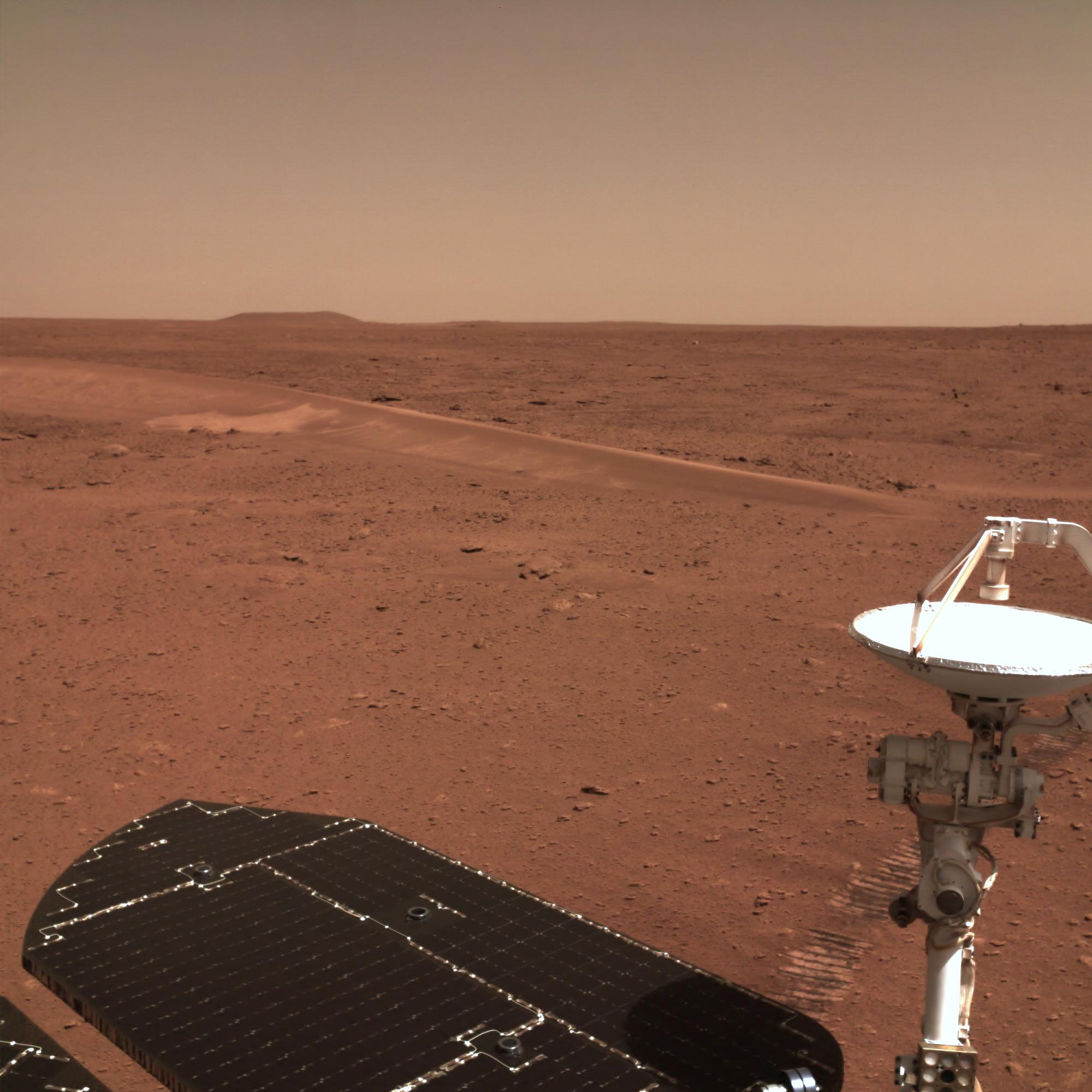}
    \includegraphics[width=0.137\textwidth]{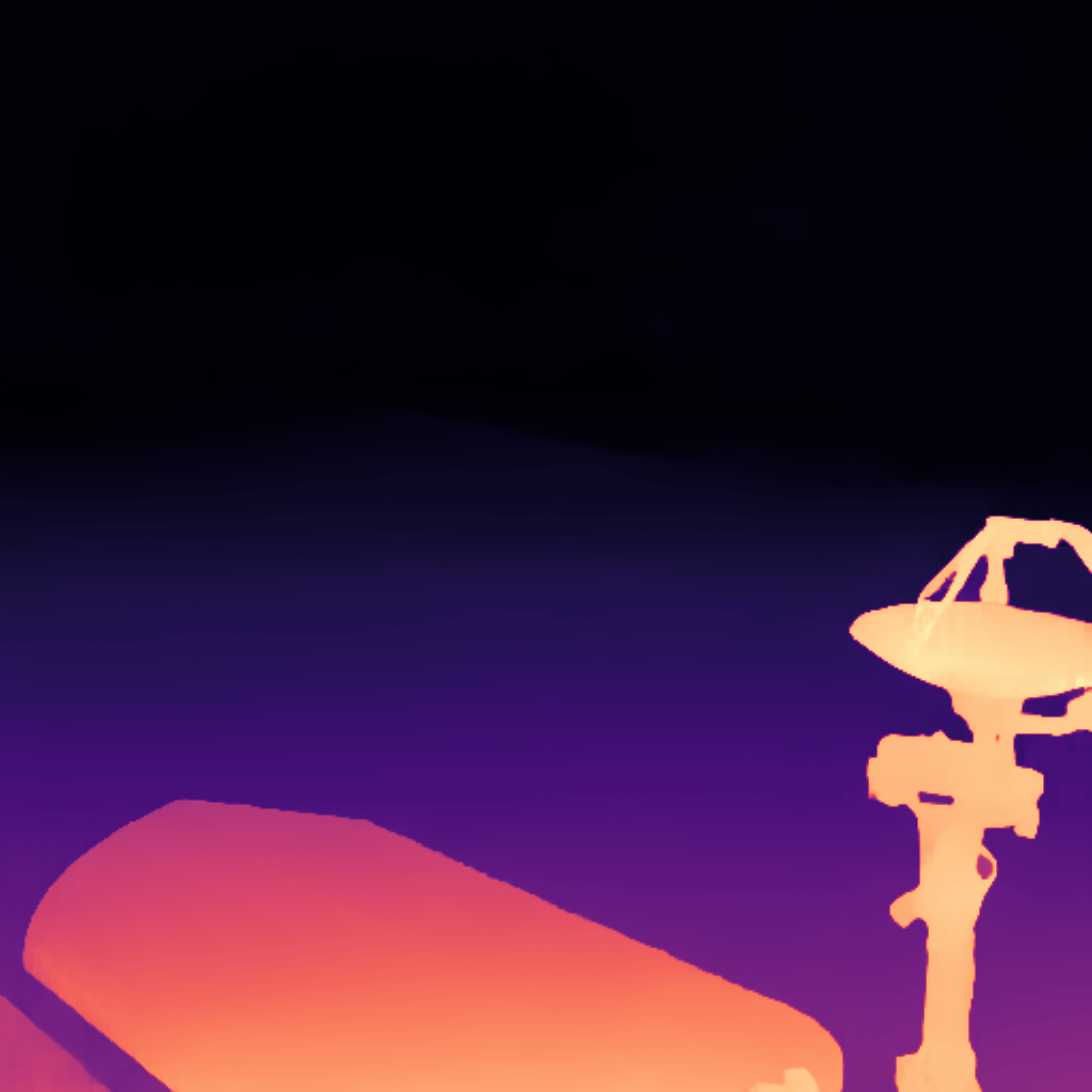}
    \includegraphics[width=0.137\textwidth]{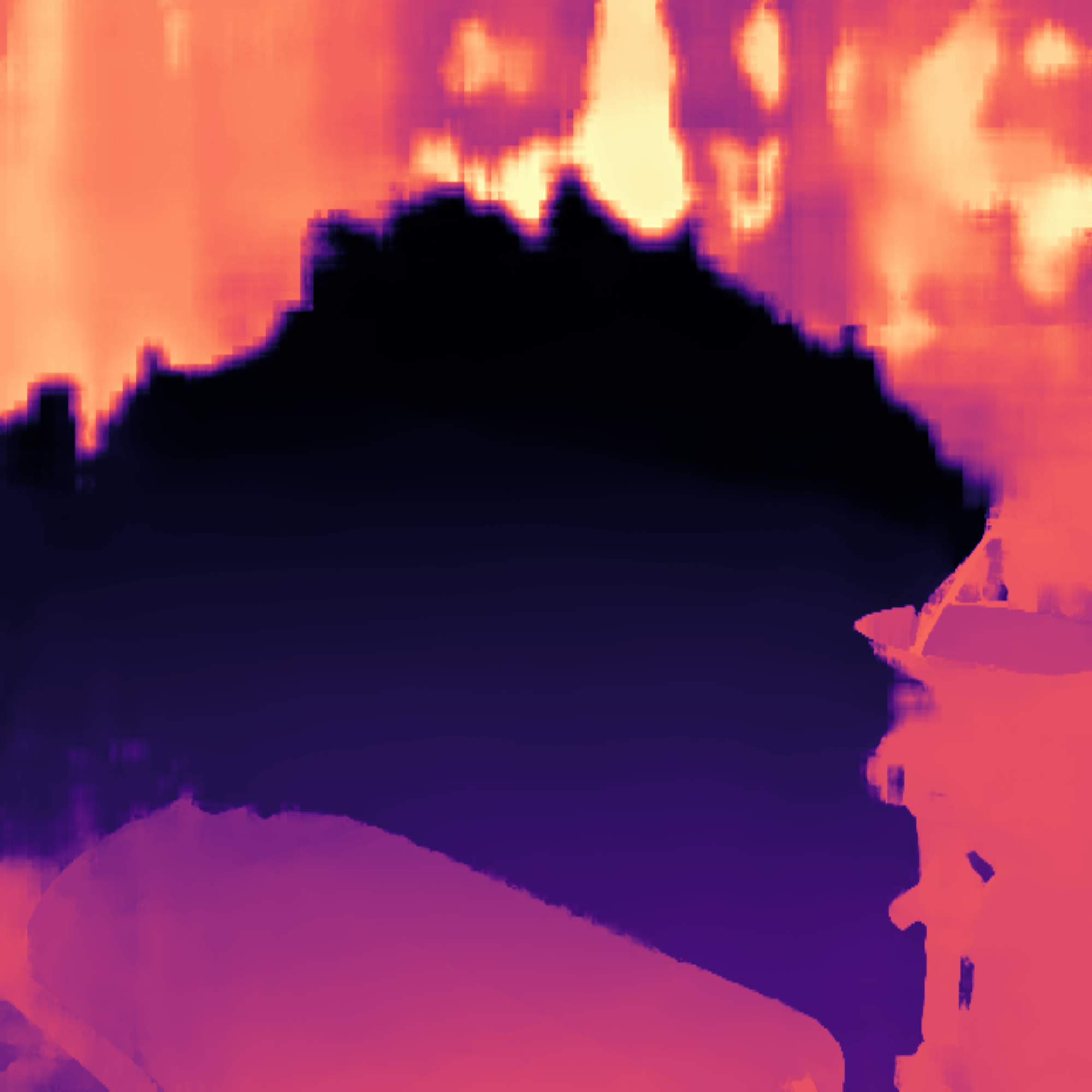}
    \includegraphics[width=0.137\textwidth]{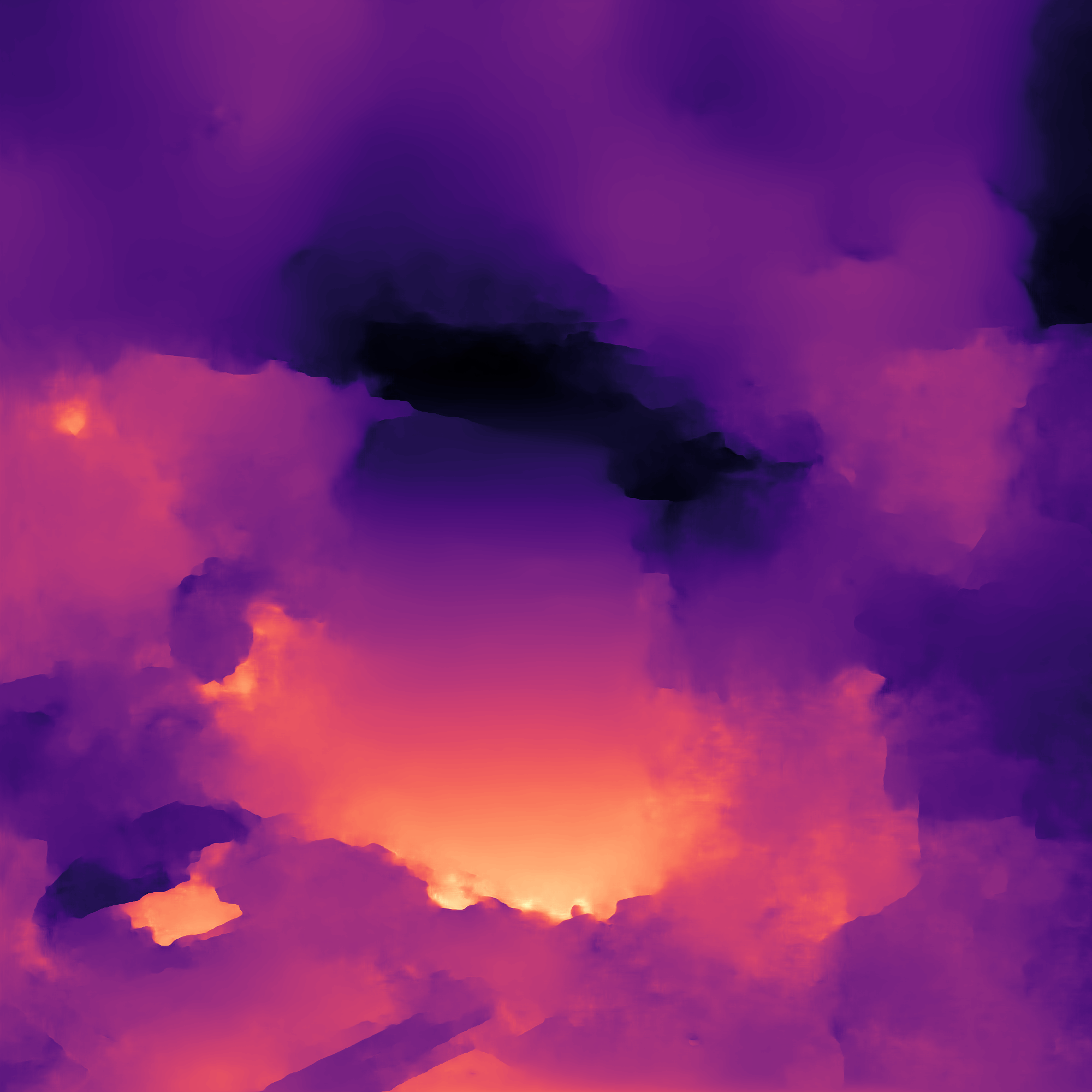}
    \includegraphics[width=0.137\textwidth]{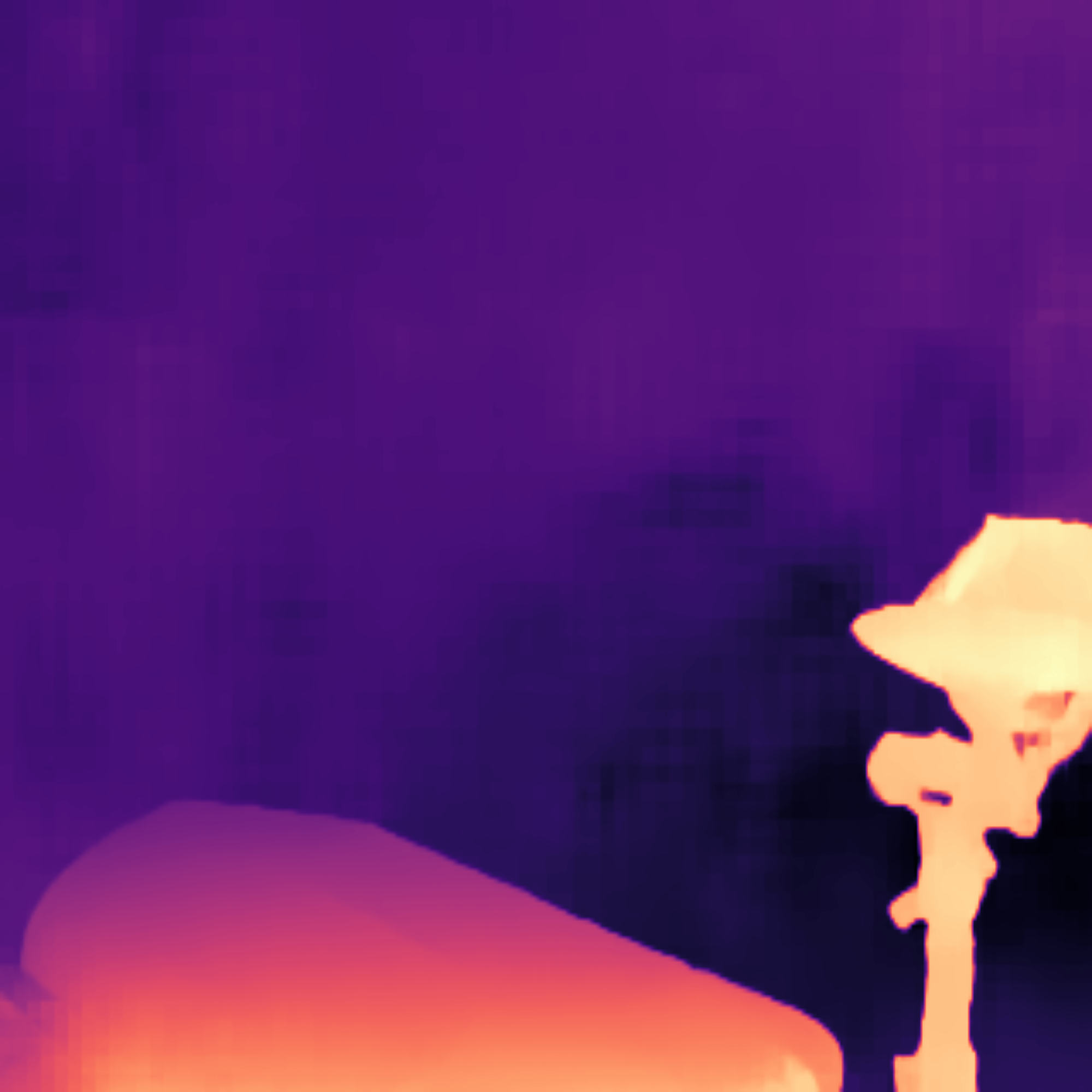}
    \includegraphics[width=0.137\textwidth]{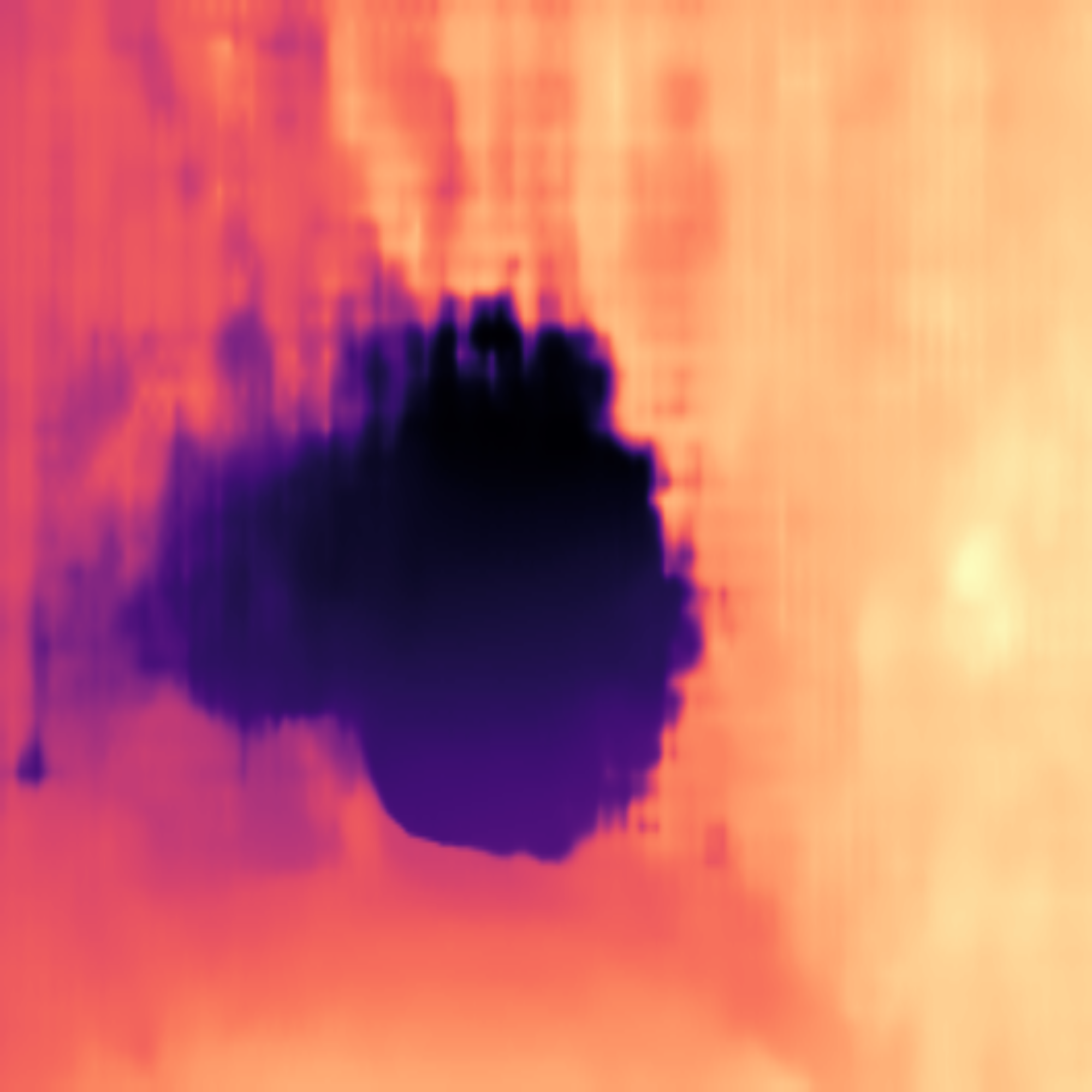}
    \includegraphics[width=0.137\textwidth]{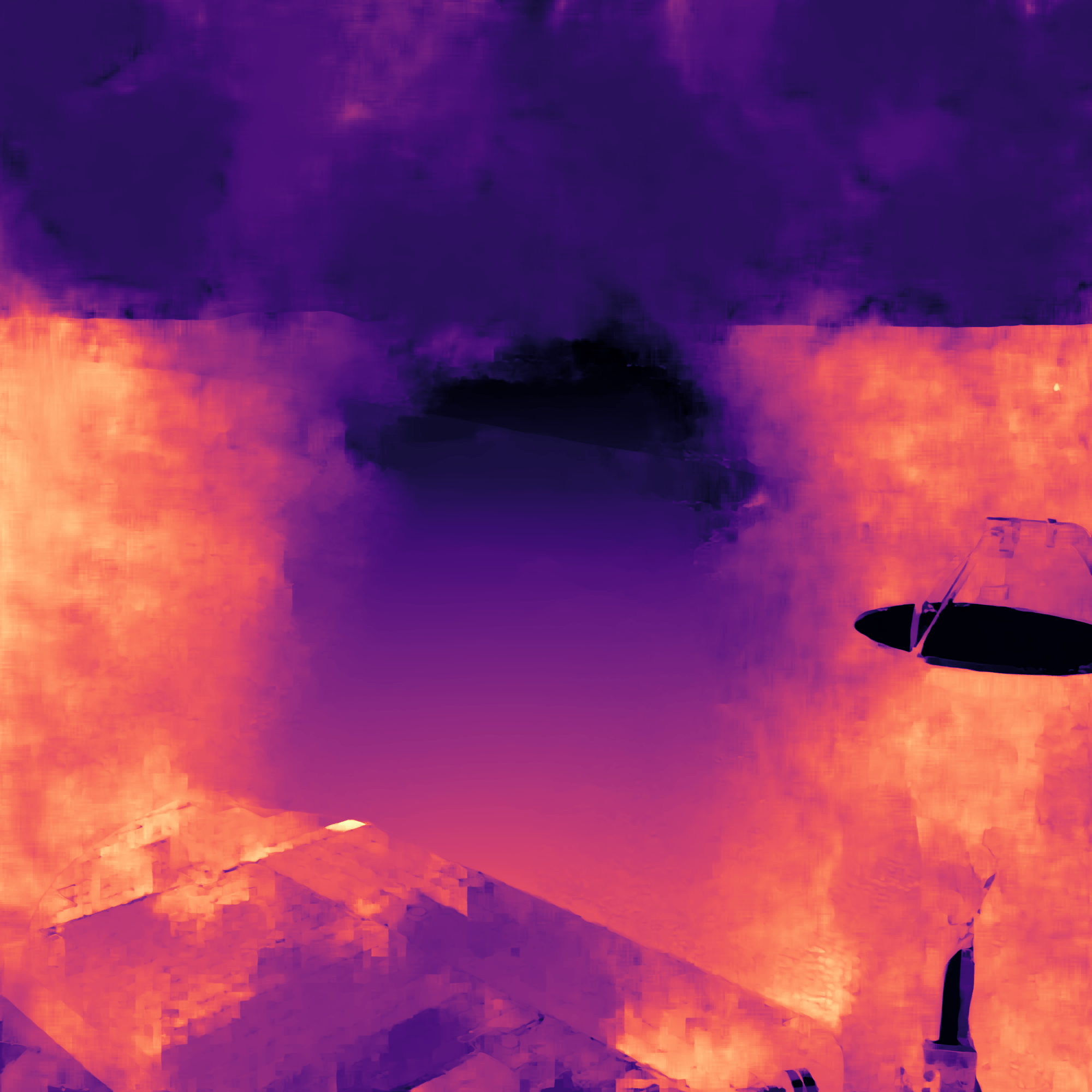}
    \\
    \makebox[0.137\textwidth]{\footnotesize Origin}
    \makebox[0.137\textwidth]{\footnotesize \textbf{M\textsuperscript{3}Depth (Ours)}}
    \makebox[0.137\textwidth]{\footnotesize Raft-Stereo}
    \makebox[0.137\textwidth]{\footnotesize HitNet-Stereo}
    \makebox[0.137\textwidth]{\footnotesize CRE-Stereo}
    \makebox[0.137\textwidth]{\footnotesize UniMVSNet-Stereo}
    \makebox[0.137\textwidth]{\footnotesize Selective-Stereo}
    \\
    \vspace{0.2em}
    \caption{Qualitative evaluation of depth estimation results on Zhurong Rover images, which demonstrates the generalization ability of our method to real-world Martian image data.} 
    \label{zhurong_qualitative}
\end{figure*}

To further demonstrate the applicability of our proposed method in real Martian scenarios, we evaluate M\textsuperscript{3}Depth on the Zhurong dataset, which consists of challenging terrain captured by the Zhurong Rover. As illustrated in \autoref{zhurong_qualitative}, our method outperforms previous state-of-the-art approaches across all samples, achieving superior depth estimation quality under complex real-world conditions.

Specifically, M\textsuperscript{3}Depth excels in handling texture-less surfaces and maintaining geometric consistency, whereas competing methods often fail in smooth and low-texture areas, such as the flat terrain in the second row, most baseline methods suffer from noisy or inconsistent depth estimations. In contrast, our method produces smooth and coherent depth maps, effectively capturing the terrain's gradual variations. Our method accurately delineates sharp structures, such as the rover's components in the first and third rows, where other approaches struggle with boundary artifacts and depth discontinuities. For example, Raft-Stereo and Selective-Stereo introduce noticeable depth bleeding near the rover body and antenna. CRE-Stereo and HitNet-Stereo display more artifacts in areas with varying lighting or occlusions, which is evident in the regions near the rover structures. However, our model remains robust, preserving the integrity of both depth smoothness and fine-grained details.

These results highlight the ability of M\textsuperscript{3}Depth to generalize well to real-world Martian imagery, overcoming challenges such as flat terrains, texture variations, and occlusions. Despite the domain gap between SimMars6K and real-world Zhurong imagery, M\textsuperscript{3}Depth demonstrates adaptability across datasets, highlighting its potential for deployment in real Martian exploration.

\subsubsection{Ablation Studies}

To assess the impact of each key component in our M\textsuperscript{3}Depth framework, we conducted ablation experiments. In \autoref{ablation}, \CheckmarkBold indicates the inclusion of the component, while \XSolidBrush indicates its absence. The two components evaluated include:

\begin{itemize}
    \item Wavelet-Enhanced Feature Extractor (WEFE): enhances the feature extraction process via wavelet transform, which is effective for the multi-frequency response, particularly in low-frequency components.
    \item Iterative Refinement Module (IRM): iteratively refines both depth and surface normal predictions through a mutual boosting mechanism, improving pixel-level depth and surface normal accuracy.
\end{itemize}

\begin{table}[thb]\centering
    \caption{Ablation Study of our proposed method on the depth estimation evaluation.}
    \label{ablation}
    \resizebox{0.48\textwidth}{!}{
    \large
    \begin{tabular}{*{8}{c||cc||cccc||c}}
        \toprule
     \multirow{2}{*}{Setup}  &  \multirow{2}{*}{WEFE}  &  \multirow{2}{*}{IRM}  & Abs Rel & Sq Rel & RMSE & log$_{10}$ & $\delta_1 < 1.25$ \\
      &  &  & $(\downarrow)$ & $(\downarrow)$ & $(\downarrow)$ & $(\downarrow)$ & $(\uparrow)$  \\
       \midrule
     (a) & \XSolidBrush  & \XSolidBrush & 0.121 & 0.083 & 0.562 & 0.055 & 0.825  \\
   (b) & \XSolidBrush  & \CheckmarkBold & 0.115 & 0.071 & 0.462 & 0.051 & 0.871 \\
   (c) & \CheckmarkBold & \XSolidBrush & 0.113 & 0.066 & 0.418 & 0.049 & 0.882  \\
   (d) & \CheckmarkBold & \CheckmarkBold & 0.089 & 0.058 & 0.314 & 0.038 & 0.905 \\
        \bottomrule
    \end{tabular}
    }
\end{table}

\textbf{Ablation Evaluation on Depth Estimation.} As summarized in \autoref{ablation}, the ablation study highlights the contributions of the WEFE and IRM module to the overall performance of our proposed M\textsuperscript{3}Depth model. The baseline model with both modules absence, \textit{i.e.,} Setup (a), performs the worst, with an Abs Rel of 0.121 and RMSE of 0.562, indicating its limited ability to handle Martian terrain challenges. Setup (b) enables only the IRM, achieving an Abs Rel of 0.115 and an RMSE of 0.462. The IRM progressively refines depth estimations, correcting local inconsistencies and enhancing fine-grained details through mutual boosting between depth and surface normal maps. This iterative process is particularly beneficial in handling depth discontinuities and boundary preservation. Similarly, Setup (c) introduces WEFE alone, which reduces Abs Rel to 0.113 and RMSE to 0.418, demonstrating its effectiveness in enhancing feature extraction for texture-less surfaces and boundary regions. 

The full model Setup (d) combines WEFE and IRM simultaneously, achieving the best performance, with an Abs Rel of 0.089 and an RMSE of 0.311. The accuracy at $\delta < 1.25$ also reaches 90.5\%, highlighting the complementary roles of WEFE and IRM in achieving accurate and robust depth estimation. WEFE provides a robust feature foundation by capturing both high-frequency and low-frequency details, while IRM iteratively refines these predictions to achieve superior accuracy and consistency.
 

\textbf{Ablation Evaluation on Surface Normal Prediction.} To further evaluate the individual contributions of the key components in our proposed framework, we conduct an ablation study on surface normal estimation. The results of the ablation study for surface normal estimation, as presented in \autoref{ablation_normal}, further demonstrate the respective contributions of the WEFE and the IRM to our model. The addition of WEFE leads to a substantial improvement in performance, with the Mean Error reduced to $25.3^\circ$, the Median Error decreased to $17.2^\circ$, and the RMSE decreased to $40.7^\circ$. Furthermore, the $11.25^\circ$ accuracy increases to 36.1\%. This highlights the importance of an enhanced feature extraction strategy, emphasizing that low-frequency signals are also essential for accurate surface normal estimation. On the other hand, the IRM module demonstrates a more pronounced impact on depth estimation than surface normal estimation. This is mainly because surface normals inherently rely on local gradients, which are less affected by iterative refinements.

\begin{table}[thb]\centering
    \caption{Ablation Study of our proposed method on the surface normal map prediction evaluation.}
    \label{ablation_normal}
    \resizebox{0.48\textwidth}{!}{
    \large
    \begin{tabular}{*{7}{c||cc||ccc||c}}
        \toprule
      \multirow{2}{*}{Setup} &  \multirow{2}{*}{WEFE} & \multirow{2}{*}{IRM} & Mean & Median & $\text{RMSE}_n $ & $11.25^\circ$ \\
       &  &  & $(\downarrow)$ & $(\downarrow)$ & $(\downarrow)$ &  $(\uparrow)$  \\
       \midrule
    (a) &  \XSolidBrush  & \XSolidBrush & 33.5 & 24.2 & 50.9 & 0.179 \\
   (b) & \XSolidBrush  & \CheckmarkBold & 29.6 & 20.8 & 46.9 & 0.275 \\
    (c) & \CheckmarkBold & \XSolidBrush & 25.3 & 17.2 & 40.7 & 0.361 \\
   (d) & \CheckmarkBold & \CheckmarkBold & 20.6 & 12.3 & 34.2 & 0.435 \\
        \bottomrule
    \end{tabular}
    }
\end{table}

\begin{figure*}[t] \centering
    \includegraphics[width=0.24\textwidth]{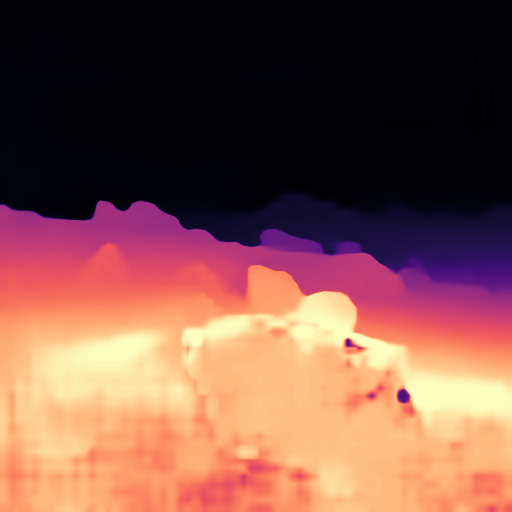}
 \includegraphics[width=0.24\textwidth]{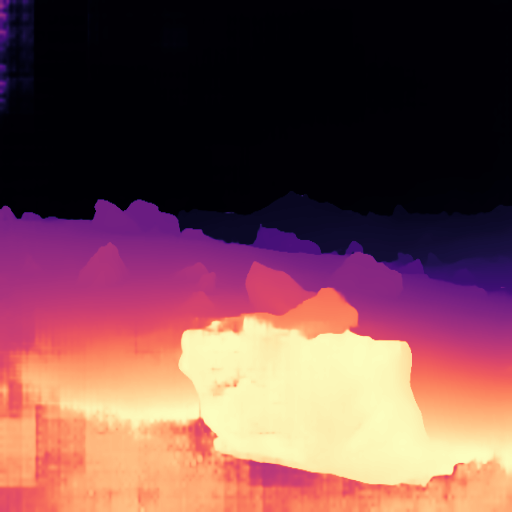}
 \includegraphics[width=0.24\textwidth]{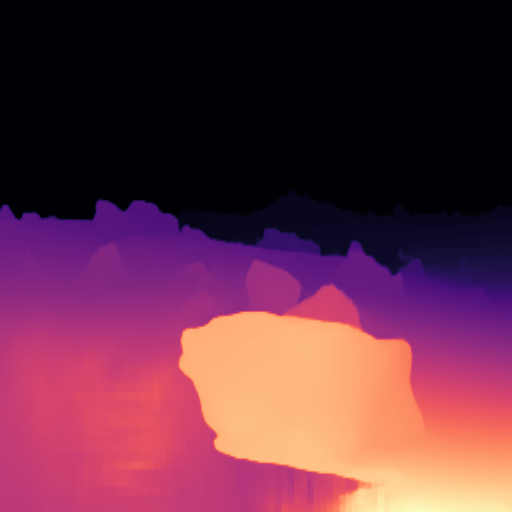}
 \includegraphics[width=0.24\textwidth]{results/ours/00134Left.png} 
 \\
\vspace{0.3em}
\includegraphics[width=0.24\textwidth]{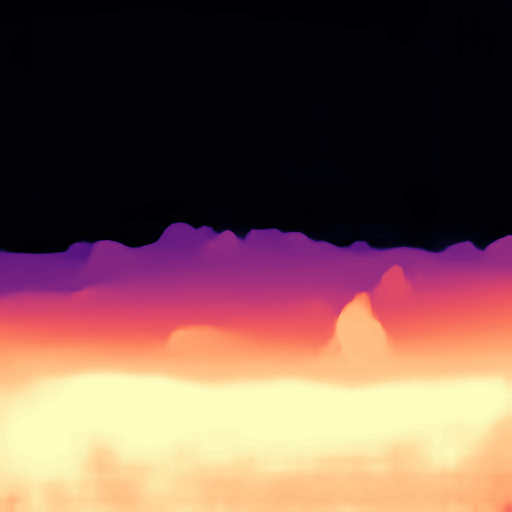}
\includegraphics[width=0.24\textwidth]{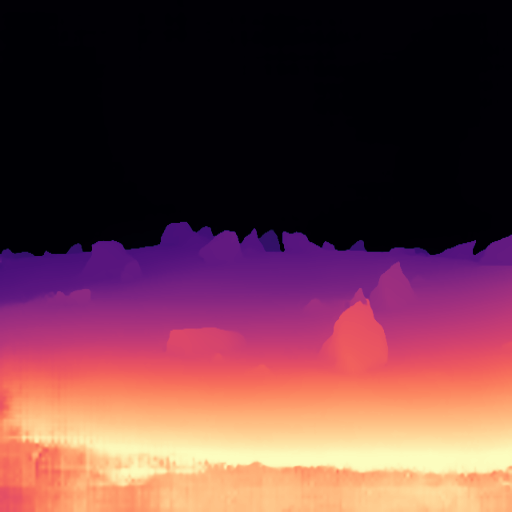}
\includegraphics[width=0.24\textwidth]{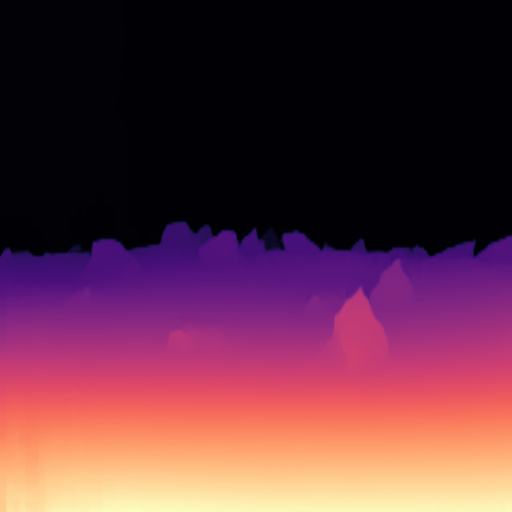}
\includegraphics[width=0.24\textwidth]{results/ours/00042Left.png} \\
    \makebox[0.24\textwidth]{\footnotesize (a) \textit{w/o} WEFE, \textit{w/o} IRM}
    \makebox[0.24\textwidth]{\footnotesize (b) \textit{w/o} WEFE, \textit{w/} IRM}
    \makebox[0.24\textwidth]{\footnotesize (c) \textit{w/} WEFE, \textit{w/o} IRM}
    \makebox[0.24\textwidth]{\footnotesize (d) \textit{w/} WEFE, \textit{w/} IRM}
    \\
    \vspace{0.2em}
    
    \caption{Ablation study visualization results. Comparison of depth estimation results under different ablation settings. \textit{w/} means ``with", \textit{w/o} means ``without".}
    \label{ablation_vis}
\end{figure*}

\textbf{Ablation Evaluation with Visualization.} To provide a more intuitive understanding of the contribution of each module in our proposed method, we visualize the depth estimation results under different ablation settings, as shown in \autoref{ablation_vis}. From the results, it is evident that the baseline model Seutp (a), which lacks both WEFE and IRM, produces depth maps with evident artifacts, particularly in texture-less regions and at object boundaries. In Setup (b), it shows sharper boundaries. However, this model struggles with texture-less regions and exhibits poor depth continuity. In Setup (c), solely introducing WEFE results in a notable improvement, which demonstrates smoother depth transitions and reduced artifacts. However, due to the absence of the surface normal as an auxiliary geometric information, the model tends to over smooth object boundaries, such as the edges of rocks. 

\textbf{Ablation on Consistency Constraints Loss Function.} To validate the effectiveness of the consistency constraints loss function, we conduct an ablation study by training different configurations of our model.

\begin{enumerate}
    \item Full network loss function configuration: $\mathcal{L}_{overall}=\lambda_d\mathcal{L}_d+\lambda_n\mathcal{L}_n + \lambda_c\mathcal{L}_c$ ($\lambda_d=2,\lambda_n=1,\lambda_s=3$).
    \item Loss function configuration without consistency constraints: $\mathcal{L}_{w/o}=\lambda_d\mathcal{L}_d+\lambda_n\mathcal{L}_n$ ($\lambda_d=2,\lambda_n=1$).
    \item Varying the $\lambda$ parameter in the loss function: different $\lambda$ values, assessing the sensitivity of the model to different hyperparameter. 
\end{enumerate}

\begin{table}[thb]
    \centering
    \caption{Ablation study of different loss function configuration.}
    \label{lambda}
    \resizebox{0.48\textwidth}{!}{
    \begin{tabular}{*{5}{c||ccc||c}}
    \toprule
      \multirow{2}{*}{Configuration} & Abs Rel & Sq Rel & RMSE & $\delta_1 < 1.25$ \\
       & $(\downarrow)$ & $(\downarrow)$ & $(\downarrow)$ &  $(\uparrow)$  \\
       \midrule
   $\mathcal{L}_{overall}$ & 0.089 & 0.058 & 0.314 &  0.905 \\
   $\mathcal{L}_{w/o}$ & 0.132 & 0.091 & 0.402 & 0.863 \\
   $\lambda_d=1, \lambda_n=2, \lambda_c=3$ & 0.115 & 0.074 & 0.370 & 0.875 \\
   $\lambda_d=3, \lambda_n=2, \lambda_c=1$ & 0.125 & 0.082 & 0.390 & 0.868 \\
   $\lambda_d=1, \lambda_n=3, \lambda_c=2$ & 0.102 & 0.065 & 0.348 & 0.891 \\
    \bottomrule
    \end{tabular}
}
\end{table}

The results of the ablation study are shown in \autoref{lambda}. From the comparison of the different configurations, we observe that the full loss function configuration $\mathcal{L}_{overall}$, which includes the consistency loss term $\mathcal{L}_c$, achieves the best performance across all metrics. In the absence of the consistency loss term $\mathcal{L}_c$, we observe a performance degradation, with Abs Rel increasing to 0.132, RMSE rising to 0.402, and $\delta_1$ dropping to 0.863. This suggests that the consistency constraint plays a crucial role in improving the coherence of depth and surface normal maps. Without $\mathcal{L}_c$, the model tends to optimize depth and normal maps independently, resulting in less coherent geometric structures. This is particularly evident from the Sq Rel values, which are higher without the consistency term. The lack of alignment between depth and normal predictions leads to depth discontinuities in the output.

We further analyze the impact of varying the hyperparameters $\lambda_d$, $\lambda_n$, and $\lambda_c$ on the performance. The results show that small adjustments in these parameters lead to noticeable differences in the model's behavior. When $\lambda_c$ is smaller than $\lambda_d$ (Row 4), the model tends to overfit local depth cues, causing an overemphasis on depth accuracy at the expense of maintaining geometric consistency. The reduced importance of $\mathcal{L}_c$ results in an inaccurate alignment between depth and normal maps, causing a decrease in the global structure. In contrast, when $\lambda_c$ is larger than $\lambda_d$ (Row 5), the model focuses more on the geometric regularization from $\mathcal{L}_c$, resulting in a more balanced solution where the depth and normal maps are better aligned. This ablation study highlights the trade-off between accuracy in depth estimation and the preservation of geometric consistency. Too much focus on depth loss (by increasing $\lambda_d$) may lead to overfitting of local details, whereas increasing $\lambda_c$ encourages the model to preserve the global structure of the scene, leading to better generalization.


\subsection{Limitations and Further Applications}
While our proposed framework demonstrates improvements in depth estimation for Martian terrains, certain aspects deserve further exploration to broaden the applicability and address future works.
Despite its strong performance in synthetic datasets, our model still has room for improvement when applied to real-world data from the Zhurong Rover, which operates in the relatively flat Utopia Planitia region. This area is characterized by smooth terrains and low-frequency variations, making it challenging to capture fine-grained details. While our framework mitigates some of these issues by amplifying low-frequency signals, reconstructing intricate surface features requires further research. Addressing this challenge may require developing novel strategies tailored to homogeneous and texture-less environments, such as integrating more advanced priors or domain-specific features. 
While semantic information, such as identifying rocks, sand, or craters, has the potential to enhance depth estimation by providing additional contextual cues, its integration comes with challenges. Incorporating semantic priors introduces computational overhead and often necessitates redesigning the network for joint learning, complicating the training process. Moreover, in regions with flat terrains or sparse visual features, semantic information may lack sufficient resolution to offer meaningful guidance, which could limit its effectiveness in improving depth estimation accuracy. As such, while the inclusion of semantic priors could be explored in future work, its impact on depth estimation needs further investigation.

%


\section{Conclusion}\label{conclusion}
In this paper, we proposed M\textsuperscript{3}Depth, a depth estimation model specifically designed for Mars rovers to address the challenges posed by the sparse and unstructured terrain on Mars. Our approach focused on tackling two critical issues, the dominance of low-frequency features in Martian imagery and the lack of explicit geometric constraints. To address the former, we incorporated a wavelet-enhanced convolutional kernel. It is capable of capturing more low-frequency components and effectively expanding the receptive field. To overcome the latter, we introduced a consistency loss that explicitly modeled the geometric complementary relationship between depth maps and surface normal maps. By leveraging surface normal as an auxiliary modal to enhance depth estimation accuracy. Additionally, we developed an iterative refinement module with a mutual boosting mechanism that jointly enhance both depth and surface normal predictions in a complementary manner. Finally, we validated the effectiveness of M\textsuperscript{3}Depth on SimMar6k dataset with ground-truth depth labels and further evaluated in real-world Martian scenarios. Experimental results demonstrated that our method outperformed SOTA approaches across multiple metrics, showcasing its accuracy and adaptability to challenging real Martian environments. 

\section*{Acknowledgments}
We thank the release of the awesome dataset SimMars6K that makes this study possible. We also acknowledge China’s Mars Mission Program for supplying the Tianwen-1 image data. The Tianwen-1 datasets used in this paper are processed and produced by Ground Research and Application System of China’s Lunar and Planetary Exploration Program. These dataset are available for application and downloaded at \url{http://moon.bao.ac.cn/}. Additionally, we sincerely thank all anonymous reviewers and editors for their insightful comments and helpful suggestions.

%
\bibliography{tgrs_r1}
\bibliographystyle{IEEEtran}

\begin{IEEEbiography}
[{\includegraphics[width=1in,height=1.25in,clip,keepaspectratio]{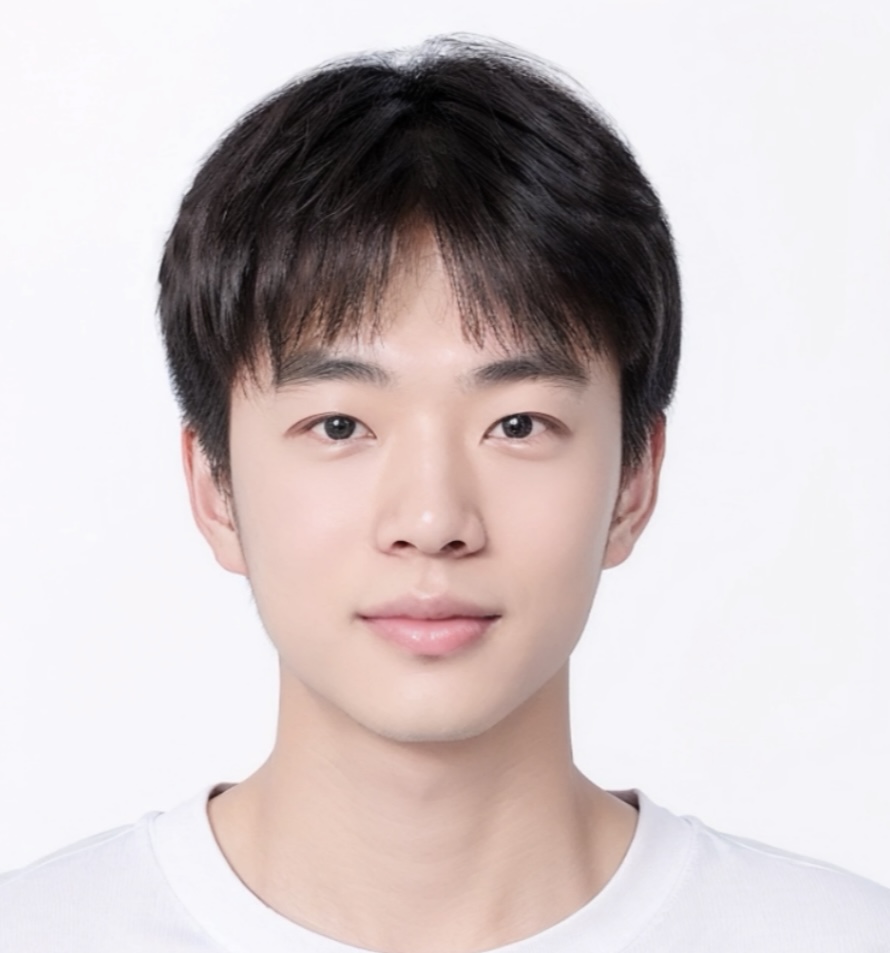}}]{Junjie Li} (Graduate Student Member, IEEE)
received the B.S. degree in 2022 from Beijing University of Posts and Telecommunications, China. He is currently pursuing his Ph.D. degree in Beijing University of Posts and Telecommunications, China. His research interests include wireless multimedia transmission and Mars scene perception. Email: junjie@bupt.edu.cn
\end{IEEEbiography}

\begin{IEEEbiography}
[{\includegraphics[width=1in,height=1.25in,clip,keepaspectratio]{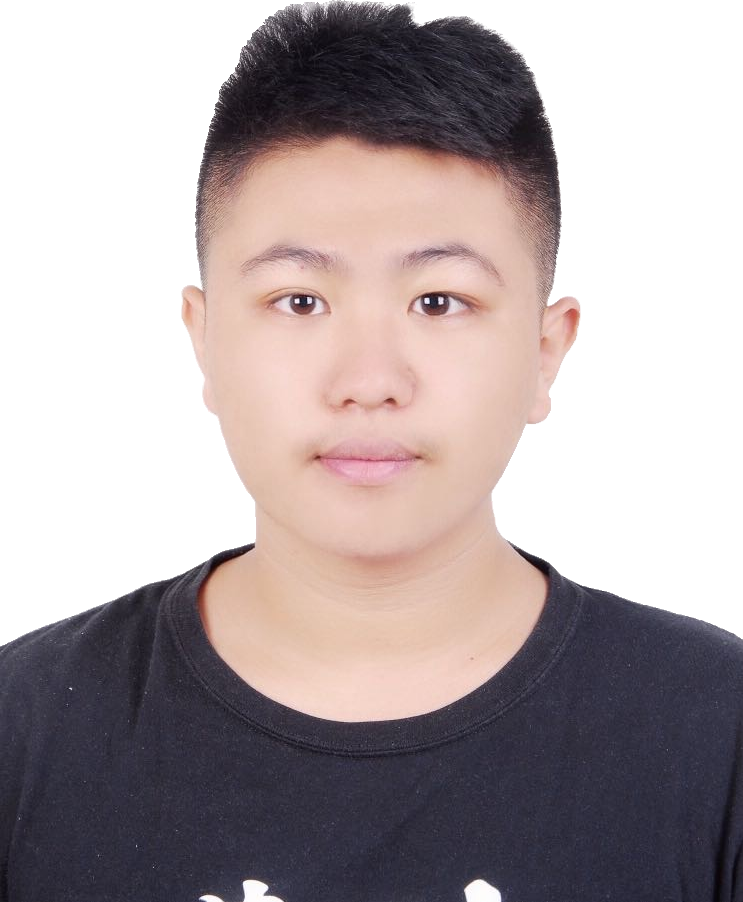}}]{Jiawei Wang} (Graduate Student Member, IEEE)
received the B.S. and Master degree from Beijing University of Posts and Telecommunications in 2020 and 2023, respectively. He is currently working toward the Ph.D. degree in State Key Laboratory of Networking and Switching Technology, Beijing University of Posts and Telecommunications. His research interests include Computer Vision and Embodied Artificial Intelligence.
\end{IEEEbiography}

\begin{IEEEbiography}
[{\includegraphics[width=1in,height=1.25in,clip,keepaspectratio]{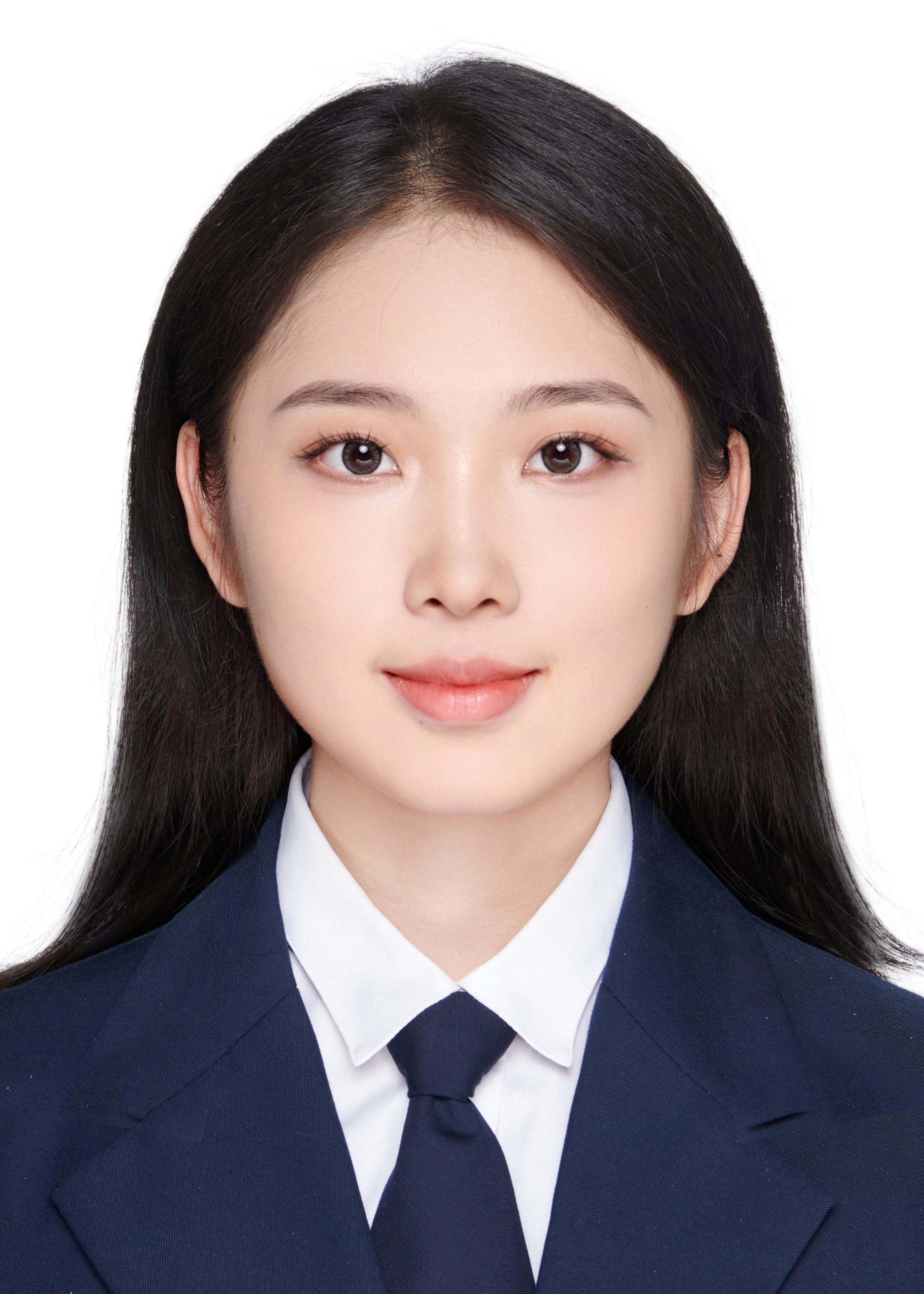}}]{Miyu Li} received the B.S. degree in 2024 from Beijing University of Posts and Telecommunications, China. She is currently pursuing his master degree in Beijing University of Posts and Telecommunications, China. Her research interests include change detection and terrain classification. Email: limy\_bupt@foxmail.com
\end{IEEEbiography}


\begin{IEEEbiography}[{\includegraphics[width=1in,height=1.25in,clip,keepaspectratio]{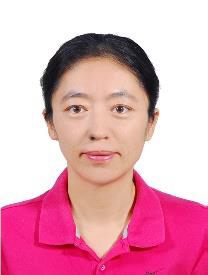}}]{Yu Liu}
(Member, IEEE) received her B.S. degree in automatic control and Ph.D. degree in signal and information processing in 2001 and 2006 respectively, both from Beijing University of Posts and Telecommunications, China. Dr. Liu is currently an associate professor and a doctoral supervisor with Beijing University of Posts and Telecommunications, China. Her main research areas include compressive sensing, blockchain security, and Mars images processing. Email: liuy@bupt.edu.cn
\end{IEEEbiography}

\begin{IEEEbiography}[{\includegraphics[width=1in,height=1.25in,clip,keepaspectratio]{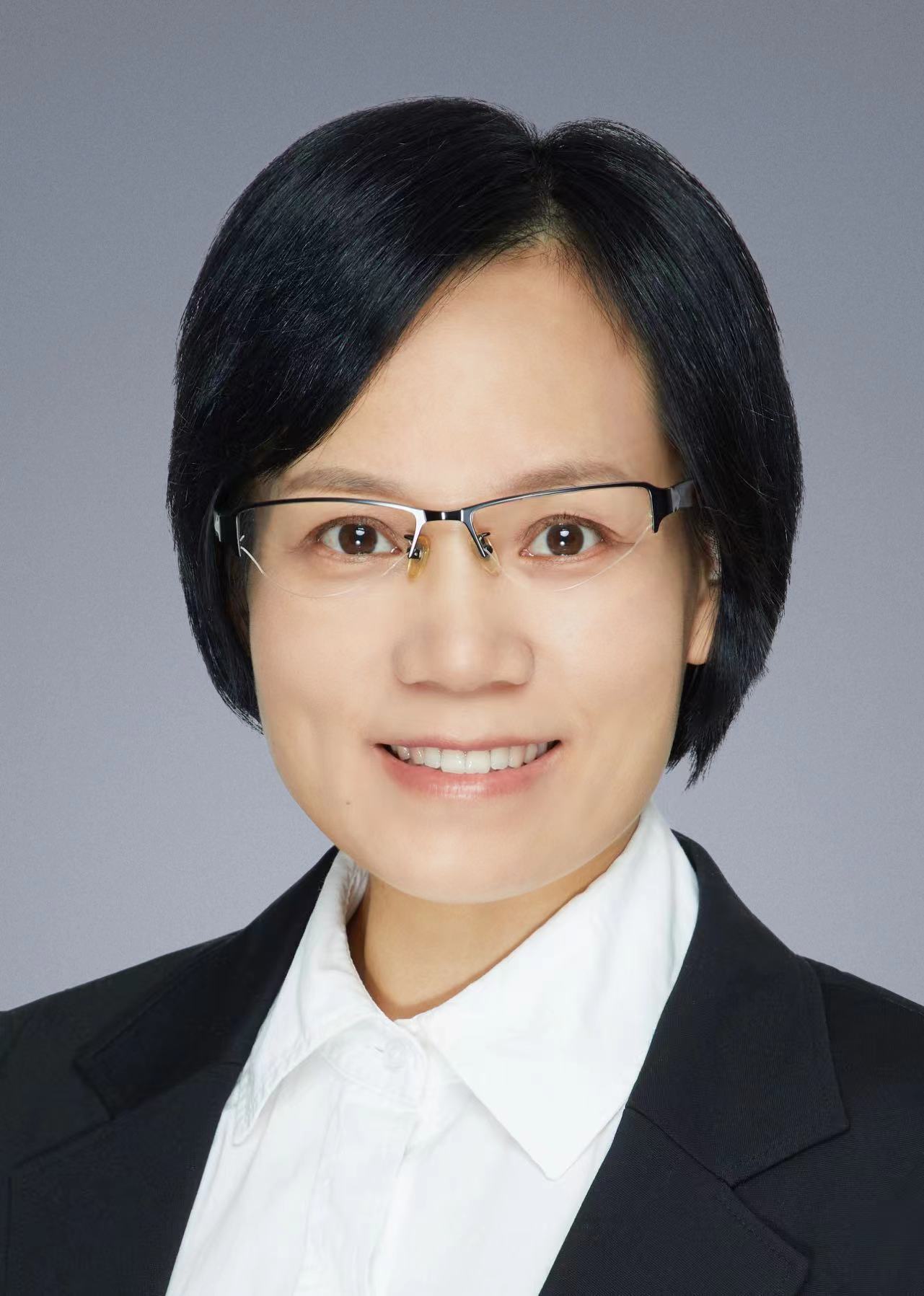}}]{Yumei Wang}
(Member, IEEE) received the Ph.D. degree in Signal and Information Processing from Beijing University of Posts and Telecommunications in 2004. She worked as an intern researcher in YRP center of NTT DoCoMo, Japan, in the summer of 2001. From April 2011 to April 2012, she was working as a visiting scholar in Lehigh University, USA. Currently she is an associate professor in Beijing University of Posts and Telecommunications. Her research interests include multimedia signal processing, cross-layer design for wireless multimedia transmission, and distributed video coding. Email: ymwang@bupt.edu.cn
\end{IEEEbiography}

\begin{IEEEbiography}
[{\includegraphics[width=1in,height=1.25in,clip,keepaspectratio]{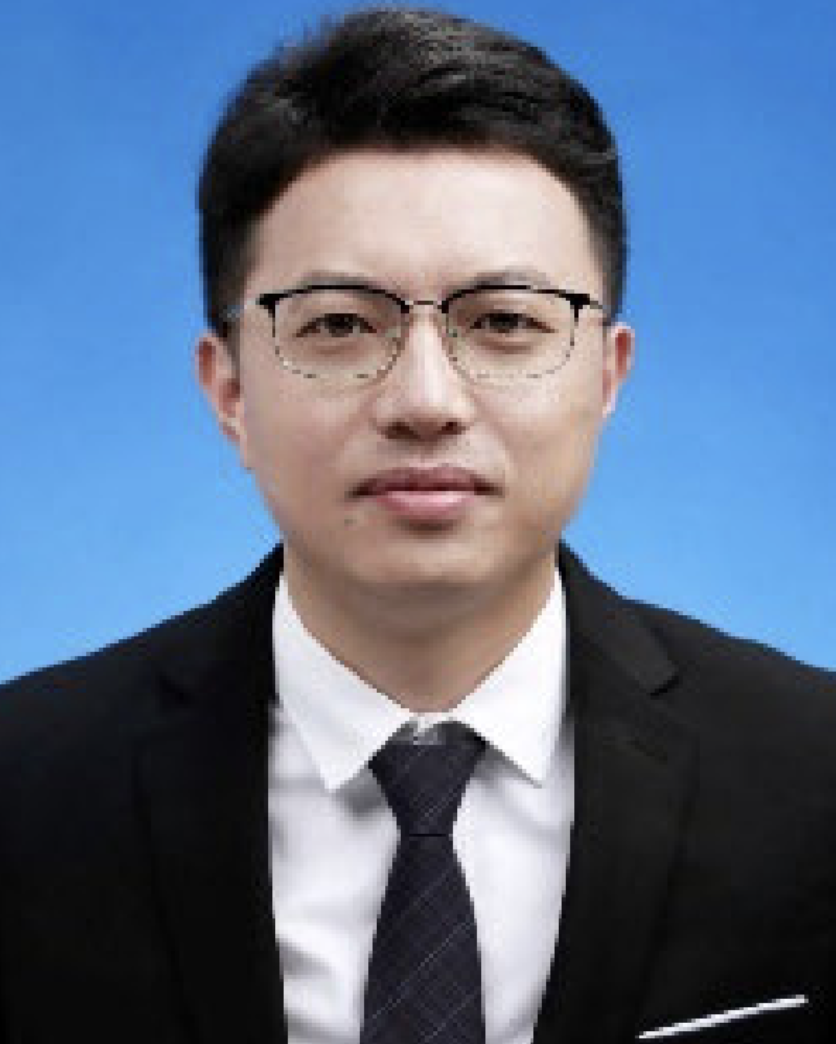}}]{Haitao Xu} received the B.E. and M.S. degrees in electrical engineering from Beijing Jiaotong University, Beijing, China, in 2005 and 2008, respectively. He received the Ph.D. degree in computer application technology with the University of Chinese Academy of Sciences, Beijing, in 2024, under the supervision of Dr. Changbin Xue. 

He is currently a Professor and a Master’s Supervisor with the Key Laboratory of Electronics Information Technology for Complex Space Systems, National Space Science Center, Chinese Academy of Sciences, Beijing. His research interests include aerospace-integrated electronic system technology and space-intelligent information processing technology. Email: xuhaitao@nssc.ac.cn
\end{IEEEbiography}

\vspace{11pt}
\vspace{11pt}
\vfill

\clearpage

\appendices\label{appendix}
\section{Consistency Constraints Formulation Between Depth And Surface Normal}
Under the pinhole camera model, the spatial gradient of the depth map in the pixel coordinate system is computed by integrating information from both the depth and surface normal maps. As illustrated in \autoref{pinhole_camera_o}, $(u,v)$ represents the pixel coordinate corresponding to the 3D point $(X,Y,Z)$, and $(u_c,u_v)$ denotes the pixel coordinate of the camera's optical center. Additionally, $f_x$ and $f_y$ represent the focal lengths along the X-axis and Y-axis, respectively.

Based on fundamental principles of photogrammetry, the pixel coordinates $(u,v)$ can be transformed into three-dimensional spatial coordinates $(X,Y,Z)$, as Eq.\eqref{eq1} depicts. 

\begin{equation}
\begin{bmatrix} u \\ v \\ 1 \end{bmatrix} = \frac{1}{Z} \begin{bmatrix} f_x & 0 & u_c \\ 0 & f_y & v_c \\ 0 & 0 & 1 \end{bmatrix} \begin{bmatrix} X \\ Y \\ Z \end{bmatrix}.
\label{eq1}
\end{equation}

Based on the pinhole camera model and the chain rule, we can yield equation below:
\begin{equation}
\begin{aligned}
&X=\frac{Z(u-u_c)}{f_x}\implies\frac{\partial X}{\partial u}=\frac{u-u_c}{f_x}\frac{\partial Z}{\partial u}+\frac{Z}{f_x}, \\
&Y=\frac{Z(v-v_c)}{f_y}\implies\frac{\partial Y}{\partial u}=\frac{v-v_c}{f_y}\frac{\partial Z}{\partial u}.
\end{aligned}
\label{eq2}
\end{equation}

Likewise, we can obtain the following equation:

\begin{equation}
\left\{
\begin{aligned}
\frac{\partial X}{\partial v} &= \frac{u-u_c}{f_x}\frac{\partial Z}{\partial v}. \\
\frac{\partial Y}{\partial v} &= \frac{v-v_c}{f_y}\frac{\partial Z}{\partial v} + \frac{Z}{f_y}.
\end{aligned}
\right.
\label{eq3}
\end{equation}

Depth and surface normal information are inherently related through the geometry of the 3D scene. To ensure consistency between these two modalities, we consider two types of constraints as follows.

\textit{Constraints 1: Geometric Mathematical Constraint.} 

The first constraint is derived from the mathematical relationship between depth gradients and surface normals. The surface normal can be expressed as the gradient of an implicit function, and this relationship provides a explicit linkage between depth gradients and normal vectors.

Under the surface function assumption, the scene can be represented as an implicit function $F(X,Y,Z)=0$. The surface normal map, denoted as $\vec{n} = (n_x,n_y,n_z)$, can be interpreted as the gradient of this function:

\begin{equation}
\nabla F=(\frac{\partial F}{\partial X},\frac{\partial F}{\partial Y},\frac{\partial F}{\partial Z}) = (n_x,n_y,n_z).
\label{eq4}
\end{equation}

Then, by taking the partial derivatives of $F(X,Y,Z)=0$ with respect to $X$ and $Y$, we can derive:
\begin{equation}
\left\{
\begin{aligned}
&\frac{\partial F}{\partial X}+\frac{\partial F}{\partial Z}\frac{\partial Z}{\partial X}=0 \implies \frac{\partial Z}{\partial X}=-\frac{\frac{\partial F}{\partial X}}{\frac{\partial F}{\partial Z}} = -\frac{n_x}{n_z}. \\
&\frac{\partial F}{\partial Y}+\frac{\partial F}{\partial Z}\frac{\partial Z}{\partial Y}=0 \implies \frac{\partial Z}{\partial Y}=-\frac{\frac{\partial F}{\partial Y}}{\frac{\partial F}{\partial Z}} = -\frac{n_y}{n_z}.
\end{aligned}
\right.
\label{eq5}
\end{equation}

By integrating from Eq.\eqref{eq2} to Eq.\eqref{eq5}, we can derive the depth gradient in the pixel coordinate space $(u,v)$ as follows:

\begin{equation}
\left\{
\begin{aligned}
&\left(\frac{\partial Z}{\partial u}\right)_1 =\frac{\partial Z}{\partial X}\frac{\partial X}{\partial u}+\frac{\partial Z}{\partial Y}\frac{\partial Y}{\partial u}, \\
&\left(1+\frac{n_x}{n_z}\frac{u-u_c}{f_x}+\frac{n_y}{n_z}\frac{v-v_c}{f_y}\right)\frac{\partial Z}{\partial u}=-\frac{n_xZ}{n_zf_x},\\
&\left(\frac{\partial Z}{\partial u}\right)_1=\frac{\left(\frac{-n_xZ}{n_zf_x}\right)}{1+\frac{n_x}{n_z}\frac{u-u_c}{f_x}+\frac{n_y}{n_z}\frac{y-u_c}{f_y}}.
\end{aligned}
\right.
\end{equation}

Similarly, we can yield equation below:

\begin{equation}\left(\frac{\partial Z}{\partial v}\right)_1=\frac{(\frac{-n_yZ}{n_zf_y})}{1+\frac{n_x}{n_z}\frac{u-u_c}{f_x}+\frac{n_y}{n_z}\frac{v-v_c}{f_y}}\end{equation}

\textit{Estimate 2: Spatial Gradient Constraint.}

The spatial gradient of the depth map can be efficiently estimated using standard image processing techniques, such as the Sobel filter, which approximates local depth variations in the image. This gradient captures changes in depth along the horizontal and vertical directions, providing a data-driven estimate of the surface geometry structure. The spatial gradient of the depth map can be computed by using a Sobel filter:

\begin{equation}
\begin{aligned}
\left(\frac{\partial Z}{\partial u},\frac{\partial Z}{\partial v}\right)_2=\left(\frac{\Delta Z}{\Delta u},\frac{\Delta Z}{\Delta v}\right). \\
\end{aligned}
\end{equation}

The consistency constraint is formulated as deviation between the two constraints of $\left(\frac{\partial Z}{\partial u}, \frac{\partial Z}{\partial v}\right)_1$ and $\left(\frac{\partial Z}{\partial u}, \frac{\partial Z}{\partial v}\right)_2$. The consistency constraint is further formulated as a consistency loss function, enabling it to guide model training by aligning the two gradient estimates. This loss is defined as the Huber norm of their deviation, as shown in Eq.\eqref{eq9}. 

\begin{equation}
\mathcal{L}_c=\left|\left(\frac{\partial Z}{\partial u},\frac{\partial Z}{\partial v}\right)_1-\left(\frac{\partial Z}{\partial u},\frac{\partial Z}{\partial v}\right)_2\right|_{\boldsymbol{H}}.
\label{eq9}
\end{equation}

\section{Module Parameter Analysis}
In this section, we analyze the impact of different parameter configurations on the performance of our model, specifically focusing on the number of iterations in the IRM and the number of levels used in the WEFE.

The performance of the IRM module is closely related to the number of refinement iterations. We evaluated settings ranging from 1 to 5 iterations, as shown in \autoref{irm_iterations}. We observe an upward trend in accuracy, particularly between 1 and 3 iterations, indicating that iterative refinement contributes to depth consistency. While the best results are achieved at 5 iterations, we also conducted preliminary tests at 6 iterations, which yielded only marginal improvements ($\delta_1$ increased from 0.912 to 0.913, and Abs Rel remained almost unchanged). This suggests that performance gains begin to saturate beyond 5 iterations. 

\begin{figure}[thb] \centering
    \includegraphics[width=0.48\textwidth]{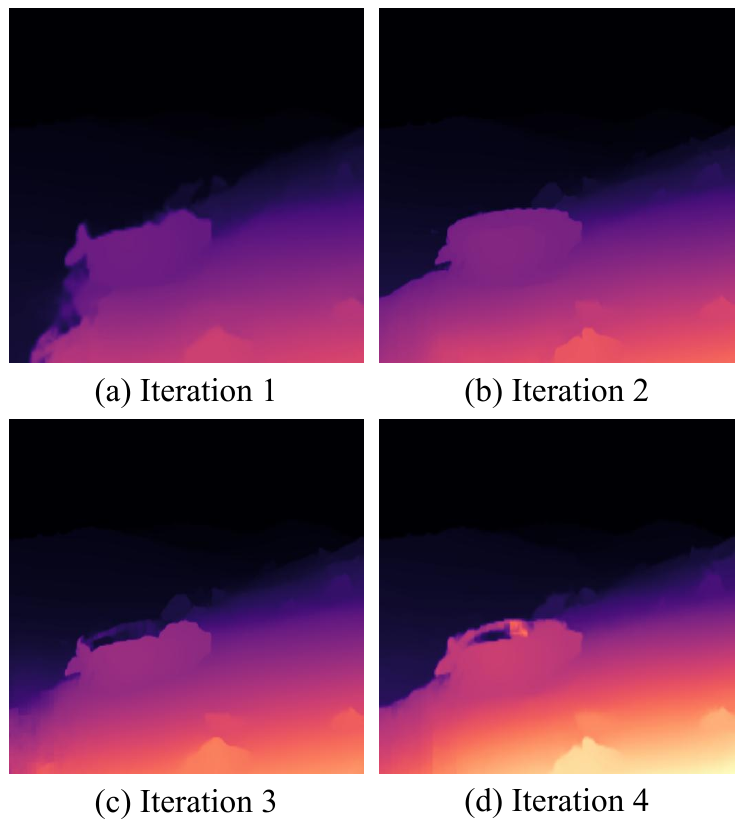}
    \caption{Visualization of dataset examples and depth distribution characteristics.} 
    \label{irm_visualization}
\end{figure}

To provide further insight into the impact of different iteration counts, we present visual examples of the depth maps generated at various stages of refinement. As shown in \autoref{irm_visualization}, the depth maps exhibit noticeable improvements from 1 to 3 iterations, particularly in texture-less regions, where finer details and depth consistency are progressively refined. However, the performance improvements become increasingly marginal. This diminishing return beyond 5 iterations is further confirmed by both the quantitative results and visual analysis. These observations highlight that, 5 iterations may achieve the best overall results, and the performance gains gradually decrease, indicating an optimal balance between accuracy and computational efficiency.

\begin{table}[htbp]
\centering
\caption{Effect of Iteration Numbers in the IRM.}
\label{irm_iterations}
\resizebox{0.48\textwidth}{!}{
\begin{tabular}{c||cccc}
\toprule
\textbf{Iterations Number} & \textbf{Abs Rel} & \textbf{Sq Rel} & \textbf{RMSE} & \textbf{$\delta_1$} \\
\midrule
1  & 0.112 & 0.076 & 0.368 & 0.876 \\
2  & 0.098 & 0.071 & 0.338 & 0.885 \\
3  & 0.089 & 0.065 & 0.314 & 0.905 \\
4  & 0.086 & 0.061 & 0.310 & 0.910 \\
5  & 0.084 & 0.059 & 0.308 & 0.912 \\
\bottomrule
\end{tabular}
}
\end{table}

The number of levels in the WEFE is another critical parameter that affects the model's performance. Increasing the number of decomposition levels allows the network to capture more global context but also increases the complexity. We evaluated different configurations of WT levels, ranging from 2 to 5, and observed the effect on depth prediction accuracy. As shown in \autoref{wt_levels}, increasing the number of WT levels results in an improvement in depth prediction accuracy, with the best performance observed at 3 levels. Interestingly, we observe that increasing WT levels beyond 3 starts to degrade performance. This may be attributed to over-decomposition, where high-frequency spatial details that are important for pixel-level disparity estimation are excessively smoothed out. This leads to less discriminative features in local matching.

\begin{table}[htbp]
\centering
\caption{Effect of Wavelet Transform Levels on Performance.}
\label{wt_levels}
\resizebox{0.48\textwidth}{!}{
\begin{tabular}{c||cccc}
\toprule
\textbf{Wavelet Levels} & \textbf{Abs Rel} & \textbf{Sq Rel} & \textbf{RMSE} & \textbf{$\delta_1$}  \\
\midrule
2  & 0.095 & 0.076 & 0.330 & 0.890 \\
3  & 0.084 & 0.059 & 0.308 & 0.912 \\
4  & 0.088 & 0.063 & 0.318 & 0.905 \\
5  & 0.093 & 0.071 & 0.327 & 0.892 \\
\bottomrule
\end{tabular}
}
\end{table}

\section{Supplementary Experiments}\label{sup_exp}
\begin{figure}[!thb] \centering
    \includegraphics[width=0.46\textwidth]{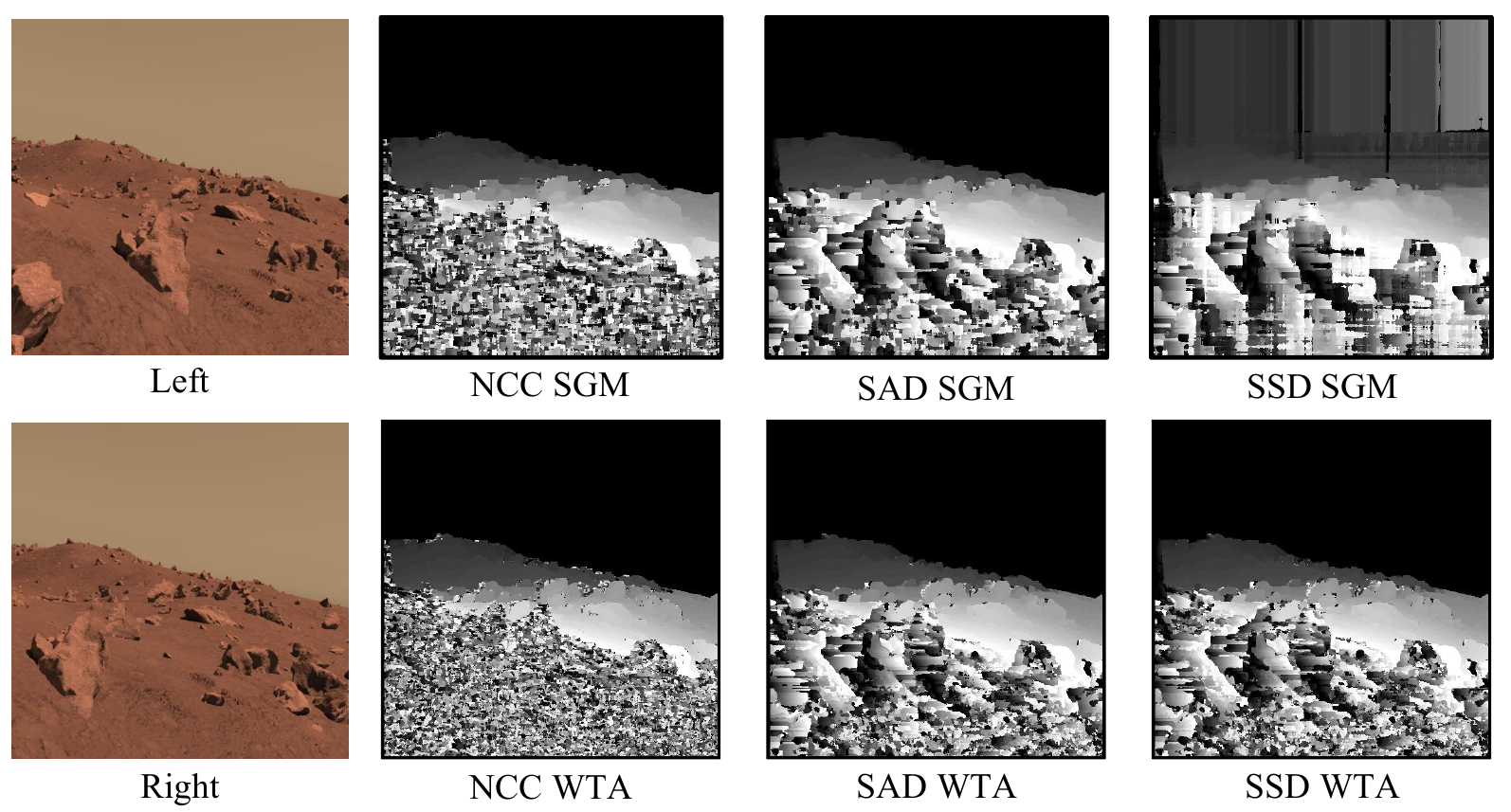}
    \caption{Visualization of dataset examples and depth distribution characteristics.} 
    \label{trad_match}
\end{figure}

Regarding the traditional stereo matching baselines, we conduct a comparative evaluation with classical methods commonly used. We evaluated classical cost volume construction methods, including the sum of squared differences (SSD), sum of absolute differences (SAD), and normalized cross-correlation (NCC). For disparity selection, both the Winner-Takes-All (WTA) strategy and the semi-global match optimization (SGM) are applied. As shown in \autoref{trad_match}, we compare SSD, SAD, and NCC as cost metrics, each paired with WTA or SGM for the selection of the disparities. Across all configurations, the resulting disparity maps exhibit substantial artifacts, such as speckle noise, depth discontinuities, and structural inconsistency. In particular, these methods suffer from severe fragmentation due to the absence of global smoothness constraints. Although SGM improves continuity to some extent, all methods still struggle in low-texture regions and around object boundaries. These visualizations highlight the limitations of traditional matching pipelines. In contrast, our learning-based method generates smoother and more coherent depth predictions, accurately preserving terrain geometry even in texture-less or ambiguous regions.



\end{document}